\documentclass[11pt]{article}

\usepackage[margin=1.5cm]{geometry}
\usepackage{pict2e}
\usepackage{tikz}
\usetikzlibrary{positioning}
\usepackage{sidecap}
\usetikzlibrary{decorations.pathreplacing}
\usepackage{amsmath}

\DeclareMathOperator*{\argmax}{argmax}
\usepackage{graphicx} %
\usepackage{amsmath,amssymb,amsfonts}
\usepackage{mathtools}

\usepackage[round]{natbib}

\usepackage{tikz}
\usetikzlibrary{arrows.meta, positioning, shapes.geometric}
\usepackage{animate}
\usepackage{float}   %
\usepackage{amsthm}
\usepackage{booktabs}
\usepackage{placeins} %
\usepackage{multirow}


\usepackage[colorlinks=true,
            linkcolor=blue,
            citecolor=blue,
            urlcolor=blue]{hyperref}
            
\newtheorem{theorem}{Theorem}[section]

\newtheorem{proposition}[theorem]{Proposition}
\newtheorem{corollary}[theorem]{Corollary}

\theoremstyle{plain}
\newtheorem{definition}[theorem]{Definition}

\theoremstyle{remark}

\newcommand{\monthyear}{%
    \ifcase\month
    \or January\or February\or March\or April\or May\or June%
    \or July\or August\or September\or October\or November\or December%
    \fi\space\number\year}

\newcommand{\AREA}{58}

\newcommand{\BIASlrpzero}{92}

\newcommand{\CNXN}{200}
\newcommand{\CNXNkeep}{173}
\newcommand{\CNXabpcgi}{0.063}

\newcommand{\CNXabpclrpnonezero}{0.216}
\newcommand{\CNXabpclrpzero}{0.216}

\newcommand{\CNXabpcxvn}{0.249}
\newcommand{\CNXabslrpab}{21}

\newcommand{\CNXkeeplrpnonezero}{41}

\newcommand{\CNXkeepxvmm}{53}
\newcommand{\CNXkeepxvn}{47}
\newcommand{\CNXlostzp}{45}

\newcommand{\CNXspearman}{0.12}

\newcommand{\CORRlrpzpsrp}{0.82}

\newcommand{\DABPCSElrpab}{0.003}

\newcommand{\DABPCSEsrpl}{0.001}
\newcommand{\DABPClrpab}{+0.000}

\newcommand{\DABPCsrpl}{-0.004}

\newcommand{\KINKN}{40}
\newcommand{\KINKdxo}{5\cdot10^{-6}}
\newcommand{\KINKgates}{176}
\newcommand{\KINKlrpab}{0.073}

\newcommand{\KINKlrpzero}{1.5}

\newcommand{\KINKlrpzp}{0.06}

\newcommand{\KINKpools}{2785}

\newcommand{\KINKunits}{13.6}
\newcommand{\KINKxvn}{0.11}

\newcommand{\NCLS}{50}

\newcommand{\NIMG}{200}
\newcommand{\NKEEP}{151}
\newcommand{\NPET}{200}

\newcommand{\RPN}{200}

\newcommand{\RPareaheadlrpzp}{21}

\newcommand{\RPcommonlrpzp}{3}

\newcommand{\RPcommonxvn}{2}

\newcommand{\RPdenslrpzp}{2.5}

\newcommand{\RPdensonelrpzp}{198}

\newcommand{\RPdensxvn}{2.1}

\newcommand{\RPerr}{8\cdot10^{-7}}

\newcommand{\RPhalffivelrpzp}{24}

\newcommand{\RPhalffivexvn}{31}

\newcommand{\RPhalffourlrpzp}{65}

\newcommand{\RPhalffourxvn}{87}

\newcommand{\RPncat}{63}
\newcommand{\RPndog}{137}
\newcommand{\RPpixheadlrpzp}{51}

\newcommand{\RPpixheadxvn}{46}

\newcommand{\RPpurityabovelrpzp}{200}

\newcommand{\RPpuritylrpzp}{76}

\newcommand{\RPpurityxvn}{73}

\newcommand{\RPseen}{581}

\newcommand{\RPsharebiaslrpzp}{17}

\newcommand{\RPsharebiasxvn}{9}

\newcommand{\RPshareheadlrpzp}{42}

\newcommand{\RPshareheadxvn}{42}

\newcommand{\RPskipboxcropped}{257}
\newcommand{\RPskipnobox}{2}
\newcommand{\RPskipnotpet}{24}
\newcommand{\RPskipsmallregion}{98}

\newcommand{\TIMESRPLMED}{93}

\title{Relevance Propagation as Entropic Optimal Transport on Filippov Neural Networks: Sinkhorn Propagation and Region Relevance in Image Classifiers}
\author{
\begin{tabular}{c}
	Olivier Gibaru \\ %
    Arts et Métiers Institute of Technology \\ 
    \url{olivier.gibaru@ensam.eu}
\end{tabular}
\begin{tabular}{c}
	  Eric Nyiri \\ %
    Arts et Métiers Institute of Technology\\ 
    \url{eric.nyiri@ensam.eu}
\end{tabular}
}
\date{\monthyear}

\begin{document}

\maketitle

\begin{abstract}
Relevance propagation, which explains a network's prediction by redistributing it backward through the layers, is usually treated as a heuristic. We show that it is instead the solution of a single variational problem, and this entropic optimal-transport formulation of relevance is our main contribution. Layer-wise Relevance Propagation (LRP) is exactly the divergence-free flow on the active computational graph of zero relative entropy to the backward transition law of the network's forward activation flow; for a non-negative reference, this is a relative entropy between Markov measures on backward paths, as in Schrödinger's problem. The classical LRP$_{z^+}$ and LRP$_{\alpha,\beta}$ rules are recovered from this one principle, LRP$_0$ included, by taking as reference the signed forward contributions themselves, through a signed extension of that entropy, and all three are instances of a single rule, LRPN$_{\alpha,\beta}$, whose weights may vary from node to node, for instance with the activity of the node, in the spirit of Hebb's principle: a more active node then sends more of its relevance to its excitatory contributions. This rests on a new, measure-theoretic formalization of feed-forward networks as directed acyclic graphs of \emph{Filippov} operators. It covers smooth and piecewise-smooth architectures (ReLU, max-pooling) alongside outright discontinuous ones, such as hard-routed Mixture-of-Experts, and carries a Clarke--Filippov generalized-gradient calculus that stays well defined wherever the network fails to be differentiable or continuous; on networks of ReLU and max-pooling nodes, LRP$_0$ is Gradient$\times$Input, with an explicit jump across an isolated kink, where LRP$_{z^+}$ stays continuous, and every rule may jump where the winner of a max-pooling node changes. Recast as a discrete-time optimal control problem, in the spirit of LeCun's original derivation, the same formalism gives backpropagation itself a variational characterization: the adjoint (costate) equation of a Lagrangian relaxation. A new rule then follows under the same constraint: taking the entropy edge by edge, rather than between backward transition laws, yields Sinkhorn Relevance Propagation, computed by a Sinkhorn-type iteration. Aggregated over a region of the input, the flow of a non-negative reference tells, with exact accounting, which hidden units carry the evidence of that region, the conditioned flow being a Schrödinger bridge whose potentials follow from one forward pass. On VGG-16, the conservation identities hold to round-off, and the rules, which share the same constraint and differ by their reference, are told apart by the sign of their maps, the share of the output they leave to the bias nodes, their faithfulness, their localization and their dependence on the class explained; on ConvNeXt-Tiny, whose hidden values take both signs, LRPN$_{\alpha,\beta}$ conserves the output, and its Hebbian weights, set by the activations of each layer, give the highest faithfulness (ABPC) of the rules tested.
\end{abstract}

\section{Introduction}
\label{sec:intro}

Artificial neural networks routinely outperform humans on narrow,
well-specified tasks -- from image classification to language
modeling -- yet the internal mechanisms behind their predictions
remain largely opaque. This ``black-box'' character, present since
\citet{Rosenblatt1958}'s perceptron, now raises acute concerns of trust,
reliability, and accountability wherever these models are deployed in
high-stakes decisions -- medical diagnosis, autonomous systems,
financial decision-making -- and has motivated a decade of work in
eXplainable Artificial Intelligence (XAI).

Two broad families of attribution methods have emerged to answer the
question of \emph{why} a network produces a given output, each
capturing a complementary facet of its behavior. \emph{Relevance
propagation}, epitomized by Layer-wise Relevance Propagation (LRP)
\citep{Bach2015}, decomposes the output score as a sum of
contributions from individual input features by propagating a
conservation signal backward through the network; it tells how the
output is shared among the inputs, not how it would respond to a
perturbation of them. \emph{Gradient-based
attribution}, such as Gradient$\times$Input
\citep{Simonyan2014,Baehrens2010}, quantifies the sensitivity of the
output to infinitesimal perturbations; it tells how the output would
respond, but ignores the global flow of information through the
network. Until now, LRP has been justified by local conservation and by
deep Taylor decomposition \citep{Montavon2017}, which derives its rules
from local expansions, but not as the solution of a stated optimization
problem. In earlier, more applied
work \citep{NyiriGibaru2025}, we proposed Relative LRP (R-LRP), a rule
free of hyperparameters that avoids the divisions by small values of
the classical rules by satisfying the conservation law only up to a
multiplicative factor, and compared it with the standard LRP variants
on image classification with convolutional, VGG and ResNet networks;
that experience is what calls for the principled foundation developed
here.

This paper replaces the conservation heuristic by a variational principle, which
also relates the two families: relevance propagation is
exactly the \emph{entropic projection} of the backward transition law of
the forward activation flow onto the set of divergence-free relevance
flows on the active computational graph, the unique such flow with zero
relative entropy to that law; for a non-negative reference, this
relative entropy is the one of two Markov measures on the backward paths
of the network. This principle recovers the classical LRP$_{z^+}$
and LRP$_{\alpha,\beta}$ rules as special cases, LRP$_0$ through a signed
extension of the same relative entropy, and gathers the three into one
rule whose weights may vary from node to node, for instance with the
activity of the node, in the spirit of Hebb's principle
\citep{Hebb1949}; LRP$_0$ is
Gradient$\times$Input on networks of ReLU and max-pooling nodes. This entropic
optimal-transport formulation of relevance is our main contribution:
the rules become the choice of a reference, conservation becomes a
constraint rather than a heuristic, and the tools of entropic transport
-- Sinkhorn iterations, Schrödinger bridges -- become available for
attribution.
Carrying this program out rigorously demands a foundation broader than
the smooth or piecewise-smooth networks classical LRP was built for:
architectures such as hard-routed Mixture-of-Experts
\citep{Shazeer2017,DeepSeekV3_2024} can be outright \emph{discontinuous},
so neither the classical nor the Clarke generalized gradient is
everywhere defined, and a gradient-based attribution applied to them
naively rests on undefined quantities exactly where the routing
decision changes.

The paper is organized as follows. Section~\ref{sec:feedforward} lays
the foundation: a new, measure-theoretic formalization of feed-forward
networks as directed acyclic graphs of \emph{Filippov} operators, with
a Clarke--Filippov generalized-gradient calculus yielding well-defined
\emph{sets} of admissible values wherever the network is
non-differentiable or discontinuous; recast as a discrete-time optimal
control problem in the spirit of \citet{LeCun1988}, the same
formalism gives backpropagation itself a variational characterization,
as the adjoint equation of a Lagrangian relaxation. Section~\ref{sec:LRP}
builds the central variational principle on this foundation, recovering
LRP$_{z^+}$ and LRP$_{\alpha,\beta}$ as the unique solutions of
relative-entropy minimization problems between backward transition
laws, LRP$_0$ included, its reference being the signed forward
contributions themselves. The three are instances of
LRPN$_{\alpha,\beta}$, a rule that splits the relevance of each node $v$
between its excitatory and its inhibitory contributions with weights
$\alpha_v$ and $\beta_v=1-\alpha_v$ that may change from node to node.
With $\bar x_v$ the value of $v$ divided by the largest value of its
layer, LRPN$_{\alpha(\bar x),\beta(1-\bar x)}$ takes $\alpha_v=\bar x_v$
and $\beta_v=1-\bar x_v$: in the spirit of Hebb's principle, a more
active node then sends more of its relevance to its excitatory
contributions. On networks of
ReLU and max-pooling nodes, each of these rules is an input times a
modified backward pass -- Gradient$\times$Input for LRP$_0$,
which jumps across a kink where LRP$_{z^+}$ stays continuous, whereas
every rule may jump where the winner of a max-pooling node changes. Taking instead the
relative entropy edge by edge, as in entropically regularized flows,
Section~\ref{sec:LRP} introduces Sinkhorn Relevance Propagation (SRP), the conservative flow
closest to the whole forward activation flow, computed by a
Sinkhorn-type iteration, and closes with applications to two image
classifiers, VGG-16 and ConvNeXt-Tiny
(Appendix~\ref{sec:architectures}). On VGG-16, the rules conserve the
output to round-off and are told apart by the sign of their maps, the
share of the output they leave to the bias nodes, their faithfulness,
their localization and their dependence on the class, SRP
following the edges and textures of the lower layers, and a
perturbation of standard deviation $10^{-5}$ moves the LRP$_0$ map by
$\KINKlrpzero\%$, against $\KINKlrpzp$ to $\KINKxvn\%$ for the other
rules, consistently with Corollary~\ref{thm:RP0ReLU}; on ConvNeXt-Tiny,
whose hidden values take both signs, LRPN$_{\alpha,\beta}$ conserves the
output and LRPN$_{\alpha(\bar x),\beta(1-\bar x)}$ has the highest ABPC of
the rules tested, whereas LRP$_{z^+}$ loses part of the output and LRP$_0$ is no
longer Gradient$\times$Input.
Section~\ref{sec:transformers} aggregates the relevance flow of a
non-negative reference over a region of the input: it tells which hidden units carry the evidence of
that region, with exact accounting, the flow conditioned on the region
being a Schrödinger bridge whose potentials follow from one forward
pass; on $\RPN$ annotated photographs of cats and dogs and VGG-16, with
LRP$_{z^+}$ and LRPN$_{\alpha(\bar x),\beta(1-\bar x)}$, the head receives
about $\RPdenslrpzp$ times as much relevance per pixel as the body, through
largely different channels.
Section~\ref{sec:conclusion} concludes and discusses perspectives.

\FloatBarrier

\section{Feed-forward Filippov neural networks}
\label{sec:feedforward}

This section lays the foundation of the paper. After the setting and the
architectures that motivate it (Section~\ref{sec:setting}), it
introduces the Filippov regularization of a switching operator
(Section~\ref{sec:filippov-regularization}), describes a network as a
directed acyclic graph of such operators (Section~\ref{sec:dag-networks}),
defines its forward propagation, its admissible branch selections, its
active graph and the Clarke--Filippov calculus of its output
(Section~\ref{sec:forward-active}), and casts backpropagation as the
adjoint equation of a discrete-time optimal control problem
(Section~\ref{sec:backprop-oc}).

\subsection{Setting}
\label{sec:setting}
The typical setting for supervised learning is to find an approximation of an unknown function $f$, given a training data set of its point values: $\{x^{(i)},f(x^{(i)})\}_{i=1,...,m}$ where $x^{(i)} \in \mathbb{R}^d$ and $f(x^{(i)}) \in \mathbb{R}^p$. For example, in classification problems for images, $d$ is the number of pixels $\times$ the number of channels in the images, typically somewhere in the range of $10^3$ to $10^6$, and for videos it is even higher. The learning problem is then to numerically produce from this data a function $\hat{f}$ that is in some sense a good predictor of $f$ on new unseen draws $x \in \mathbb{R} ^d$. In the classical formulation the input space is $\mathbb{R}^d$,
where each coordinate corresponds to a fixed feature
(e.g. a pixel in an image).

Large language models such as Claude \citep{Anthropic2023Claude}, Mistral \citep{Jiang2023Mistral,Jiang2024Mixtral}, 
GPT \citep{Brown2020GPT3,
OpenAI2023GPT4}, LLaMA \citep{Touvron2023LLaMA,Touvron2023LLaMA2} or DeepSeek \citep{DeepSeekV3_2024,dai2024deepseekmoe}, 
are trained on collections of tokenized documents rather than on fixed-dimensional feature vectors.
A training sample is a sequence of tokens
$
x=(t_1,\ldots,t_n)\in\mathcal T^n,
$
where $\mathcal T$ is a finite vocabulary and $n$ may vary from one example to another.
The learning task consists in approximating the
conditional distribution $p(t_{n+1}\mid t_1,\ldots,t_n)$,
that is, predicting the next token given the previous context.
After embedding, each token is represented by a vector in a latent space $\mathbb R^d$ and the input to the network becomes a sequence
$
X=(x_1,\ldots,x_n)
\in
(\mathbb R^d)^n .
$
Modern foundation models are typically based on
transformer architectures combining attention layers, feed-forward networks and normalization operators. 
Several recent models additionally employ Mixture-of-
Experts (MoE) \citep{Shazeer2017} layers in order to increase the number of parameters without increasing the computational cost
proportionally. In a MoE block, a routing network computes a score vector. A subset of
experts is then selected according to the largest routing scores. In the hard-routing setting, the selection is typically performed through a top-$k$ operator or an
argmax rule.

Unlike attention and feed-forward blocks, which remain smooth or at worst piecewise-smooth (as with ReLU),
the hard-routing level selects a different expert map across routing boundaries, where the network map itself, not merely one of its derivatives, may jump, unless the experts agree there. Neither the classical gradient nor the Clarke generalized gradient is then defined at such points, so that a gradient-based explanation applied naively to a hard-routed architecture rests on undefined quantities exactly where the routing decision changes, unless a selection convention is fixed. We therefore introduce, in place of ordinary differentiability, a Filippov set-valued regularization that remains well defined across these discontinuities, coincides with the classical gradient at regular points and contains the Clarke generalized gradient where the network is continuous.

\subsection{Filippov regularization}
\label{sec:filippov-regularization}

Let us introduce piecewise-$C^1$ switching operator.

\begin{definition}
    \label{def:piecewise-switching-operator}
    Let $\Omega\subseteq\mathbb R^n$ be an open set. A map
    $\Phi:\Omega\setminus\Sigma\to\mathbb R^m$ ( $m \in \mathbb N^*$) is called a
    \emph{piecewise-$C^1$ switching operator} on $\Omega$, with
    \emph{switching region} $\Sigma$, if there exist:
    \begin{itemize}
        \item a locally finite family $(\Omega^{(k)})_{k\in K}$ of pairwise disjoint
        open subsets of $\Omega$, indexed by a finite or countable set $K$, such
        that, writing
        \[
            \Sigma:=\Omega\setminus\bigcup_{k\in K}\Omega^{(k)},
        \]
        the complement satisfies $\mu(\Sigma)=0$ where $\mu$ denotes the
        Lebesgue measure -- an explicit requirement on the family, not
        automatic from local finiteness and disjointness alone;
        \item for every $k\in K$, an open set $U^{(k)}\subseteq\mathbb R^n$
        with $\overline{\Omega^{(k)}}\cap\Omega\subseteq U^{(k)}$ and a
        $C^1$ map $\Phi^{(k)}:U^{(k)}\to\mathbb R^m$, called the
        \emph{branch} of $\Phi$ on $\Omega^{(k)}$, such that
        $\Phi=\Phi^{(k)}$ on $\Omega^{(k)}$; we write $\Phi^{(k)}$ again
        for the restriction of this map to $\overline{\Omega^{(k)}}\cap
        \Omega$. In particular $\Phi^{(k)}$ is $C^1$, with continuous Jacobian
        $D\Phi^{(k)}\in\mathbb R^{m\times n}$, on a full open neighborhood
        of every point of $\overline{\Omega^{(k)}}\cap\Omega$, including
        points of the switching region $\Sigma$ itself, rather than merely
        admitting a continuous extension from inside $\Omega^{(k)}$.
    \end{itemize}
    For $x\in\Omega$, the \emph{active index set} of $\Phi$ at $x$ is
    $J_\Phi(x):=\bigl\{k\in K:x\in\overline{\Omega^{(k)}}\bigr\}$.
    We write $\Phi\in\mathcal S^1(\Omega,\mathbb R^m)$ for the class of all such
    operators, and simply $\Phi\in\mathcal S^1(\Omega)$ when $m=1$.
\end{definition}

The set $\Sigma$ is automatically closed in $\Omega$.
Moreover, we will show that $J_\Phi(x)$ is finite and nonempty for every $x\in\Omega$.
Since $\Phi^{(k)}$ is $C^1$ on the open neighborhood $U^{(k)}$, its
restriction to $\overline{\Omega^{(k)}}\cap\Omega$ automatically agrees
with the limit taken from inside $\Omega^{(k)}$,
\[
    \Phi^{(k)}(z)
    =
    \lim_{\substack{\zeta\to z\\\zeta\in \Omega^{(k)}}}
    \Phi(\zeta),
    \qquad
    \text{for every} \; z\in\overline{\Omega^{(k)}}\cap\Omega,
\]
so the branch is unambiguously determined by $\Phi$ on $\Omega^{(k)}$
alone; the open-neighborhood extension is what additionally guarantees
that $\Phi^{(k)}$ stays differentiable, not just continuous,
arbitrarily close to a switching point.

The partition $(\Omega^{(k)})_{k\in K}$ is motivated by the fact that
many operators used in modern neural architectures are naturally
described by different formulas on different regions of their input
space; any discontinuity or non-smooth behavior is then confined to the
switching region $\Sigma$, contained in the union of the region
boundaries.
A first example is the max-pooling operator
$
\Phi(z_1,\ldots,z_n)
=
\max_{1\le i\le n} z_i
$
on $\Omega=\mathbb R^n$. For each $k\in K=\{1,\ldots,n\}$, let
$
\Omega^{(k)}
=
\{z\in\mathbb R^{n}:
z_k > z_j \text{ for all } j\neq k\},
$
the open set where the $k$-th coordinate is the unique maximum. On
$\Omega^{(k)}$, the operator reduces to the linear map
$
\Phi^{(k)}(z)=z_k,
$
and the switching region
$\Sigma=\{z:\ z_i=z_j=\max_\ell z_\ell\ \text{for some}\ i\neq j\}$
is the set of ties between \emph{maximal} coordinates -- a closed subset
of finitely many hyperplanes, hence a null set. Max-pooling is thus
piecewise linear and continuous, and it is not differentiable exactly
on $\Sigma$, where several branches $\Phi^{(k)}(z)=z_k$ coincide in value
but have different gradients $e_k$.
A second example is a hard-routing mixture-of-experts block
\citep{Shazeer2017, Fedus2022, dai2024deepseekmoe, Jiang2024Mixtral}.
Let $g_1,\ldots,g_M$ be $C^1$ routing score functions and
$E_1,\ldots,E_M$ be $C^1$ expert maps. The routing rule selects the
expert
$
r(z)\in\argmax_{1\le j\le M} g_j(z),
$
and the local operator is
$
\Phi(z)
=
E_{r(z)}(z).
$
For each expert $j$, the routing region
$
\Omega^{(j)}
=
\{z:
g_j(z) > g_\ell(z)
\text{ for all }
\ell\neq j\}
$
is open, and on it the operator is the branch
$
\Phi^{(j)}=E_j.
$
The switching region is the set of routing ties
$g_j(z)=g_\ell(z)=\max_i g_i(z)$, $j\neq\ell$; it is a null set as soon as
no difference $g_j-g_\ell$ vanishes on a set of positive measure, which
holds for instance for affine scores $g_j(z)=a_j^{\top}z+c_j$ with
pairwise distinct $a_j$. Unlike max-pooling, the operator is in general
discontinuous across such a tie, since the two experts need not agree
there.

In both examples the local operator is $C^1$ inside each region
$\Omega^{(k)}$ and loses regularity only on $\Sigma$: differentiability
for max-pooling, possibly continuity for hard routing; the usual
activation functions are further special cases, listed after
Definition~\ref{def:DAG}. The switching region is precisely where
generalized notions of differentiation become relevant.
For max-pooling or ReLU operators,
the network map remains continuous and locally Lipschitz, so \citet{Clarke1975}'s generalized gradients are well defined.
For hard-routing architectures,
the network map may become discontinuous across routing boundaries. In that case the Clarke framework is generally no longer applicable. For this, we introduce a Filippov set-valued extension based on \citep{Filippov1988,Quincampoix2021,AubinCellina1984,AubinFrankowska2009}.
In what follows, for $x\in\mathbb R^n$ and $\delta>0$,
$B(x,\delta):=\{y\in\mathbb R^n:\|y-x\|<\delta\}$ denotes the open
ball of radius $\delta$ centred at $x$, $\overline{\operatorname{co}}\,A$
the closed convex hull of a set $A$, $\operatorname{co}\,A$ its convex
hull, and $\mu$ the Lebesgue measure.
Since $\Phi$ and $D\Phi$ are only defined on $\Omega\setminus\Sigma$,
an image $\Phi(A)$, or a set $\{(\Phi(y),D\Phi(y)):y\in A\}$, is taken
over $y\in A\cap(\Omega\setminus\Sigma)$. As $\mu(\Sigma)=0$, this
removal of $\Sigma$ changes nothing in the intersections over null sets
$\mathcal N$ below, which discard $\Sigma$ like any other negligible
set. The motivation for the double intersection is discussed, with
examples, in Appendix~\ref{proof:thm:filippov-piecewise}.

\begin{theorem}
    \label{thm:filippov-piecewise}
    Let $\Phi\in\mathcal S^1(\Omega,\mathbb R^m)$ be a switching operator
    (see Definition~\ref{def:piecewise-switching-operator}) and $x\in\Omega$.
    Define the \emph{Filippov regularization} of $\Phi$ at $x$ and the
    \emph{Filippov extension} of $(\Phi,D\Phi)$ by
    \[
        \mathcal F[\Phi](x):=
        \bigcap_{\mu(\mathcal N)=0}\ \bigcap_{\delta>0}
        \overline{\operatorname{co}}\,\Phi\bigl(B(x,\delta)\setminus\mathcal N\bigr),
        \qquad
        \mathfrak F[\Phi](x):=
        \bigcap_{\mu(\mathcal N)=0}\ \bigcap_{\delta>0}
        \overline{\operatorname{co}}\bigl\{(\Phi(y),D\Phi(y)):y\in B(x,\delta)\setminus\mathcal N\bigr\}.
    \]

    \smallskip\noindent
    (i) For every $x\in\Omega$,
    \[
        \mathcal F[\Phi](x)=\operatorname{co}\bigl\{\Phi^{(k)}(x):k\in J_\Phi(x)\bigr\}
        \subset\mathbb R^m,
        \qquad
        \mathfrak F[\Phi](x)=
        \operatorname{co}\bigl\{\bigl(\Phi^{(k)}(x),D\Phi^{(k)}(x)\bigr):k\in J_\Phi(x)\bigr\}
        \subset\mathbb R^m\times\mathbb R^{m\times n}.
    \]

    \smallskip\noindent
    (ii) If $x\notin\Sigma$, there is a unique $k_0\in K$ with
    $x\in\Omega^{(k_0)}$; then $J_\Phi(x)=\{k_0\}$, and
    \[
        \mathcal F[\Phi](x)=\bigl\{\Phi^{(k_0)}(x)\bigr\}=\bigl\{\Phi(x)\bigr\},
        \qquad
        \mathfrak F[\Phi](x)=\bigl\{\bigl(\Phi(x),D\Phi(x)\bigr)\bigr\}.
    \]
\end{theorem}
Proof of the theorem is provided in
Appendix~\ref{proof:thm:filippov-piecewise}.

In the Filippov regularization formula, the first intersection is taken over the sets $\mathcal N \subset \mathbb R^n$ with Lebesgue measure $\mu(\mathcal N)$ equal to zero.
By (ii), the Filippov regularization coincides with the classical evaluation of $\Phi$ at every regular point.

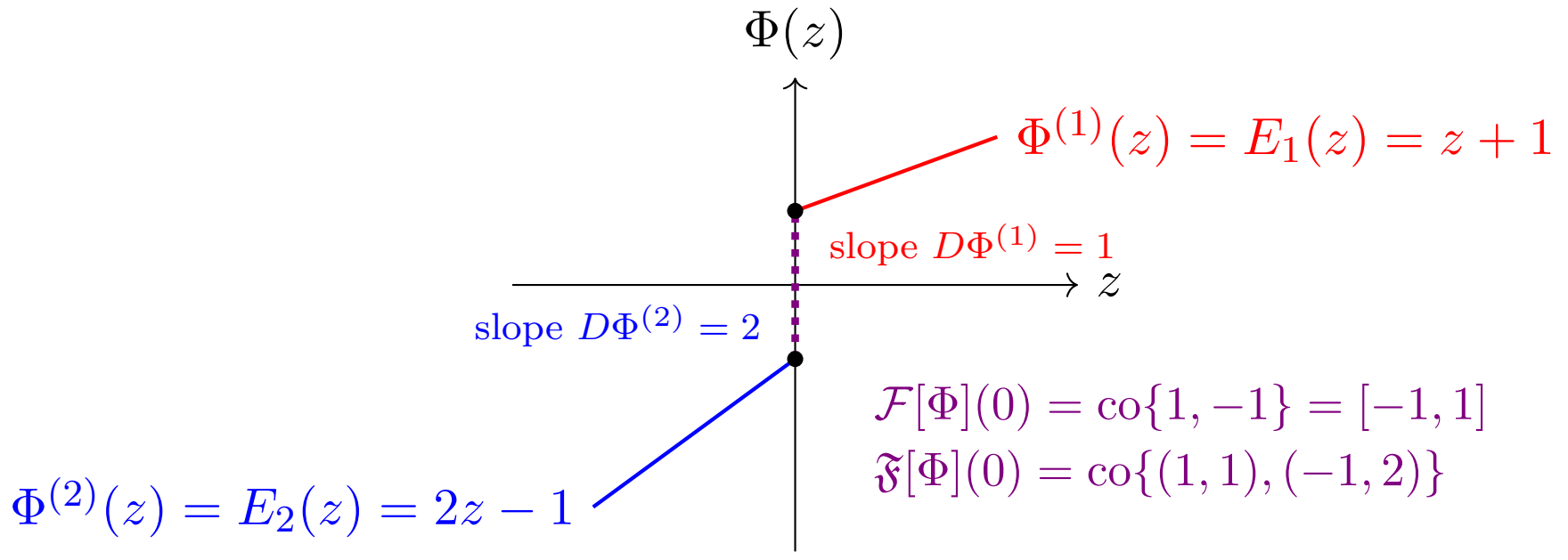
\begin{figure}[!ht]
    \centering
    \begin{tikzpicture}[x=1.5cm,y=0.55cm]
        \draw[->] (-1.4,0) -- (1.4,0) node[right] {$z$};
        \draw[->] (0,-3.6) -- (0,2.8) node[above] {$\Phi(z)$};
        \draw[thick,blue,domain=-1:0,samples=2] plot (\x,{2*\x-1});
        \node[blue,left] at (-1,-3) {$\Phi^{(2)}(z)=E_2(z)=2z-1$};
        \draw[thick,red,domain=0:1,samples=2] plot (\x,{\x+1});
        \node[red,right] at (1,2) {$\Phi^{(1)}(z)=E_1(z)=z+1$};
        \draw[line width=1.6pt,dotted,violet] (0,-1) -- (0,1);
        \filldraw (0,1) circle (1.5pt);
        \filldraw (0,-1) circle (1.5pt);
        \node[right,font=\small,violet,align=left] at (0.3,-2.1)
            {$\mathcal F[\Phi](0)=\operatorname{co}\{1,-1\}=[-1,1]$\\[2pt]
             $\mathfrak F[\Phi](0)=\operatorname{co}\{(1,1),(-1,2)\}$};
        \node[red,right,font=\scriptsize] at (0.08,0.55) {slope $D\Phi^{(1)}=1$};
        \node[blue,left,font=\scriptsize] at (-0.08,-0.55) {slope $D\Phi^{(2)}=2$};
    \end{tikzpicture}
    \caption{The hard-routing MoE example: $\Phi$ jumps from $-1$ to
    $1$ across the routing boundary $z=0$, where $g_1(z)=g_2(z)$. The
    Filippov regularization $\mathcal F[\Phi](0)$ fills this jump with
    the whole segment $[-1,1]$ (dotted, in violet) rather than either
    endpoint alone, and the Filippov extension $\mathfrak F[\Phi](0)$ is
    the segment joining the pairs $(\Phi^{(k)}(0),D\Phi^{(k)}(0))$ of the
    two branches, $(1,1)$ and $(-1,2)$.}
    \label{fig:moe-filippov-example}
\end{figure}

We illustrate Theorem~\ref{thm:filippov-piecewise} on the hard-routing
Mixture-of-Experts block introduced above, with $M=2$ experts on a
scalar input $z\in\mathbb R$: routing scores $g_1(z):=z$,
$g_2(z):=-z$, and affine experts $E_1(z):=z+1$, $E_2(z):=2z-1$, so
that $\Omega^{(1)}=\{z>0\}$, $\Omega^{(2)}=\{z<0\}$, and
$\Sigma=\{0\}$. At the switching point $z=0$, both branches are
active, $J_\Phi(0)=\{1,2\}$, with $\Phi^{(1)}(0)=E_1(0)=1$ and
$\Phi^{(2)}(0)=E_2(0)=-1$ actually different values -- unlike a ReLU
kink, where every branch agrees at the switching point itself, this
is exactly the network-map discontinuity that motivates the Filippov
construction. Theorem~\ref{thm:filippov-piecewise} gives
\[
\mathcal F[\Phi](0)
=\operatorname{co}\bigl\{\Phi^{(1)}(0),\Phi^{(2)}(0)\bigr\}
=\operatorname{co}\{1,-1\}
=[-1,1],
\]
an interval of admissible output values at the routing
boundary, rather than the single value either expert alone would
report (Figure~\ref{fig:moe-filippov-example}). Including the branch
derivatives $D\Phi^{(1)}(0)=E_1'(0)=1$ and $D\Phi^{(2)}(0)=E_2'(0)=2$,
the Filippov extension is likewise a non-degenerate segment,
\[
\mathfrak F[\Phi](0)=\operatorname{co}\bigl\{(1,1),(-1,2)\bigr\},
\]
with both coordinates varying across branches -- to be contrasted
with the ReLU-kink example of Figure~\ref{fig:DAG example}, where the
value coordinate collapses to a point and only the Jacobian coordinate
remains set-valued.

To the best of our knowledge, Filippov regularization has not been applied to the discontinuous hard-routing maps of feed-forward Mixture-of-Experts architectures. Earlier work addresses related issues through other constructions. \citet{Jacobs1991} interpret the softmax outputs of the gating network as the probabilities of a stochastic one-out-of-$n$ selector, and train the system through a smooth objective of the mixture. \citet{Shazeer2017} note that sparse gating \emph{``creates some theoretically scary discontinuities''}, without reporting practical difficulties attributable to them; their noisy gating also gives a differentiable estimate of the load of each expert, used for load balancing. \citet{Fedus2022} keep the softmax probability of the selected expert as a multiplicative gate: away from the routing boundaries, backpropagation then computes the classical derivative of the active branch, but at a boundary where the branch values differ, these branch derivatives are neither a classical nor a Clarke gradient of the composite map. The Filippov construction below addresses these boundaries.

\subsection{Directed Acyclic graph for networks}
\label{sec:dag-networks}

\begin{figure}[tbp]
    \centering
    \begin{tikzpicture}[
        x=1cm,
        y=1cm,
        scale=0.75,
        transform shape,
        >=stealth,
        neuron/.style={circle, draw, minimum size=0.8cm, thick, inner sep=0pt, font=\small},
        attn/.style={->, orange!80!black},
        res/.style={->, dashed, gray!70!black},
        every node/.style={font=\small}
    ]

    \def\ya{0}    %
    \def\yb{1.8}  %
    \def\yc{4.3}  %
    \def\yd{5.7}  %
    \def\ye{7.1}  %
    \def\yf{8.5}  %

    \foreach \t/\c in {1/0, 2/4.2, 3/8.4}{
        \node[neuron, blue]                  (x\t) at (\c,\ya) {$i_{\t}$};
        \node[neuron, green!45!black]        (q\t) at (\c-1.05,\yb) {$q_{\t}$};
        \node[neuron, green!45!black]        (k\t) at (\c,\yb) {$k_{\t}$};
        \node[neuron, green!45!black]        (v\t) at (\c+1.05,\yb) {$v_{\t}$};
        \node[neuron, orange!80!black]       (a\t) at (\c,\yc) {$a_{\t}$};
        \node[neuron, violet]                (h\t) at (\c,\yd) {$h_{\t}$};
        \node[neuron, violet, double]        (f\t) at (\c,\ye) {$f_{\t}$};
        \node[neuron, red]                   (y\t) at (\c,\yf) {$o_{\t}$};

        \draw[->, blue] (x\t) -- (q\t);
        \draw[->, blue] (x\t) -- (k\t);
        \draw[->, blue] (x\t) -- (v\t);
        \draw[attn] (q\t.north) -- (a\t.south);
        \draw[->, orange!80!black] (a\t) -- (h\t);
        \draw[->, violet] (h\t) -- (f\t);
        \draw[->, violet] (f\t) -- (y\t);
        \draw[res] (x\t.west) .. controls +(-1.9,1.2) and +(-1.9,-1.2) .. (h\t.west);
        \draw[res] (h\t.east) .. controls +(0.9,0.5) and +(0.9,-0.5) .. (y\t.east);
    }

    \foreach \t in {1,2,3}{
        \foreach \s in {1,...,\t}{
            \draw[attn] (k\s.north) -- (a\t.south);
            \draw[attn] (v\s.north) -- (a\t.south);
        }
    }

    \node[anchor=east] at (-2.3,\ya) {\textbf{tokens}};
    \node[anchor=east] at (-2.3,\yb) {\textbf{Q, K, V}};
    \node[anchor=east] at (-2.3,\yc) {\textbf{attention}};
    \node[anchor=east] at (-2.3,\yd) {\textbf{residual}};
    \node[anchor=east] at (-2.3,\ye) {\textbf{FFN / MoE}};
    \node[anchor=east] at (-2.3,\yf) {\textbf{outputs}};
    \end{tikzpicture}

    \caption{A causal Transformer block on three tokens, drawn as a
    directed acyclic graph with input nodes $i_t$ (the tokens) and output
    nodes $o_t$. Each node stands for a group of neurons (one
    $d$-dimensional vector per token); expanding it coordinate by
    coordinate gives a DAG in the sense of Definition~\ref{def:DAG}. The
    attention node $a_t$ receives the query $q_t$ and the keys and values
    $k_s,v_s$ of every earlier token $s\le t$ (solid orange edges), so the
    in-degree grows with the context length; dashed edges are residual
    connections skipping one or several stages, which prevent the graph
    from being layered (Definition~\ref{def:layered-DAG}). The
    double-circled node $f_t$ is the feed-forward sub-layer, which becomes
    a switching operator in the sense of
    Definition~\ref{def:piecewise-switching-operator} when it is a hard-routed
    Mixture-of-Experts. CNN, ResNet, DenseNet, Inception and U-Net
    architectures give DAGs of the same kind.}
    \label{fig:scale_free_dag}
\end{figure}
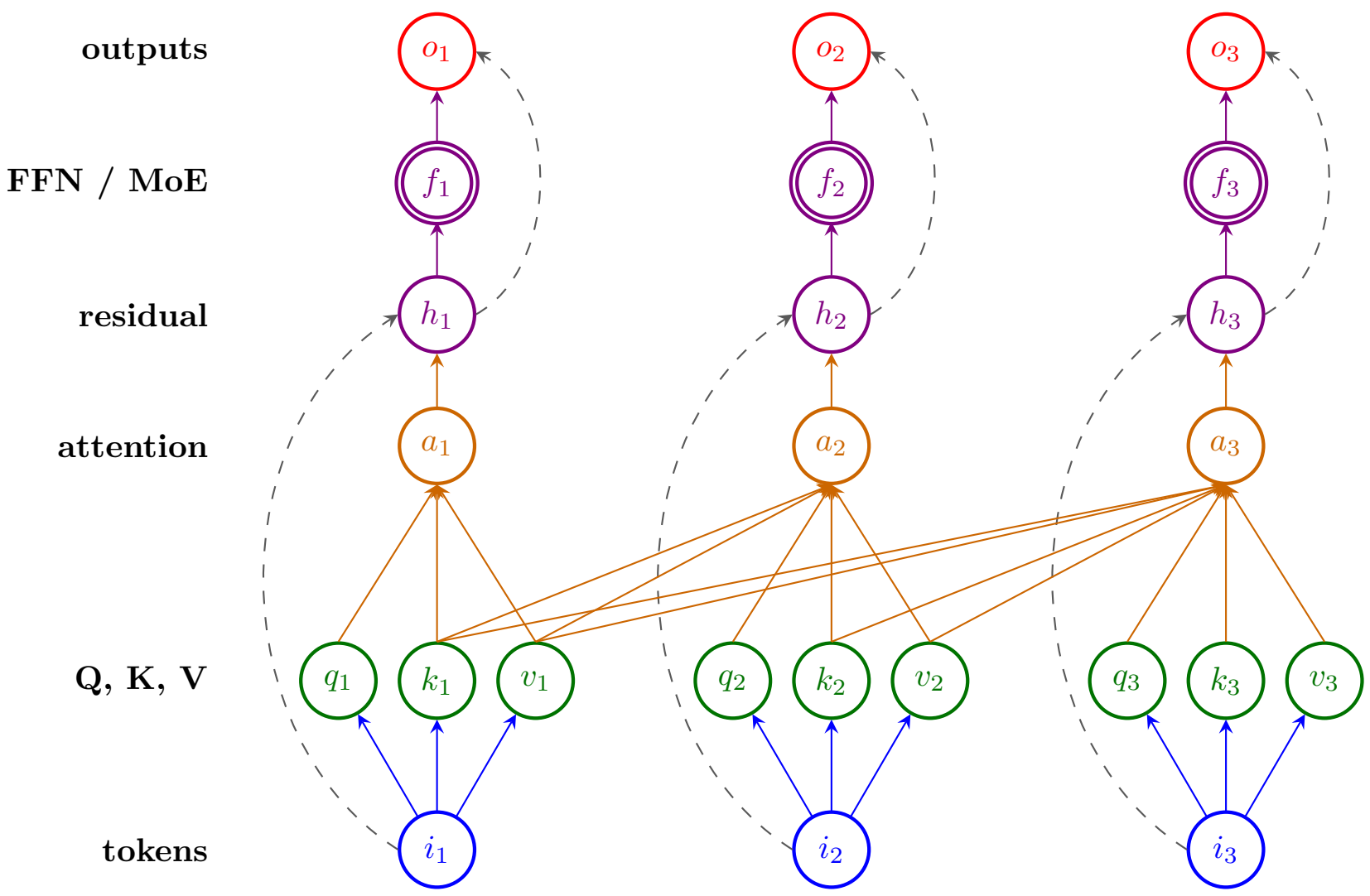

Let us consider a directed acyclic graph (DAG). They are used in many feed-forward neural network architectures (see Figure \ref{fig:scale_free_dag}). We then propose a new definition of feed-forward neural networks based on DAGs, following \citet{DeVore2021,DeVore2022} and adapted to our general Filippov framework.

\begin{definition}
    \label{def:DAG}
    Let $\mathcal G=(\mathcal V,\mathcal E)$ be a finite directed acyclic graph.
    The set $\mathcal V$ is the set of nodes and $\mathcal E$ is the set of
    directed edges. We denote by
    $
    \mathcal I := \{v\in\mathcal V:\operatorname{In}(v)=\varnothing\}
    $
    the set of input nodes, and by
    $
    \mathcal O := \{v\in\mathcal V:\operatorname{Out}(v)=\varnothing\}
    $
    the set of output nodes where
    $
    \operatorname{In}(v):=\{u\in\mathcal V:(u,v)\in\mathcal E\},
    $ and
    $\operatorname{Out}(v):=\{u\in\mathcal V:(v,u)\in\mathcal E\}.
    $
    The following properties define a feed-forward Filippov neural network
    $\mathcal N_{\mathcal F}(\mathcal G)$ associated to $\mathcal G$:
    \begin{itemize}
    \item To every edge $e=(u,v)\in\mathcal E$ is associated a scalar weight
    $w_{u,v}\in\mathbb R$.
    \item For any node $i\in\mathcal I$, we associate a scalar value. For
    some input nodes, we prescribe the value to be fixed and equal to $+1$.
    This subset $\mathcal B\subset\mathcal I$ is called the bias set.
    \item For every non-input node $v\in\mathcal V\setminus\mathcal I$, we
    assume that:
        \begin{itemize}
            \item A local operator
                $\Phi_v\in\mathcal S^1\bigl(\mathbb R^{|\operatorname{In}(v)|}\bigr)$
                is associated (see Definition~\ref{def:piecewise-switching-operator}),
                with switching region $\Sigma_v$, branches
                $(\Phi_v^{(k)})_{k=1}^{m_v}$, regions
                $\Omega_v^{(1)},\ldots,\Omega_v^{(m_v)}$, and active index set
                $J_v:=J_{\Phi_v}$.
            \item For any input vector
            $x=(x_i)_{i\in\mathcal I\setminus\mathcal B}\in\mathbb R^d$ with
            $d:=|\mathcal I\setminus\mathcal B|$, 
            \begin{itemize}
                \item
                We denote the value of any input node by $X_i(x)=\{x_i\}$
                if $i\in\mathcal I\setminus
                \mathcal B$, and by $X_i(x)=\{+1\}$ for any $i\in
                \mathcal B$.

                \item We define $Z_v(x):=\displaystyle\prod_{u\in\operatorname{In}(v)}z_{u,v}
                \subset\mathbb R^{|\operatorname{In}(v)|}$,
                with $z_{u,v} := w_{u,v}\times X_u(x)$ for any $u\in \operatorname{In}(v)$.

                \item We then define recursively the
                value $X_v(x)$ of each node $v\in\mathcal V\setminus\mathcal I$ by the
                propagated Filippov set of
                $\Phi_v$ at $Z_v(x)$, extending the pointwise Filippov
                regularization $\mathcal F[\Phi_v](\zeta)$ of
                Theorem~\ref{thm:filippov-piecewise} to the whole box $Z_v(x)$ by
                \[
                    X_v(x)=\mathcal F[\Phi_v]\bigl(Z_v(x)\bigr)
                    :=
                    \underbrace{\bigcup_{k\in J_v(Z_v(x))}
                    \bigl\{\Phi_v^{(k)}(\zeta):\zeta\in Z_v(x)\cap \Omega_v^{(k)}\bigr\}}_{\text{regular points}}
                    \ \cup\
                    \underbrace{\bigcup_{\zeta\in Z_v(x)\cap\Sigma_v}
                    \mathcal F[\Phi_v](\zeta)}_{\text{switching
                    points}},
                \]
                where $J_v(Z_v(x)):=\bigl\{k:\Omega_v^{(k)}\cap Z_v(x)\neq
                \varnothing\bigr\}$ is the set of regions met by the box
                $Z_v(x)$, extending the pointwise active index set
                $J_v(\zeta)$ of Definition~\ref{def:piecewise-switching-operator}.
                Each term of the first union is the image, under the single
                $C^1$ branch $\Phi_v^{(k)}$, of the part of the box lying in
                $\Omega_v^{(k)}$; since $\Phi_v=\Phi_v^{(k)}$ on $\Omega_v^{(k)}$,
                this union is the set $\{\Phi_v(\zeta):\zeta\in
                Z_v(x)\setminus\Sigma_v\}$ of the classical values at the
                regular points of the box, while the second union collects the
                local convex hulls at its switching points.
            \end{itemize}
        \end{itemize}
    \item We denote by $\mathcal H:=\mathcal V\setminus(\mathcal I\cup
    \mathcal O)$ the hidden nodes.
    \end{itemize}
\end{definition}

The graph $\mathcal G$ fixes the topology of
computation: since $\mathcal G$ is directed and acyclic, a topological
order on $\mathcal V$ always exists, so the recursive definition of
$Z_v(x)$ and $X_v(x)$ is well founded and terminates after finitely many
steps -- there is no risk of circular dependency, unlike in recurrent
architectures. {The weights $w_{u,v}$} act exactly as in a classical
network: they rescale the (set-valued) output of a node before it
reaches the next one. Since scalar multiplication by $w_{u,v}$ maps
convex compact sets to convex compact sets, this operation is
compatible with the Filippov formalism at every step.
The bias set $\mathcal B$ plays the same role as in a
classical affine layer: fixing $X_i(x)=\{+1\}$ for $i\in\mathcal B$
allows constant (bias) terms to be absorbed into the weighted-sum
structure of $Z_v(x)$ without special-casing the definition.

As we can see, each node is associated to one neuron.
The DAGs admit topological sorting and no cycles \citep{Bondy1976}.
This allows us to deduce relationships in a logical order.
Consequently, for any given feed-forward neural network,
we can compute the list of all paths from one input node of $\mathcal{I}$
to any output node of $\mathcal{O}$ and conversely.
Moreover, for any $e=(u,v) \in \mathcal{E}$ the parameters $w_e$ are then trainable.
This definition encompasses virtually all network architectures used
in practice -- fully connected, convolutional, residual, U-Net, U$^2$-Net,
and Transformer networks -- with input, bias, hidden and output
units corresponding to nodes of $\mathcal G$, directed edges representing
dependencies between nodes, and the (possibly multi-valued) local operators $\Phi_v$
representing the computation performed at each node.

The layer types of a standard architecture are instances of
Definition~\ref{def:DAG} obtained by a choice of fan-in and of local
operator: a convolution reads the weights of all its edges in one kernel $\theta$, $w_{u,v}=\theta_{c(u),c(v),p(u)-p(v)}$ for $u$ in the receptive field of $v$ ($c$ being the channel and $p$ the spatial position of a unit); a max-pooling unit is one node carrying the
max-of-$|\operatorname{In}(v)|$ switching operator of
Section~\ref{sec:filippov-regularization}, whose ties are the only place
where branches compete; and a softmax over $n$ scores is $n$ smooth nodes $v_1,\dots,v_n$, whose
switching regions are empty, $\Sigma_{v_j}=\varnothing$: it never
switches, unlike the hard-routing $\operatorname{argmax}$ that replaces it
in a Mixture-of-Experts, which switches where two scores tie. The local operator is where
all the nonsmooth, possibly discontinuous behavior is concentrated --
smooth activations, ReLU, sign-type relays and quantization are all
finitely many $C^1$ branches -- and the propagated value
$\mathcal F[\Phi_v](Z_v(x))$ is the union of the \emph{pointwise}
Filippov values over $\zeta\in Z_v(x)$, not a global convexification of
the incoming box, which is what keeps it nonempty and compact.

We now single out the special case where the nodes organize into a
sequence of layers, each feeding only the next.

\begin{definition}
    \label{def:layered-DAG}
    A feed-forward Filippov neural network $\mathcal N_{\mathcal F}(\mathcal G)$
    (see Definition~\ref{def:DAG}) is \emph{layered} if there
    exists a rank function $\ell:\mathcal V\to\{0,1,\dots,K\}$ with
    $\ell(i)=0$ for every non-bias input $i\in\mathcal I\setminus
    \mathcal B$, $\ell(o)=K$ for every output $o\in\mathcal O$, and
    $\ell(v)=\ell(u)+1$ for every edge $(u,v)\in\mathcal E$. The level
    sets of $\ell$ are the successive \emph{layers} of the network.
\end{definition}

The classical sequential architectures underlying \citet{Bach2015}'s
Layer-wise Relevance Propagation -- and, by inheritance, the acronyms LRP
and SRP used throughout this paper -- are exactly the layered case of
Definition~\ref{def:layered-DAG}. The active graph $G(x)$ of a general
feed-forward Filippov network need not be layered: skip connections,
hard-routed Mixture-of-Experts, and attention-induced edges all produce
active graphs whose nodes cannot be assigned a single consistent rank.
Every result of this paper is stated for the general, not necessarily
layered, DAG $\mathcal G$; the layered case of Definition~\ref{def:layered-DAG}
is recovered as the special instance considered in the classical literature.

\subsection{Forward Propagation and the Active Graph}
\label{sec:forward-active}

The recursion of Definition~\ref{def:DAG} propagates, at every switching
node, the whole local convex hull of the active branches, so that node
values are in general sets. An ordinary forward pass, or an
automatic-differentiation library, computes none of these sets: it
commits at each switching node to one active branch and propagates a
single real number. The following definition formalizes this choice as
a \emph{branch selection}, and characterizes, through the active index
sets $J_v$, the selections that are consistent with the values they
themselves produce.

\begin{definition}
    \label{def:branch}
    Let $\mathcal N_{\mathcal F}(\mathcal G)$ be a feed-forward Filippov
    neural network (see Definition~\ref{def:DAG}) on a finite directed acyclic
    graph $\mathcal G=(\mathcal V,\mathcal E)$, and let $x\in\mathbb R^d$,
    $d:=|\mathcal I\setminus\mathcal B|$. A \emph{branch selection} is a
    family
    \[
        \kappa=(\kappa_v)_{v\in\mathcal V\setminus\mathcal I}\ \in\
        \prod_{v\in\mathcal V\setminus\mathcal I}\{1,\ldots,m_v\},
    \]
    assigning to every non-input node $v$ one of the $m_v$ branches of
    $\Phi_v\in\mathcal S^1(\Omega_v)$. It determines the recursion of
    Definition~\ref{def:DAG} in which the Filippov value
    $\mathcal F[\Phi_v]$ is replaced, at every node $v$, by the single
    branch $\Phi_v^{(\kappa_v)}$: forward along a topological order of
    $\mathcal G$,
    \begin{equation}
        \label{eq:branch-forward}
        X_i^{(\kappa)}(x):=X_i(x)\ \ (i\in\mathcal I),
        \qquad
        Z_v^{(\kappa)}(x):=\prod_{a\in\operatorname{In}(v)}w_{a,v}\,
        X_a^{(\kappa)}(x),
        \qquad
        X_v^{(\kappa)}(x):=\Phi_v^{(\kappa_v)}\bigl(Z_v^{(\kappa)}(x)\bigr)
        \ \ (v\in\mathcal V\setminus\mathcal I).
    \end{equation}
    Since $X_i(x)=\{x_i\}$ or $\{+1\}$, every $Z_v^{(\kappa)}(x)$ and
    $X_v^{(\kappa)}(x)$ is a singleton, whose unique element we denote by
    $z_v^{(\kappa)}(x)\in\mathbb R^{|\operatorname{In}(v)|}$ and
    $x_v^{(\kappa)}(x)\in\mathbb R$ respectively. The selection $\kappa$ is
    \emph{admissible at $x$} if it picks, at every node, a branch that is
    active at the pre-activation it has itself produced,
    \begin{equation}
        \label{eq:branch-admissible}
        \kappa_v\ \in\ J_v\bigl(z_v^{(\kappa)}(x)\bigr)
        \qquad\text{for every }v\in\mathcal V\setminus\mathcal I,
    \end{equation}
    and we write
    \[
        \mathcal M(x):=\Bigl\{\kappa\ :\ \kappa\in
        \prod_{v\in\mathcal V\setminus\mathcal I}
        J_v\bigl(z_v^{(\kappa)}(x)\bigr)\Bigr\}
    \]
    for the set of branch selections admissible at $x$. For
    $\kappa\in\mathcal M(x)$ and $o\in\mathcal O$, the scalar
    $x_o^{(\kappa)}(x)$ is the \emph{output value along $\kappa$}.
\end{definition}

A branch selection is attached to the network, not to the input: like
the weights $w_{u,v}$, $\kappa$ is given before any $x$, and it is the
set $\mathcal M(x)$ of admissible ones that varies with $x$.
Condition \eqref{eq:branch-admissible} is equivalent to
$z_v^{(\kappa)}(x)\in\overline{\Omega_v^{(\kappa_v)}}$, the condition
under which the evaluation in \eqref{eq:branch-forward} is meaningful at
all: an admissible selection is one with all node values well
defined. Along a topological order it is a recipe, not a test -- once the predecessors of $v$ carry values,
$J_v(z_v^{(\kappa)}(x))$ is nonempty, so an admissible $\kappa_v$ is
available and $\mathcal M(x)\neq\varnothing$
(Theorem~\ref{thm:ffnn-output}(ii)) -- and it carries information only
at a switching node.

Two superscripts must be distinguished: $(\kappa)$ is the global
selection, $(\kappa_v)$ the local branch index it prescribes at $v$.
Since that branch changes the values downstream, the product of
Definition~\ref{def:branch} has $\kappa$-dependent factors:
$\mathcal M(x)$ is a set of fixed points, not a product of fixed sets,
and it is \emph{fibered} -- along a topological order $v_1,v_2,\ldots$,
each $\kappa_{v_j}$ ranges over $J_{v_j}$ at the pre-activation produced
upstream, read at the point $z_{v_j}^{(\kappa)}(x)$ and not on the box
$Z_{v_j}(x)$, whose index set is empty at a pure switching point. Hence
$\mathcal M(x)\subseteq\prod_v\bigcup_{\zeta\in Z_v(x)}J_v(\zeta)$, in
general strictly (Appendix~\ref{thm:ffnn-output_example}), with equality
when the branches agree at the switching points, as for ReLU, maxout or
max-pooling.

Alongside a branch selection, it is useful to isolate the nodes and
edges of $\mathcal G$ that actually carry a nonzero contribution for a
given input, since a network typically activates only a subgraph of
the full DAG.

\begin{definition}
    \label{def:active-graph}
    Let $\mathcal N_{\mathcal F}(\mathcal G)$ be a feed-forward Filippov neural network, then for any input $x\in \mathbb R^d$ ($d:=|\mathcal I\setminus\mathcal B|$), we then define $G(x)=(V(x),E(x))$ the active graph of $\mathcal{G}$ where $V(x)= \{v\in\mathcal H \cup \mathcal I:\; X_v(x)\neq \{0\}\}  \cup \mathcal O$ is the set of active nodes and $E(x)=\{(u,v)\in\mathcal E:\ u,v\in V(x),\ z_{u,v} \neq \{0\}\}$  is the set of active edges where
    $z_{u,v}=w_{u,v}X_u(x)$ is the weighted contribution of neuron $u$ to neuron $v$ for any edge $(u,v) \in E(x)$. We write $\mathcal I(x):=\mathcal I\cap V(x)$ for the set of active input nodes.
\end{definition}

The active graph $G(x)$ is the graph of the nonzero forward contributions: a node of value zero leaves it with its edges, even if, at a linear node, the derivative of the output with respect to its value does not vanish. No relevance being attached to such a node, zero-valued nodes create no non-generic attribution ambiguity.

Switching nodes are exceptional for a \emph{generic} input, but nothing
said so far prevents a node from being switching for \emph{every} input
at once; this is what the non-degeneracy hypothesis of
Theorem~\ref{thm:ffnn-output}(iii) below excludes.

We now show that forward propagation is well-posed.

\begin{theorem}
    \label{thm:ffnn-output}
    Let $\mathcal N_{\mathcal F}(\mathcal G)$ be a feed-forward Filippov
    neural network (see Definition~\ref{def:DAG}) and $x\in\mathbb R^d$.

    \smallskip\noindent
    (i) Every node value is the nonempty compact set
    $
        X_v(x)=\bigcup_{\zeta\in Z_v(x)}
        \bigl[\min_{k\in J_v(\zeta)}\Phi_v^{(k)}(\zeta),\,
        \max_{k\in J_v(\zeta)}\Phi_v^{(k)}(\zeta)\bigr]
    $
    ($v\in\mathcal V\setminus\mathcal I$), and the network output is the box
    $\mathcal S_{\mathcal N}(x)=\prod_{o\in\mathcal O}X_o(x)\subset
    \mathbb R^{|\mathcal O|}$.

    \smallskip\noindent
    (ii) $\mathcal M(x)$ is nonempty and finite, with
    $x_v^{(\kappa)}(x)\in X_v(x)$ at every node $v\in\mathcal V$ and every
    $\kappa\in\mathcal M(x)$. For $\kappa\in\mathcal M(x)$,
    $v\in\mathcal V$ and $o\in\mathcal O$, the output
    $x_o^{(\kappa)}$ of \eqref{eq:branch-forward} is continuously
    differentiable with respect to the value $x_v^{(\kappa)}$ of $v$, and
    \begin{equation}
        \label{eq:branch-derivative}
        g^{(\kappa)}_{v,o}(x):=\frac{\partial x_o^{(\kappa)}}
        {\partial x_v^{(\kappa)}}(x)
        =\sum_{\mathbf p\in\mathcal P(v,o)}\ \prod_{(a,b)\in\mathbf p}
        w_{a,b}\,\Bigl[D\Phi_b^{(\kappa_b)}\bigl(z_b^{(\kappa)}(x)\bigr)
        \Bigr]_a ,
    \end{equation}
    where $\mathcal P(v,o)$ is the set of directed paths from $v$ to $o$.
    Hence the \emph{branch-admissible set of value-gradient pairs}
    \begin{equation}
        \label{eq:admissible-value-gradient}
        \mathfrak F_{v,o}(x):=\operatorname{co}
        \Bigl\{\bigl(x_o^{(\kappa)}(x),g^{(\kappa)}_{v,o}(x)\bigr):
        \kappa\in\mathcal M(x)\Bigr\}
    \end{equation}
    is a nonempty convex compact subset of $\mathbb R^2$.

    \smallskip\noindent
    (iii) Let
    $\mathcal D:=\{x\in\mathbb R^d:\ Z_v(x)\cap\Sigma_v\neq\varnothing
    \text{ for some } v\in\mathcal V\setminus\mathcal I\}$ be the
    exceptional set. If $x\notin\mathcal D$, then every $X_v(x)$ is a
    singleton, $\mathcal M(x)=\{\kappa\}$, and
    $\mathfrak F_{v,o}(x)=\bigl\{\bigl(x_o(x),\,\partial x_o/\partial x_v(x)\bigr)\bigr\}$
    with $\partial x_o/\partial x_v(x)=g^{(\kappa)}_{v,o}(x)$. If moreover
    the network is \emph{non-degenerate}, in the sense that
    \[
        \mu\Bigl(\bigl\{x\in\mathbb R^d\ :\ z_v^{(\kappa)}(x)\in\Sigma_v
        \ \text{ for some }\kappa\in\mathcal M(x)\bigr\}\Bigr)=0
        \qquad\text{for every } v\in\mathcal V\setminus\mathcal I ,
    \]
    then $\mu(\mathcal D)=0$.

    \smallskip\noindent
    (iv) If the branches agree on the switching sets
    ($\Phi_v^{(k)}(\zeta)=\Phi_v^{(k')}(\zeta)$ for every
    $\zeta\in\Sigma_v$ and $k,k'\in J_v(\zeta)$, as for ReLU, maxout or
    max-pooling), then $x\mapsto x_o(x)$ is single-valued
    and locally Lipschitz and $x_o^{(\kappa)}(x)=x_o(x)$ for every
    $\kappa\in\mathcal M(x)$. If $x\notin\mathcal D$, then
    $\partial^\circ x_o(x)=\bigl\{\bigl(g^{(\kappa)}_{i,o}(x)\bigr)_{i\in
    \mathcal I\setminus\mathcal B}\bigr\}$ for the unique
    $\kappa\in\mathcal M(x)$, and
    \[
        \partial^\circ x_o(x)\ \subseteq\
        \operatorname{co}\Bigl\{\bigl(g^{(\kappa)}_{i,o}(x)\bigr)_{i\in
        \mathcal I\setminus\mathcal B}:\kappa\in\mathcal M(x)\Bigr\}
        \qquad\text{for every }x\in\mathbb R^d .
    \]
\end{theorem}
Proof and a worked example are in
Appendices~\ref{thm:ffnn-output_proof} and~\ref{thm:ffnn-output_example}.

Non-degeneracy constrains preimages, not images: each $\Sigma_v$ is
null, but a pre-activation may pull it back to a large set. It is not
implied by nonzero weights: at the second node of
$x\mapsto\operatorname{ReLU}(\operatorname{ReLU}(x))$, the pre-activation
vanishes on $(-\infty,0)$, and a max-pooling window that follows ReLU
nodes is a tie on an open set as soon as all its values vanish, which
puts a whole neighborhood of a photograph in $\mathcal D$ for a network
such as VGG-16. Sections \ref{sec:LRP}--\ref{sec:transformers} do not
need it: the branches of these operators agree on the switching sets, so
that the node values do not branch, by (iv), at every $x$.

The derivative in \eqref{eq:branch-derivative} is the
computational-graph one: the value of $v$ becomes a free variable, its
descendants are recomputed with the branches prescribed by $\kappa$, the
other nodes frozen -- well defined since each $\Phi_u^{(\kappa_u)}$ is
$C^1$ near $z_u^{(\kappa)}(x)$ even in $\Sigma_u$. At an input or bias
node it is the ordinary partial derivative with respect to the
prescribed value. At a regular point we
write $x_v$ for the unique element of $X_v(x)$, the convention used
below; Sections \ref{sec:LRP}--\ref{sec:transformers} are stated at the
inputs where the node values do not branch. An
automatic-differentiation pass resolving each switching node
consistently with its propagated values returns a pair of
$\mathfrak F_{v,o}(x)$, and why $\mathcal M(x)$ is global rather than
per path: $z_u^{(\kappa)}(x)$ depends on all ancestors of $u$.

At $x\in\mathcal D$ it is conservative both ways: its value
coordinate may miss part of $X_o(x)$
(Appendix~\ref{thm:ffnn-output_example}), and the inclusion of (iv) may
be strict -- for $o=x_{h_1}-x_{h_2}$ with two ReLU nodes fed by one
input, $\partial^\circ x_o(0)=\{0\}$ while
independent choices at the common kink give $[-1,1]$. We keep
$\mathcal M(x)$: what a deterministic implementation can choose. The
boxes, for their part, are outer enclosures: their products forget the
correlations between the descendants of a node. For $h=\operatorname{sign}(x)$
and $o=h-h$ through two relays, $X_o(0)=[-2,2]$, while the composite map
vanishes identically.

\begin{figure}[H]
    \centering
    \begin{tikzpicture}[
        scale=0.74,
        transform shape,
        x=3.9cm,
        y=2.6cm,
        >=stealth,
        neuron/.style={circle,draw,font=\scriptsize,inner sep=1pt,minimum size=1.5cm,thick},
        wl/.style={inner sep=1.5pt,font=\scriptsize,fill=white,fill opacity=0.85,
                   text opacity=1,rounded corners=1pt},
        every node/.style={font=\small}
    ]

    \node[neuron,blue] (x1) at (0,1.8)
        {\shortstack{$X_{i_1}=\{-2\}$\\[-1pt]{\color{blue}$g^{(\kappa)}_{i_1,o_1}\!\in\!\{-1,0\}$}}};
    \node[neuron,green!30!black!70] (x2) at (0,0.6)
        {\shortstack{$X_{i_2}=\{1\}$\\[-1pt]{\color{green!30!black!70}$g^{(\kappa)}_{i_2,o_1}\!\in\!\{2,4\}$}}};
    \node[neuron] (b1) at (0,-0.6) {\shortstack{$X_{b_1}=\{+1\}$\\[-1pt]$g_{b_1,o_1}=2$}};

    \node[neuron,red,thick] (h1) at (2,1.15)
        {\shortstack{$X_{h_1}=\{0\}$\\[-1pt]{\color{red}$D\Phi^{(k)}_{h_1}\!\in\!\{0,1\}$}\\[-1pt]{\tiny(kink, inactive)}}};
    \node[neuron,teal!70!black] (h2) at (2,-0.05) {\shortstack{$X_{h_2}=\{4\}$\\[-1pt]$D\Phi_{h_2}=1$}};
    \node[neuron] (b2) at (2,-1.25) {\shortstack{$X_{b_2}=\{+1\}$\\[-1pt]$g_{b_2,o_1}=0$}};

    \node[neuron,olive] (y1) at (4,1.8) {\shortstack{$X_{o_1}=\{6\}$\\[-1pt]$D\Phi_{o_1}=1$}};
    \node[neuron,olive] (y2) at (4,0.6) {\shortstack{$X_{o_2}=\{7\}$\\[-1pt]$D\Phi_{o_2}=1$}};

    \node at (0,2.75) {\textbf{Input}};
    \node at (2,2.75) {\textbf{Hidden ReLU}};
    \node at (4,2.75) {\textbf{Output}};

    \draw[->,blue] (x1) -- node[wl,sloped,above,pos=0.52,text=blue] {$z_{i_1,h_1}\!=\!\{-2\}$} (h1);
    \draw[->,green!30!black!70] (x2) -- node[wl,sloped,above,pos=0.74,text=green!30!black!70] {$z_{i_2,h_1}\!=\!\{2\}$} (h1);
    \draw[->,blue] (x1) -- node[wl,sloped,below,pos=0.26,text=blue] {$z_{i_1,h_2}\!=\!\{2\}$} (h2);
    \draw[->,green!30!black!70] (x2) -- node[wl,sloped,below,pos=0.52,text=green!30!black!70] {$z_{i_2,h_2}\!=\!\{1\}$} (h2);
    \draw[->,dashed] (b1) -- node[wl,sloped,above,pos=0.17] {$z_{b_1,h_1}\!=\!\{0\}$} (h1);
    \draw[->] (b1) -- node[wl,sloped,below,pos=0.34] {$z_{b_1,h_2}\!=\!\{1\}$} (h2);

    \draw[->,red] (h1) -- node[wl,sloped,above,pos=0.5,text=red] {$z_{h_1,o_1}\!=\!\{0\}$} (y1);
    \draw[->,teal!70!black] (h2) -- node[wl,sloped,above,pos=0.2,text=teal!70!black] {$z_{h_2,o_1}\!=\!\{8\}$} (y1);
    \draw[->,red] (h1) -- node[wl,sloped,below,pos=0.25,text=red] {$z_{h_1,o_2}\!=\!\{0\}$} (y2);
    \draw[->,teal!70!black] (h2) -- node[wl,sloped,below,pos=0.44,text=teal!70!black] {$z_{h_2,o_2}\!=\!\{8\}$} (y2);
    \draw[->,dashed] (b2) -- node[wl,sloped,above,pos=0.72] {$z_{b_2,o_1}\!=\!\{0\}$} (y1);
    \draw[->] (b2) -- node[wl,sloped,below,pos=0.16] {$z_{b_2,o_2}\!=\!\{-1\}$} (y2);

    \draw[->,blue] (x1) -- node[wl,above,pos=0.5,text=blue] {$z_{i_1,o_1}\!=\!\{-2\}$} (y1);

    \end{tikzpicture}
    \caption{The running example at the switching point
    $x=(-2,1)\in\mathcal D$.}
    \label{fig:DAG example}
\end{figure}
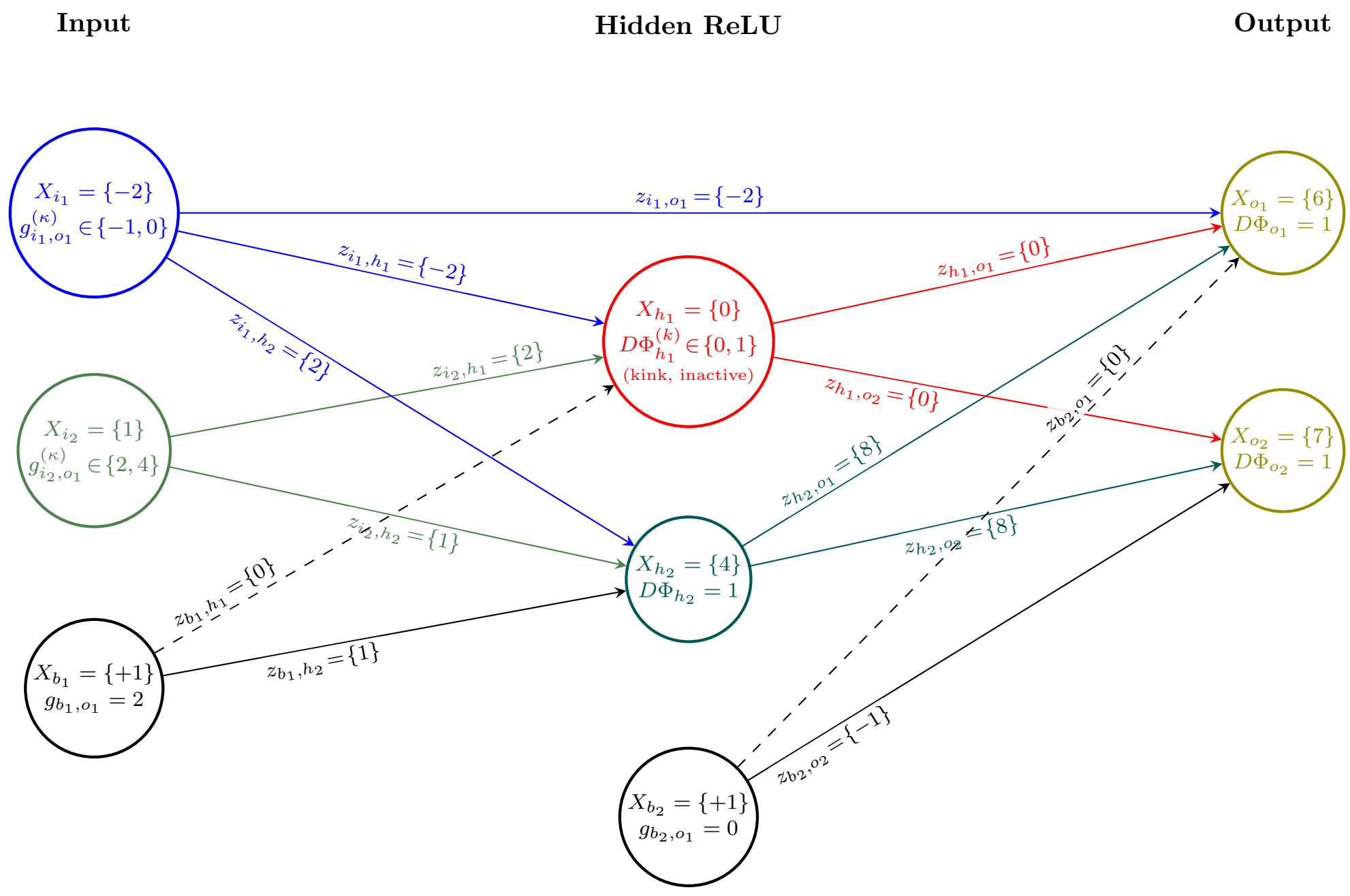

The example of Figure~\ref{fig:DAG example} illustrates
Definition~\ref{def:DAG} at $x=(-2,1)$: ReLU hidden nodes $h_1,h_2$,
linear outputs $o_1,o_2$, a skip edge $i_1\to o_1$ and bias nodes
$b_1,b_2$ of value $+1$, with weights
$w_{i_2,h_1}=w_{h_2,o_1}=w_{h_2,o_2}=2$,
$w_{i_1,h_2}=w_{h_1,o_2}=w_{b_2,o_2}=-1$, zero on the two dashed edges
and $1$ on the others. Each arrow carries its contribution
$z_{u,v}=w_{u,v}X_u(x)$, a set like every value here, so that the box
$Z_v(x)=\prod_{u\in\operatorname{In}(v)}z_{u,v}$ is read off the arrows
entering $v$, and each node
its value $X_v(x)$ together with the derivative $D\Phi^{(k)}_v(z_v(x))$
of its operator over the active branches $k\in J_v(z_v(x))$, written as
a single number since every operator here sums its inputs, so that the
entry $[D\Phi^{(k)}_v]_a$ entering \eqref{eq:branch-derivative} is the
same for every predecessor $a$ --
replaced, at the input and bias nodes, which carry no operator, by the
path-gradient $g^{(\kappa)}_{v,o_1}(x)$ of
\eqref{eq:branch-derivative} over $\kappa\in\mathcal M(x)$. Both are
\emph{branch-wise} values, whose convex hulls give the Filippov
extension and the gradient coordinate of $\mathfrak F_{v,o_1}(x)$, here
$[-1,0]$ at $i_1$ and $[2,4]$ at $i_2$. Each arrow takes the colour of the node it leaves, and red marks
the switching node; the colours therefore label nodes, not paths. Paths
and branches should not be confused on the picture either: $i_1$ reaches $o_1$ along three paths -- the
skip edge and one through each hidden node -- which
\eqref{eq:branch-derivative} \emph{sums}, so the two values displayed at
$i_1$ are one per admissible selection and not one per path, the two
elements of $\mathcal M(x)$ differing only at the kink node $h_1$.

The arrows into $h_1$ carry $z_{i_1,h_1}=\{-2\}$,
$z_{i_2,h_1}=\{2\}$ and $z_{b_1,h_1}=\{0\}$, so the box
$Z_{h_1}(x)=\{(-2,2,0)\}$ is the single point $(-2,2,0)$ of $\Sigma_{h_1}$ -- a ReLU kink, whence
$x\in\mathcal D$ -- while $Z_{h_2}(x)=\{(2,1,1)\}$ lies in the open
region where the ReLU is the identity. The operator carried by $h_1$ is
$\Phi_{h_1}(\zeta)=\max\bigl(\zeta_{i_1}+\zeta_{i_2}+\zeta_{b_1},0\bigr)$
on $\mathbb R^{|\operatorname{In}(h_1)|}=\mathbb R^3$, whose switching
set is the hyperplane where that sum vanishes; two of its branches are
then admissible, the constant $\Phi_{h_1}^{(1)}\equiv0$ and the sum
$\Phi_{h_1}^{(2)}(\zeta)=\zeta_{i_1}+\zeta_{i_2}+\zeta_{b_1}$, and they
agree at the kink: $X_{h_1}(x)=\{0\}$
under either, so every downstream value stays a singleton,
$X_{o_1}(x)=\{6\}$ and $X_{o_2}(x)=\{7\}$ --
Theorem~\ref{thm:ffnn-output}(iv)'s guarantee that a ReLU network's
output value never actually branches, only its derivative can. At a
regular input such as $x=(1,1)$, where $Z_{h_1}(x)=\{(1,2,0)\}$ and
$Z_{h_2}(x)=\{(-1,1,1)\}$ avoid the switching sets, $\mathcal M(x)$ is a
singleton and every set in the figure collapses to one number.

For
ReLU, the kink and $X_{h_1}(x)=\{0\}$ are the same event, so
$h_1\notin V(x)$ (Definition~\ref{def:active-graph}): $h_1$ is
\emph{inactive} at this $x$, carrying no forward contribution to
either output. This does not shrink the sum defining $g_{i_1,o_1}$: formula
\eqref{eq:branch-derivative} sums over the paths of the full graph
$\mathcal G$, not over the smaller active graph $G(x)$,
precisely so that a branch which would carry signal through a
currently-inactive node is still represented. Dropping the path
through $h_1$ and keeping only the skip edge and the path through
$h_2$ -- as restricting to $G(x)$ would -- gives the single value
$-1$ below, not the set that follows; the branch $\Phi_{h_1}^{(2)}$
is what recovers the second extreme point. The two branches disagree
on the Jacobian $D\Phi_{h_1}^{(\kappa_{h_1})}$ entering
\eqref{eq:branch-derivative}, $0$ against $1$,
so for $v=i_1$, $o=o_1$, summing the skip edge and the two hidden
paths gives
\[
g^{(\kappa_{h_1}=1)}_{i_1,o_1}(x)
=\underbrace{w_{i_1,o_1}}_{\text{skip}=1}
+\underbrace{w_{i_1,h_1}\cdot0\cdot w_{h_1,o_1}}_{=0}
+\underbrace{w_{i_1,h_2}\cdot1\cdot w_{h_2,o_1}}_{=-2}
=-1,
\qquad
g^{(\kappa_{h_1}=2)}_{i_1,o_1}(x)
=1+\underbrace{1\cdot1\cdot1}_{=1}-2
=0,
\]
so that
\[
\mathfrak F_{i_1,o_1}(x)
=\operatorname{co}\bigl\{(6,-1),(6,0)\bigr\}
=\{6\}\times[-1,0]:
\]
the value coordinate degenerates to the single point $x_{o_1}=6$ --
ReLU's continuity leaves it no room to branch -- while the gradient
coordinate is a nondegenerate segment, exactly the Clarke--Filippov
ambiguity of an automatic-differentiation pass free to resolve the
$h_1$ tie either way. The same computation for $o_2$ gives
$\mathfrak F_{i_1,o_2}(x)=\{7\}\times[-3,-2]$.

\subsection{Backpropagation: Optimal Control}
\label{sec:backprop-oc}

Formula \eqref{eq:branch-derivative} has a second derivation, in the
spirit of \citet{LeCun1988}, which never enumerates paths. Read the
evaluation along a branch selection as a constrained problem: the
quantity of interest is the output $\xi_o$, while the state
$\xi=(\xi_v)_{v\in\mathcal V}$ is tied by the \emph{state equations}
$\xi_v=\Phi^{(\kappa_v)}_v(\zeta_v(\xi))$ at the non-input nodes and by
the \emph{data equations} $\xi_i=\hat x_i$ at the inputs -- a
discrete-time \emph{optimal control} problem whose control is the
weights, which enter nowhere but in the state equations. Adjoining every
constraint with its own multiplier gives the Lagrangian
\eqref{eq:lagrangian-backprop}, and the construction pays off twice. On
the trajectory all constraint terms vanish, so
$\mathcal L^{(\kappa)}=x_o^{(\kappa)}(x)$ \emph{whatever} $\lambda$: the
multipliers are free, and can be spent on cancelling the first-order
effect of the state variations instead of tracking it. Imposing that
cancellation is exactly the adjoint equation
\eqref{eq:adjoint-recursion}.

\begin{theorem}
    \label{thm:backprop-adjoint}
    Let $x\in\mathbb R^d$, $o\in\mathcal O$ and $\kappa\in\mathcal M(x)$
    (see Definition~\ref{def:branch}), and write $\hat x_i$ for the value
    prescribed at an input node, $\hat x_i:=x_i$ if
    $i\in\mathcal I\setminus\mathcal B$ and $\hat x_i:=1$ if
    $i\in\mathcal B$. For state variables $\xi=(\xi_v)_{v\in\mathcal V}$
    and multipliers $\lambda=(\lambda_v)_{v\in\mathcal V}$, let
    \begin{equation}
        \label{eq:lagrangian-backprop}
        \mathcal L^{(\kappa)}(\xi,\lambda):=\xi_o
        +\sum_{v\in\mathcal V\setminus\mathcal I}\lambda_v\Bigl(
        \Phi_v^{(\kappa_v)}\bigl(\zeta_v(\xi)\bigr)-\xi_v\Bigr)
        +\sum_{i\in\mathcal I}\lambda_i\,(\hat x_i-\xi_i),
        \qquad
        \zeta_v(\xi):=\bigl(w_{a,v}\,\xi_a\bigr)_{a\in\operatorname{In}(v)},
    \end{equation}
    for $\xi$ near $\xi^{(\kappa)}:=\bigl(x_v^{(\kappa)}(x)\bigr)_{v\in
    \mathcal V}$. Then
    $\partial\mathcal L^{(\kappa)}/\partial\xi_v(\xi^{(\kappa)},\lambda)=0$
    for every $v\in\mathcal V$ if and only if
    \begin{equation}
        \label{eq:adjoint-recursion}
        \lambda_v=\mathbf 1_{\{v=o\}}+\sum_{u\in\operatorname{Out}(v)}
        \lambda_u\,w_{v,u}\,\bigl[D\Phi_u^{(\kappa_u)}\bigl(z_u^{(\kappa)}(x)
        \bigr)\bigr]_v,
        \qquad v\in\mathcal V .
    \end{equation}
    This backward recursion has a unique solution $\lambda^{(\kappa)}$,
    with $\lambda^{(\kappa)}_v=g^{(\kappa)}_{v,o}(x)$ for every
    $v\in\mathcal V$. Consequently
    $\mathfrak F_{v,o}(x)=\operatorname{co}\{(x_o^{(\kappa)}(x),
    \lambda_v^{(\kappa)}):\kappa\in\mathcal M(x)\}$, and
    $\lambda_v^{(\kappa)}=\partial x_o/\partial x_v(x)$ if
    $x\notin\mathcal D$.
\end{theorem}

Proof of the theorem is provided in
Appendix~\ref{proof:thm:backprop-adjoint}. The state variables $\xi_v$
are free real numbers, distinct from the input $x$, which is fixed:
stationarity in $\lambda$ returns the constraints, that is the forward
recursion \eqref{eq:branch-forward}, whose unique solution is
$\xi^{(\kappa)}$. The Lagrangian is well defined and $C^1$ near
$\xi^{(\kappa)}$ even when $x\in\mathcal D$, because each branch
$\Phi_v^{(\kappa_v)}$ is $C^1$ on an open neighborhood of
$z_v^{(\kappa)}(x)$ (see Definition~\ref{def:piecewise-switching-operator}).
The last sum in \eqref{eq:lagrangian-backprop} enforces the prescribed
input and bias values; its multipliers $\lambda_i$ are, as usual in
optimal control, the sensitivities of the output to those data.

Backpropagation is thus exactly the adjoint equation of the
Lagrangian~\eqref{eq:lagrangian-backprop}: at every node $v$, the
costate $\lambda_v$ collects, from each successor $u$, the local
Jacobian entry $[D\Phi_u^{(\kappa_u)}(z_u^{(\kappa)}(x))]_v$ scaled by
the weight $w_{v,u}$ and by the already computed costate $\lambda_u$ --
one multiplier per node, obtained in a single backward pass, instead of
the sum over the possibly exponentially many paths of
\eqref{eq:branch-derivative}. Because the weights enter
$\mathcal L^{(\kappa)}$ only through the state equations, the same
costates also give, at no further cost, the gradient with respect to
every weight -- the control variable, and the quantity of actual
interest for training.

\begin{corollary}
    \label{cor:backprop-weight-gradient}
    Let $x\notin\mathcal D$, $o\in\mathcal O$, $\kappa$ the unique element
    of $\mathcal M(x)$ and $\lambda=\lambda^{(\kappa)}$ the solution of
    \eqref{eq:adjoint-recursion}. Then, for every edge
    $(u,v)\in\mathcal E$,
    \[
        \frac{\partial x_o}{\partial w_{u,v}}(x)
        =
        \lambda_v\,x_u\,
        \bigl[D\Phi_v^{(\kappa_v)}(z_v(x))\bigr]_u
        =x_u\,\delta_{u,v},
    \]
    where the edge costates
    $\delta_{u,v}:=\lambda_v\,[D\Phi_v^{(\kappa_v)}(z_v(x))]_u$ are the
    unique solution of the backward recursion
    \begin{equation}
        \label{eq:edge-costate-recursion}
        \delta_{u,v}=\bigl[D\Phi_v^{(\kappa_v)}(z_v(x))\bigr]_u
        \Bigl(\mathbf 1_{\{v=o\}}+\sum_{r\in\operatorname{Out}(v)}
        w_{v,r}\,\delta_{v,r}\Bigr),
        \qquad (u,v)\in\mathcal E .
    \end{equation}
\end{corollary}

Proof of the corollary is provided in
Appendix~\ref{proof:cor:backprop-weight-gradient}.
The recursion \eqref{eq:edge-costate-recursion} computes every weight
gradient in the same backward pass as \eqref{eq:adjoint-recursion},
which it rewrites edge by edge: the bracket is $\lambda_v$, and
$\lambda_u=\mathbf 1_{\{u=o\}}+\sum_{v\in\operatorname{Out}(u)}w_{u,v}\,\delta_{u,v}$.
For a neuron $\Phi_v(\zeta)=\sigma_v\bigl(\sum_a\zeta_a\bigr)$, the
Jacobian entry is $\sigma_v'(\tilde z_v)$ for every $u$, so that
$\delta_{u,v}=\lambda_v\,\sigma_v'(\tilde z_v)$ is the classical
``delta'' of backpropagation, one per neuron, and
$\partial x_o/\partial w_{u,v}=x_u\,\delta_v$.

\FloatBarrier

\section{Relevance Propagation as an Entropic Projection}
\label{sec:LRP}

\subsection{Relevance flows on the active graph}
\label{sec:admissible-flows}

As outlined in Section~\ref{sec:intro}, classical relevance
propagation arises from a single variational principle: the
\emph{entropic projection} \citep{Villani2008,Peyre2019} of the backward
transition law induced by the forward activation flow onto the set of
divergence-free relevance flows consistent with the output, a projection
of value zero, the flow it selects having exactly that transition law.
Specifically, given the active
computational graph $G(x)$ induced by an input $x$, we define the
admissible set $\mathcal{A}_o(x)$ of flow relevances as those
satisfying a conservation constraint at every hidden node -- a
discrete analogue of Kirchhoff's current law -- and we show in this
section that the classical LRP$_{z^+}$ and LRP$_{\alpha,\beta}$ rules of
\citet{Bach2015} (see also \citet{Montavon2017,Montavon2018,Montavon2019}
for extensions and overviews) are recovered as the unique solutions of
relative-entropy minimization problems over this set, the relative
entropy being taken between backward transition laws; LRP$_0$ is
recovered in the same way, its reference being the signed forward
contributions $z$ themselves. For a non-negative reference and a
non-negative flow, this is the relative entropy of two Markov measures
on the backward paths of the network
(Appendix~\ref{proof:path-space}); for a signed reference, it is an
extension under a sign constraint, without this reading.

This variational formulation offers three key advantages.
First, it provides a \emph{rigorous foundation} for relevance
propagation, replacing the heuristic redistribution rules of
classical LRP with a well-posed optimization problem whose
solution is unique. Second, it reveals the
\emph{geometric structure} of attribution: the propagation
kernels $K_{u,v}$ are the transition laws of these minimizers, and the
conservation of relevance is a constraint of the problem at every
node, from which global conservation follows on the DAG, rather than an
assumption. Third,
it opens the door to \emph{new attribution methods} that keep the
same conservation constraint: changing the reference measure in the
KL divergence changes the rule, and taking the relative entropy edge
by edge yields Sinkhorn Relevance Propagation
(Section~\ref{sec:SRP}).

Section~\ref{sec:admissible-flows} defines the admissible relevance flows and their
conservation (Proposition~\ref{thm:RelevanceConservation}).
Section~\ref{sec:entropic-projection} proves the projection theorem
(Theorem~\ref{thm:RP0}), derives from it LRP$_0$, LRP$_{z^+}$,
LRP$_{\alpha,\beta}$ and LRPN$_{\alpha,\beta}$
(Corollary~\ref{cor:lrp-references}), and describes these rules on
networks of ReLU and max nodes (Corollary~\ref{thm:RP0ReLU}):
Gradient$\times$Input for LRP$_0$, continuity and jumps at a kink and at
a change of winner. Section~\ref{sec:SRP} introduces SRP, and
Section~\ref{sec:app-images} compares the rules on VGG-16 and
ConvNeXt-Tiny.

We now formally define, for an output node $o\in\mathcal O$, the node
relevances $\mu^o_u$ together with the associated flow relevances
$\pi^o_{u,v}$ along each active edge of the DAG.
\begin{definition}
    \label{def:admissible-set}
    Let $\mathcal N_{\mathcal F}(\mathcal G)$ be a feed-forward Filippov
    neural network, $x\in\mathbb R^d$ and $\kappa\in\mathcal M(x)$. Write
    $x_v:=x_v^{(\kappa)}(x)$ for the node values along $\kappa$ and
    $G^{(\kappa)}(x)=(V^{(\kappa)}(x),E^{(\kappa)}(x))$ for the active
    graph they determine, as in Definition~\ref{def:active-graph}. For an
    active output $o\in\mathcal O$, define the admissible set
    \begin{multline*}
    \mathcal A_{o}^{(\kappa)}(x)
    :=
    \Bigl\{
    \pi^{o,\kappa}\in\mathbb{R}^{E^{(\kappa)}(x)}\ :\
    (\operatorname{div}_{\scriptscriptstyle G^{(\kappa)}(x)}\pi^{o,\kappa})(v)=0
    \ \ \forall v\in H^{(\kappa)}(x),\\
    \sum_{u:u\to o}\pi_{u,o}^{o,\kappa} = x_o^{(\kappa)},
    \qquad
    \sum_{u:u\to o'}\pi_{u,o'}^{o,\kappa} = 0\ \ \forall o'\in\mathcal O\setminus\{o\}
    \Bigr\}
    \end{multline*} where $H^{(\kappa)}(x) = V^{(\kappa)}(x)\cap \mathcal H$ are the hidden active nodes and
    \[
            (\operatorname{div}_{\scriptscriptstyle G^{(\kappa)}(x)}\pi^{o,\kappa})(v)
            =
            \sum_{u\in V^{(\kappa)}(x):u\to v}\pi^{o,\kappa}_{u,v}
            -
            \sum_{w\in V^{(\kappa)}(x):v\to w}\pi^{o,\kappa}_{v,w}
    \]
    is the divergence operator at node $v$. Both the active graph and the
    boundary mass $x_o^{(\kappa)}$ are read off the values along $\kappa$, so
    $\mathcal A_{o}^{(\kappa)}(x)$ does depend on the selection; when the
    node values do not branch -- at every $x\notin\mathcal D$
    (Theorem~\ref{thm:ffnn-output}(iii)), and at every $x$ when the
    branches agree on the switching sets -- it is the same for every
    selection, and we then drop $\kappa$, writing $\pi^{o}_{u,v}$,
    $\mu^{o}_u$ and $x_o$, as we do throughout Sections
    \ref{sec:LRP}--\ref{sec:transformers}. The $\pi^{o,\kappa}_{u,v}$ are
    called the flow relevances for the active output $o\in\mathcal O$. They
    are carried by the active edges alone, no variable being attached to
    the others. The relevance of a node $u\in\mathcal V\setminus\mathcal O$
    is their sum over the active edges leaving it,
    \[
    \mu^{o,\kappa}_u={\displaystyle \sum_{v\in V^{(\kappa)}(x):{u}\to v}}
    \pi^{o,\kappa}_{u,v},
    \]
    an empty sum, hence $\mu^{o,\kappa}_u=0$, at a node from which no
    active edge leaves -- in particular at every such node outside
    $V^{(\kappa)}(x)$. The relevance of an output is the sum of the flows
    entering it: $\mu^{o,\kappa}_o=x_o^{(\kappa)}$, and
    $\mu^{o,\kappa}_{o'}=0$ for $o'\neq o$.
\end{definition}

\begin{figure}[H]
    \centering
    \begin{tikzpicture}[
        scale=0.74,
        transform shape,
        x=3.9cm,
        y=2.6cm,
        >=stealth,
        neuron/.style={circle,draw,font=\scriptsize,inner sep=1pt,minimum size=1.4cm,thick},
        wl/.style={inner sep=1.5pt,font=\scriptsize,fill=white,fill opacity=0.85,
                   text opacity=1,rounded corners=1pt},
        rel/.style={->,thick,red!75!black},
        fl/.style={inner sep=1.5pt,font=\scriptsize,red!75!black,fill=white,
                   fill opacity=0.85,text opacity=1,rounded corners=1pt},
        faint/.style={gray!55,dashed},
        every node/.style={font=\small}
    ]

    \node[neuron,blue] (x1) at (0,1.8)
        {\shortstack{$X_{i_1}=\{-2\}$\\[-1pt]{\color{red!75!black}$\mu^{o_1,\kappa}_{i_1}$}}};
    \node[neuron,green!30!black!70] (x2) at (0,0.6)
        {\shortstack{$X_{i_2}=\{1\}$\\[-1pt]{\color{red!75!black}$\mu^{o_1,\kappa}_{i_2}$}}};
    \node[neuron] (b1) at (0,-0.6)
        {\shortstack{$X_{b_1}=\{+1\}$\\[-1pt]{\color{red!75!black}$\mu^{o_1,\kappa}_{b_1}$}}};

    \node[neuron,gray!55] (h1) at (2,1.15)
        {\shortstack{{\color{gray!55}$X_{h_1}=\{0\}$}\\[-1pt]{\color{gray!55}\tiny $h_1\notin V(x)$}}};
    \node[neuron,teal!70!black] (h2) at (2,-0.05)
        {\shortstack{$X_{h_2}=\{4\}$\\[-1pt]{\color{red!75!black}$\mu^{o_1,\kappa}_{h_2}$}}};
    \node[neuron] (b2) at (2,-1.25)
        {\shortstack{$X_{b_2}=\{+1\}$\\[-1pt]{\color{red!75!black}$\mu^{o_1,\kappa}_{b_2}=0$}}};

    \node[neuron,olive] (y1) at (4,1.8)
        {\shortstack{$X_{o_1}=\{6\}$\\[-1pt]{\color{red!75!black}$\mu^{o_1,\kappa}_{o_1}=x_{o_1}$}}};
    \node[neuron,olive] (y2) at (4,0.6)
        {\shortstack{$X_{o_2}=\{7\}$\\[-1pt]{\color{red!75!black}$\mu^{o_1,\kappa}_{o_2}=0$}}};

    \node at (0,2.6) {\textbf{Input}};
    \node at (2,2.6) {\textbf{Hidden ReLU}};
    \node at (4,2.6) {\textbf{Output}};

    \draw[->,faint] (x1) -- (h1);
    \draw[->,faint] (x2) -- (h1);
    \draw[->,faint] (b1) -- (h1);
    \draw[->,faint] (h1) -- (y1);
    \draw[->,faint] (h1) -- (y2);
    \draw[->,faint] (b2) -- (y1);

    \draw[->,blue] (x1) -- node[wl,sloped,above,pos=0.3,text=blue] {$z_{i_1,h_2}\!=\!\{2\}$} (h2);
    \draw[->,green!30!black!70] (x2) -- node[wl,sloped,above,pos=0.3,text=green!30!black!70] {$z_{i_2,h_2}\!=\!\{1\}$} (h2);
    \draw[->] (b1) -- node[wl,sloped,above,pos=0.3] {$z_{b_1,h_2}\!=\!\{1\}$} (h2);
    \draw[->,teal!70!black] (h2) -- node[wl,sloped,above,pos=0.3,text=teal!70!black] {$z_{h_2,o_1}\!=\!\{8\}$} (y1);
    \draw[->,teal!70!black] (h2) -- node[wl,sloped,below,pos=0.62,text=teal!70!black] {$z_{h_2,o_2}\!=\!\{8\}$} (y2);
    \draw[->] (b2) -- node[wl,sloped,below,pos=0.16] {$z_{b_2,o_2}\!=\!\{-1\}$} (y2);
    \draw[->,blue] (x1) -- node[wl,below,pos=0.5,text=blue] {$z_{i_1,o_1}\!=\!\{-2\}$} (y1);

    \draw[rel] (x1.10) -- node[fl,above,pos=0.5] {$\pi^{o_1,\kappa}_{i_1,o_1}$} (y1.170);
    \draw[rel] (x1.-41.7) -- node[fl,sloped,below,pos=0.3] {$\pi^{o_1,\kappa}_{i_1,h_2}$} (h2.158.3);
    \draw[rel] (x2.-22.2) -- node[fl,sloped,below,pos=0.3] {$\pi^{o_1,\kappa}_{i_2,h_2}$} (h2.177.8);
    \draw[rel] (b1.0.4) -- node[fl,sloped,below,pos=0.3] {$\pi^{o_1,\kappa}_{b_1,h_2}$} (h2.200.4);
    \draw[rel] (h2.21.7) -- node[fl,sloped,below,pos=0.62] {$\pi^{o_1,\kappa}_{h_2,o_1}$} (y1.221.7);

    \end{tikzpicture}
    \caption{Relevance flows and node relevances for the output $o_1$, on
    the network of Figure~\ref{fig:DAG example} at $x=(-2,1)$, for
    either selection $\kappa\in\mathcal M(x)$.}
    \label{fig:DAG relevance}
\end{figure}
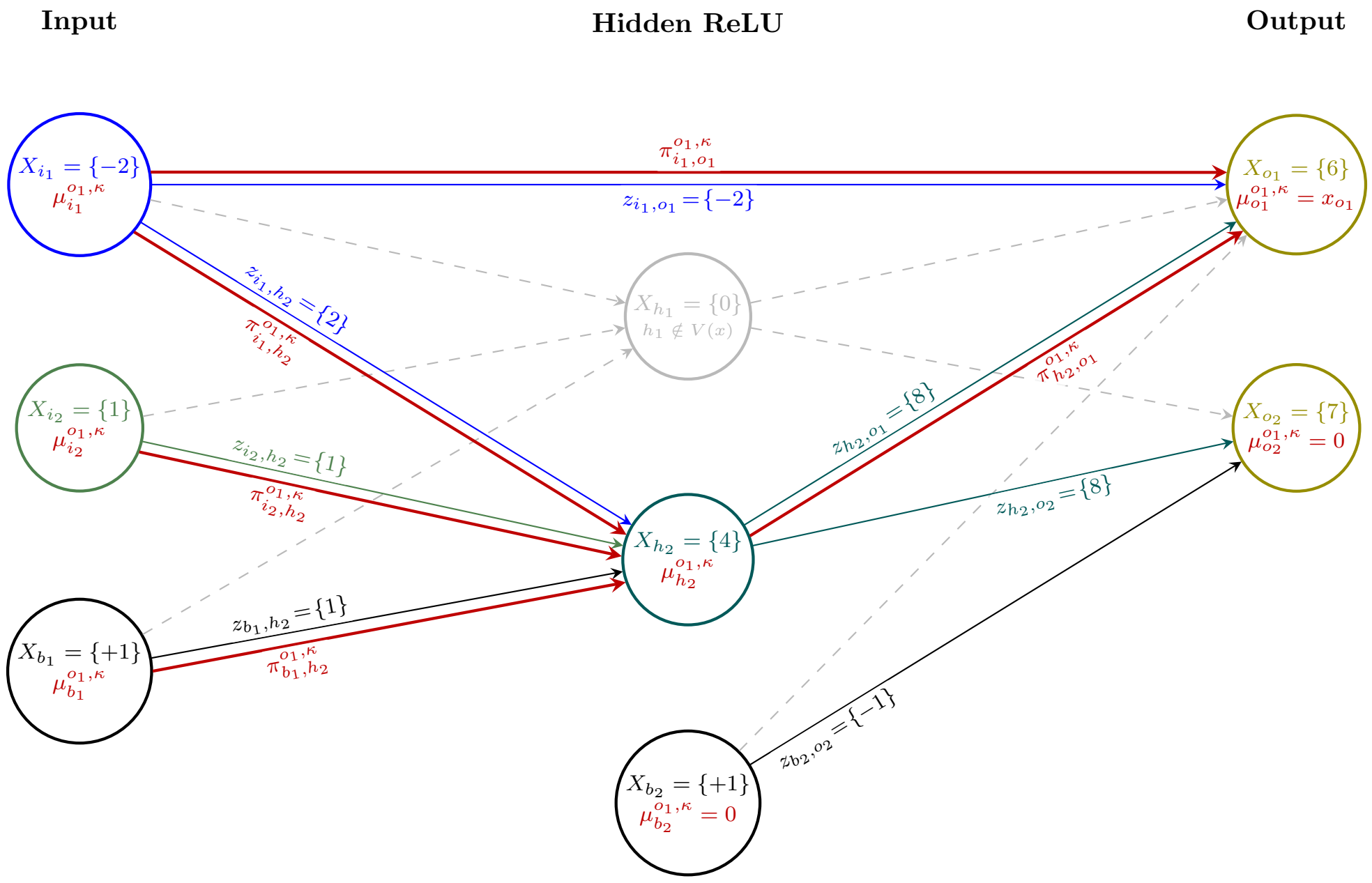

Figure~\ref{fig:DAG relevance} illustrates these notions on the running
example of Figure~\ref{fig:DAG example}, at the same input
$x=(-2,1)$ and for the output $o_1$. The kink node is the one casualty:
$X_{h_1}(x)=\{0\}$, so $h_1\notin V(x)$ and the five edges it carries
leave $E(x)$, together with the zero-weight edges $b_1\to h_1$ and
$b_2\to o_1$. On what remains, a relevance flow $\pi^{o_1,\kappa}_{u,v}$ is
attached to every active edge, the relevance of a non-output node is the
sum of the flows leaving it, $\mu^{o_1,\kappa}_u=\sum_{v\in V^{(\kappa)}(x):u\to v}\pi^{o_1,\kappa}_{u,v}$,
the flows entering $o_1$ add up to its value,
$\pi^{o_1,\kappa}_{h_2,o_1}+\pi^{o_1,\kappa}_{i_1,o_1}=x_{o_1}^{(\kappa)}=6$, and no flow enters
the other output: $\mu^{o_1,\kappa}_{b_2}=0$, its only active edge feeding
$o_2$. Here $\mathcal M(x)$ has two elements: the outputs being linear
and $h_2$ active, $\kappa_{o_1}=\kappa_{o_2}=1$ and $\kappa_{h_2}=2$ are
forced, while $\kappa_{h_1}\in\{1,2\}$ is free at the kink. The two
branches of $h_1$ agreeing in value, both selections give the same node
values, hence the same active graph and the same flows: the figure holds
for either.

Any admissible flow in $\mathcal A_o(x)$ already conserves the output
relevance globally:

\begin{proposition}
    \label{thm:RelevanceConservation}
    Let $\mathcal N_{\mathcal F}(\mathcal G)$ be a feed-forward Filippov
    neural network, $x\in\mathbb R^d$, $\kappa\in\mathcal M(x)$ and
    $o\in\mathcal O$. Every admissible relevance flow
    $\pi^{o,\kappa}\in\mathcal A_{o}^{(\kappa)}(x)$ satisfies
    $
    {\displaystyle \sum_{u\in \mathcal I} \mu^{o,\kappa}_{u}}
    = x_o^{(\kappa)}(x).
    $
\end{proposition}
Proof of the proposition is provided in
Appendix~\ref{proof:thm:RelevanceConservation}. An input outside $\mathcal I(x)$ has
$x_u=0$, hence no active outgoing edge and, by
Definition~\ref{def:admissible-set}, $\mu_u^{o,\kappa}=0$; so
$\sum_{u\in\mathcal I}\mu_u^{o,\kappa}
=\sum_{u\in\mathcal I(x)}\mu_u^{o,\kappa}$: the
conservation law is equivalently stated over the active input nodes
$\mathcal I(x)$ (Definition~\ref{def:active-graph}) -- the form used
from here on, with $\kappa$ dropped as agreed whenever the node values
do not branch, so that the right-hand side is simply $x_o$.

Consequently, the relevance associated with output $o$ is neither created nor destroyed inside the network; it is only redistributed among the active paths of $G(x)$.

\subsection{The entropic projection and the classical rules}
\label{sec:entropic-projection}

We fix an input $x$, a selection $\kappa\in\mathcal M(x)$ and an active
output $o\in\mathcal O$; the node values, the active graph and the
admissible sets are those of Definition~\ref{def:admissible-set}; we
keep $\kappa$ in the statements of Theorem~\ref{thm:RP0} and
Corollary~\ref{cor:lrp-references} and drop it elsewhere, as agreed
there. For
$a\in\mathbb R$ we write $a=a^+-a^-$ with $a^\pm:=\max(\pm a,0)$, so that
the edge contributions $z_{u,v}=w_{u,v}x_u$ split into two non-negative
measures $z^\pm=(z^\pm_{u,v})_{(u,v)\in E(x)}$ on the active graph, and
we write $\tilde z^\pm_v:=\sum_{u:u\to v}z^\pm_{u,v}$.

A relevance propagation rule is a \emph{backward transition law}: at
every node $v$, it splits the relevance $\mu_v$ of $v$ among its
predecessors, following a \emph{reference} $\rho\in\mathbb R^{E(x)}$
read off the forward pass, of either sign. Theorem~\ref{thm:RP0}
characterizes this rule as the unique admissible flow with zero
relative entropy to the reference, the entropy being taken between
backward transition laws rather than edge by edge. For a flow
$\pi\in\mathbb R^{E(x)}$ we write $m_v(\pi):=\sum_{u:u\to v}\pi_{u,v}$
and $n_v(\pi):=\sum_{u:u\to v}|\pi_{u,v}|$ for its inflow and absolute
inflow at $v$.

\begin{theorem}
    \label{thm:RP0}
    Let $x\in\mathbb R^d$, $\kappa\in\mathcal M(x)$, $o\in\mathcal O$ an
    active output, and $\rho\in\mathbb R^{E^{(\kappa)}(x)}$ with
    $\tilde\rho_v:=\sum_{u:u\to v}\rho_{u,v}\neq0$ for every
    $v\in H^{(\kappa)}(x)\cup\{o\}$. Define the kernel of $\rho$ and its absolute law,
    \[
        K_{u,v}:=\frac{\rho_{u,v}}{\tilde\rho_v},
        \qquad
        \bar K_{u,v}:=\frac{|\rho_{u,v}|}{\sum_{r:r\to v}|\rho_{r,v}|},
    \]
    both set to $0$ if $\tilde\rho_v=0$, and the \emph{conditional
    relative entropy} of $\pi\in\mathbb R^{E^{(\kappa)}(x)}$,
    \begin{equation}
        \label{eq:cond-entropy}
        \mathcal H(\pi\,\|\,K):=\sum_{v:\,n_v(\pi)\neq0}n_v(\pi)\;
        \mathrm{KL}\Bigl(\frac{|\pi_{\cdot,v}|}{n_v(\pi)}\,\Big\|\,\bar K_{\cdot,v}\Bigr)
    \end{equation}
    if $\pi$ is \emph{sign-coherent with $K$} -- at every node $v$ there
    is $\sigma_v\in\{-1,1\}$ with $\sigma_v\,\pi_{u,v}K_{u,v}\ge0$ for
    every edge $(u,v)\in E^{(\kappa)}(x)$ entering $v$ -- and $\mathcal H(\pi\|K):=+\infty$ otherwise. Then the
    problem
    \[
        \min_{\pi^{o,\kappa}\in\mathcal A_o^{(\kappa)}(x)}\ \mathcal H(\pi^{o,\kappa}\,\|\,K)
    \]
    has a unique minimizer $\pi^{o,\kappa,\ast}$, its minimal value is $0$, and
    \[
        \pi^{o,\kappa,\ast}_{u,v}=K_{u,v}\,\mu^{o,\kappa}_v,\qquad (u,v)\in E^{(\kappa)}(x),
    \]
    where the node relevances satisfy the backward recursion
    \[
        \mu^{o,\kappa}_o=x^{(\kappa)}_o,\qquad
        \mu^{o,\kappa}_{o'}=0\ \ (o'\in\mathcal O\setminus\{o\}),
        \qquad
        \mu^{o,\kappa}_u=\sum_{v:\,u\to v}K_{u,v}\,\mu^{o,\kappa}_v\ \
        (u\in V^{(\kappa)}(x)\setminus\mathcal O).
    \]
    In particular, $\sum_{u\in\mathcal I(x)}\mu^{o,\kappa}_u=x^{(\kappa)}_o$.
    Letting $\kappa$ range over $\mathcal M(x)$, the relevance of a node
    $u$ is the set
    $\operatorname{co}\{\mu_u^{o,\kappa}:\kappa\in\mathcal M(x)\}$, a
    singleton whenever neither the node values nor the reference depend on
    $\kappa$.
\end{theorem}
Proof of the theorem is provided in Appendix~\ref{proof:thm:RP0}.
Both ingredients of \eqref{eq:cond-entropy} are needed for a signed
reference: with $K=(2,-1)$ and inflow $1$, the naive form
$\sum|\pi_{u,v}|\log\bigl(|\pi_{u,v}|/(|K_{u,v}||m_v(\pi)|)\bigr)$ is
$-\log3<0$ at $\pi=(2/3,1/3)$. Normalizing $|K|$ into $\bar K$ restores
Gibbs' inequality, and sign coherence excludes $(2/3,1/3)$, leaving
$(2,-1)=K$ as the only flow of inflow $1$ with zero entropy.

The theorem turns the choice of a rule into the choice of a reference.
It characterizes the flow once the reference is fixed, and only through
the kernel $K$: rescaling the column $\rho_{\cdot,v}$ of a node by a
nonzero constant changes neither $K$ nor the minimizer. It does not select
a canonical reference, nor guarantee the faithfulness of the result,
which Section~\ref{sec:app-images} measures.
The minimizer rescales $\rho$ at the head of each edge only,
$\pi_{u,v}=\rho_{u,v}\,\mu_v/\tilde\rho_v$: the entropy fixes the
proportions $|\pi_{u,v}|/n_v(\pi)$ to those of $|\rho_{u,v}|$, and the
coherence constraint fixes the signs, relevance of the sign of $\mu_v$
flowing back along the edges where $\rho$ has the sign of
$\tilde\rho_v$, and of the opposite sign along the others. A
non-negative reference therefore yields relevances of the sign of
$x_o$, a signed one relevances of both signs. The natural references
are built from the forward contributions $z$ and their sign channels
$z^\pm$, and the three classical rules are three ways of weighting
these channels; we set $\tilde z_v:=\sum_{u:u\to v}z_{u,v}
=\tilde z^+_v-\tilde z^-_v$.

\begin{corollary}
    \label{cor:lrp-references}
    Let $x\in\mathbb R^d$, $\kappa\in\mathcal M(x)$ and $o\in\mathcal O$
    an active output, with contributions
    $z_{u,v}:=w_{u,v}\,x^{(\kappa)}_u$ on $E^{(\kappa)}(x)$. The minimizer
    $\pi^{o,\kappa,\ast}$ of Theorem~\ref{thm:RP0} is, for each of the
    following references $\rho$ (those of (iii) and (iv) being given on the
    edges entering $H^{(\kappa)}(x)\cup\{o\}$, and $0$ on the edges
    entering the other outputs):

    \smallskip\noindent
    (i) \emph{LRP$_0$.} If $\rho=z$ and $\tilde z_v\neq0$ for every
    $v\in H^{(\kappa)}(x)\cup\{o\}$, the minimizer is
    \[
        \pi^{o,\kappa,\ast}_{u,v}=\frac{z_{u,v}}{\tilde z_v}\,\mu^{o,\kappa}_v .
    \]
    If moreover $\tilde z_{o'}\neq0$ for every $o'\in\mathcal O\setminus\{o\}$,
    then, for $\pi^{o,\kappa}\in\mathcal A_o^{(\kappa)}(x)$,
    \[
        \mathcal H(\pi^{o,\kappa}\,\|\,K)
        =\sum_{v:\,n_v(\pi^{o,\kappa})\neq0}n_v(\pi^{o,\kappa})\;
        \mathrm{KL}\Bigl(\frac{|\pi^{o,\kappa}_{\cdot,v}|}{n_v(\pi^{o,\kappa})}\,\Big\|\,
        \frac{|z_{\cdot,v}|}{\sum_{r:r\to v}|z_{r,v}|}\Bigr)
    \]
    when, at every $v$, the products $\pi^{o,\kappa}_{u,v}z_{u,v}$ over the
    edges $(u,v)\in E^{(\kappa)}(x)$ entering $v$ have a common sign, and
    $\mathcal H(\pi^{o,\kappa}\|K)=+\infty$ otherwise.

    \smallskip\noindent
    (ii) \emph{LRP$_{z^+}$.} If $\rho=z^+$ and $\tilde z^+_v>0$ for every
    $v\in H^{(\kappa)}(x)\cup\{o\}$, the minimizer is
    \[
        \pi^{o,\kappa,\ast}_{u,v}=\frac{z^+_{u,v}}{\tilde z^+_v}\,\mu^{o,\kappa}_v .
    \]

    \smallskip\noindent
    (iii) \emph{LRP$_{\alpha,\beta}$.} If $\alpha,\beta\in\mathbb R$ with
    $\alpha+\beta=1$,
    $\rho_{u,v}=\alpha\,z^+_{u,v}/\tilde z^+_v+\beta\,z^-_{u,v}/\tilde z^-_v$,
    and $\tilde z^+_v>0$, $\tilde z^-_v>0$ for every
    $v\in H^{(\kappa)}(x)\cup\{o\}$, the minimizer is
    \[
        \pi^{o,\kappa,\ast}_{u,v}=\Bigl(\alpha\,\frac{z^+_{u,v}}{\tilde z^+_v}
        +\beta\,\frac{z^-_{u,v}}{\tilde z^-_v}\Bigr)\mu^{o,\kappa}_v .
    \]

    \smallskip\noindent
    (iv) \emph{LRPN$_{\alpha,\beta}$ (LRP$_{\alpha,\beta}$ with node-wise
    weights).} If, for every $v\in H^{(\kappa)}(x)\cup\{o\}$,
    $\alpha_v,\beta_v\in\mathbb R$ (which may depend on $x$) satisfy
    $\alpha_v+\beta_v=1$ and $\tilde z^+_v+\tilde z^-_v>0$, and
    \[
        \rho_{u,v}=
        \begin{cases}
            \alpha_v\,z^+_{u,v}/\tilde z^+_v+\beta_v\,z^-_{u,v}/\tilde z^-_v
            & \text{if } \tilde z^+_v>0 \text{ and } \tilde z^-_v>0,\\
            z^+_{u,v}/\tilde z^+_v & \text{if } \tilde z^-_v=0,\\
            z^-_{u,v}/\tilde z^-_v & \text{if } \tilde z^+_v=0,
        \end{cases}
    \]
    then $\tilde\rho_v=1$, and the minimizer is
    \[
        \pi^{o,\kappa,\ast}_{u,v}=\rho_{u,v}\,\mu^{o,\kappa}_v .
    \]
\end{corollary}
Proof of the corollary is provided in
Appendix~\ref{proof:cor:lrp-references}.

The three references are of the form
$\rho_{u,v}=c^+_v\,z^+_{u,v}-c^-_v\,z^-_{u,v}$, with
$(c^+_v,c^-_v)=(1,1)$, $(1,0)$ and
$(\alpha/\tilde z^+_v,-\beta/\tilde z^-_v)$; for $\beta\le0\le\alpha$
they are the rules of \citet{Bach2015,Montavon2017,Montavon2019},
extended to the general active DAG $G(x)$. Case (iv) contains the other
three wherever they apply: (iii) is $(\alpha_v,\beta_v)=(\alpha,\beta)$
at every node, (ii) is $(\alpha_v,\beta_v)=(1,0)$, and (i) is
$(\alpha_v,\beta_v)=(\tilde z^+_v/\tilde z_v,-\tilde z^-_v/\tilde z_v)$.
Only (ii) yields relevances of the sign of $x_o$ for every $x$; (iv) does
so whenever $\alpha_v,\beta_v\in[0,1]$, the reference being then
non-negative.

Weights that depend on $x$ give case (iv) a Hebbian-inspired reading. If
$\alpha_v$ grows with the activity $x_v$ of $v$, then, at a given
relevance $\mu_v\ge0$ and given channel shares $z^\pm_{u,v}/\tilde z^\pm_v$,
a more active node sends a larger share $\alpha_v$ of its relevance to its
excitatory contributions $z^+_{u,v}=(w_{u,v}x_u)^+$, and a weakly activated
node a larger share $\beta_v$ to its inhibitory ones, much as Hebb's
principle \citep{Hebb1949} favours the connections whose two ends are
active together. The analogy concerns this split only: the relevance of
an edge is not monotone in the product of the activities of its two
ends, the shares and $\mu_v$ depending on $x$ too, and no weight is
updated. Section~\ref{sec:app-images} takes $\alpha_v=\bar x_v$, the value
of $v$ divided by the largest value of its layer.

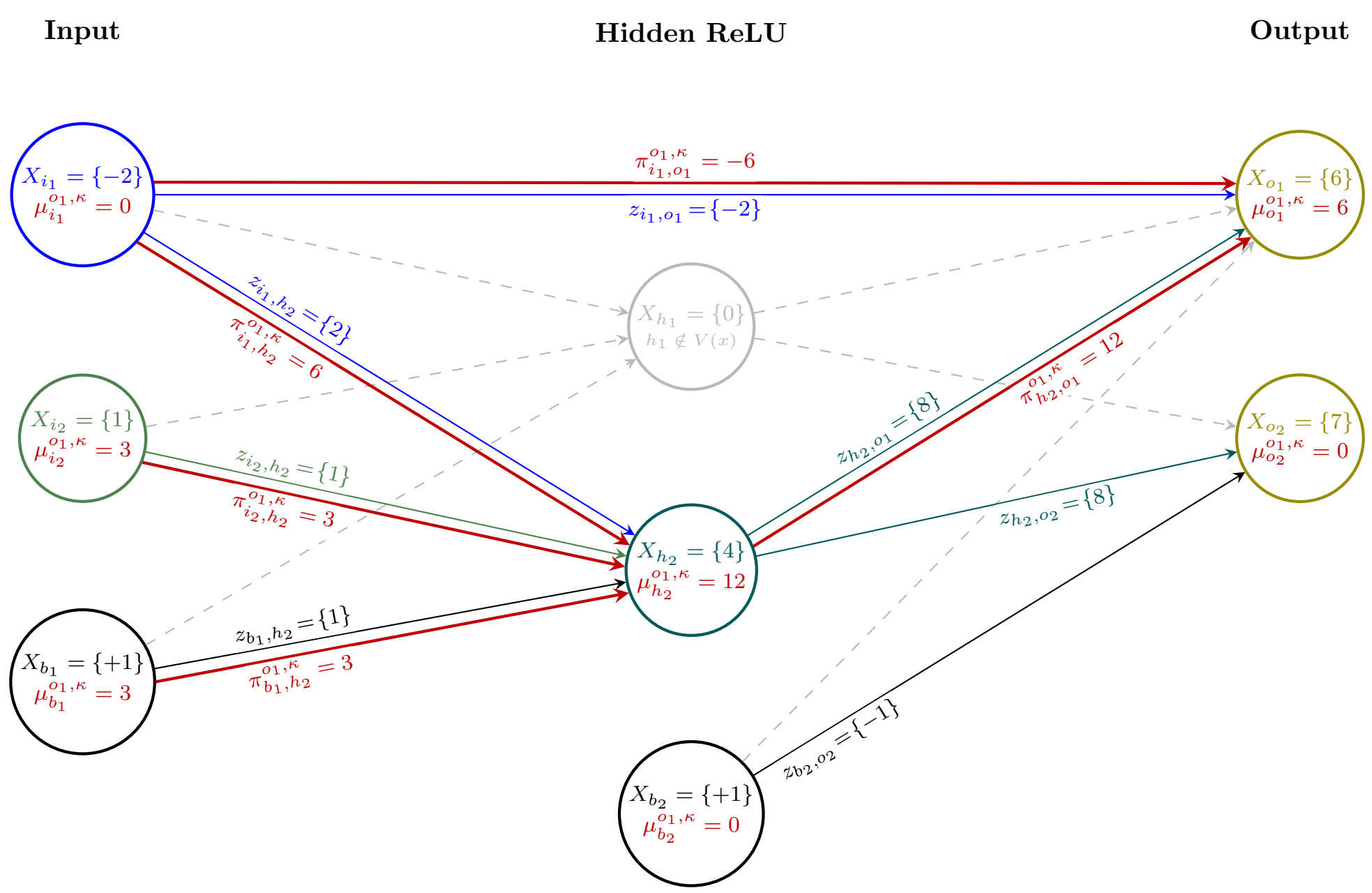
\begin{figure}[H]
    \centering
    \begin{tikzpicture}[
        scale=0.74,
        transform shape,
        x=3.9cm,
        y=2.6cm,
        >=stealth,
        neuron/.style={circle,draw,font=\scriptsize,inner sep=1pt,minimum size=1.4cm,thick},
        wl/.style={inner sep=1.5pt,font=\scriptsize,fill=white,fill opacity=0.85,
                   text opacity=1,rounded corners=1pt},
        rel/.style={->,thick,red!75!black},
        fl/.style={inner sep=1.5pt,font=\scriptsize,red!75!black,fill=white,
                   fill opacity=0.85,text opacity=1,rounded corners=1pt},
        faint/.style={gray!55,dashed},
        every node/.style={font=\small}
    ]

    \node[neuron,blue] (x1) at (0,1.8)
        {\shortstack{$X_{i_1}=\{-2\}$\\[-1pt]{\color{red!75!black}$\mu^{o_1,\kappa}_{i_1}=0$}}};
    \node[neuron,green!30!black!70] (x2) at (0,0.6)
        {\shortstack{$X_{i_2}=\{1\}$\\[-1pt]{\color{red!75!black}$\mu^{o_1,\kappa}_{i_2}=3$}}};
    \node[neuron] (b1) at (0,-0.6)
        {\shortstack{$X_{b_1}=\{+1\}$\\[-1pt]{\color{red!75!black}$\mu^{o_1,\kappa}_{b_1}=3$}}};

    \node[neuron,gray!55] (h1) at (2,1.15)
        {\shortstack{{\color{gray!55}$X_{h_1}=\{0\}$}\\[-1pt]{\color{gray!55}\tiny $h_1\notin V(x)$}}};
    \node[neuron,teal!70!black] (h2) at (2,-0.05)
        {\shortstack{$X_{h_2}=\{4\}$\\[-1pt]{\color{red!75!black}$\mu^{o_1,\kappa}_{h_2}=12$}}};
    \node[neuron] (b2) at (2,-1.25)
        {\shortstack{$X_{b_2}=\{+1\}$\\[-1pt]{\color{red!75!black}$\mu^{o_1,\kappa}_{b_2}=0$}}};

    \node[neuron,olive] (y1) at (4,1.8)
        {\shortstack{$X_{o_1}=\{6\}$\\[-1pt]{\color{red!75!black}$\mu^{o_1,\kappa}_{o_1}=6$}}};
    \node[neuron,olive] (y2) at (4,0.6)
        {\shortstack{$X_{o_2}=\{7\}$\\[-1pt]{\color{red!75!black}$\mu^{o_1,\kappa}_{o_2}=0$}}};

    \node at (0,2.6) {\textbf{Input}};
    \node at (2,2.6) {\textbf{Hidden ReLU}};
    \node at (4,2.6) {\textbf{Output}};

    \draw[->,faint] (x1) -- (h1);
    \draw[->,faint] (x2) -- (h1);
    \draw[->,faint] (b1) -- (h1);
    \draw[->,faint] (h1) -- (y1);
    \draw[->,faint] (h1) -- (y2);
    \draw[->,faint] (b2) -- (y1);

    \draw[->,blue] (x1) -- node[wl,sloped,above,pos=0.3,text=blue] {$z_{i_1,h_2}\!=\!\{2\}$} (h2);
    \draw[->,green!30!black!70] (x2) -- node[wl,sloped,above,pos=0.3,text=green!30!black!70] {$z_{i_2,h_2}\!=\!\{1\}$} (h2);
    \draw[->] (b1) -- node[wl,sloped,above,pos=0.3] {$z_{b_1,h_2}\!=\!\{1\}$} (h2);
    \draw[->,teal!70!black] (h2) -- node[wl,sloped,above,pos=0.3,text=teal!70!black] {$z_{h_2,o_1}\!=\!\{8\}$} (y1);
    \draw[->,teal!70!black] (h2) -- node[wl,sloped,below,pos=0.62,text=teal!70!black] {$z_{h_2,o_2}\!=\!\{8\}$} (y2);
    \draw[->] (b2) -- node[wl,sloped,below,pos=0.16] {$z_{b_2,o_2}\!=\!\{-1\}$} (y2);
    \draw[->,blue] (x1) -- node[wl,below,pos=0.5,text=blue] {$z_{i_1,o_1}\!=\!\{-2\}$} (y1);

    \draw[rel] (x1.10) -- node[fl,above,pos=0.5] {$\pi^{o_1,\kappa}_{i_1,o_1}=-6$} (y1.170);
    \draw[rel] (x1.-41.7) -- node[fl,sloped,below,pos=0.3] {$\pi^{o_1,\kappa}_{i_1,h_2}=6$} (h2.158.3);
    \draw[rel] (x2.-22.2) -- node[fl,sloped,below,pos=0.3] {$\pi^{o_1,\kappa}_{i_2,h_2}=3$} (h2.177.8);
    \draw[rel] (b1.0.4) -- node[fl,sloped,below,pos=0.3] {$\pi^{o_1,\kappa}_{b_1,h_2}=3$} (h2.200.4);
    \draw[rel] (h2.21.7) -- node[fl,sloped,below,pos=0.62] {$\pi^{o_1,\kappa}_{h_2,o_1}=12$} (y1.221.7);

    \end{tikzpicture}
    \caption{LRPN$_{2,-1}$ relevances for the output $o_1$, on the network of
    Figure~\ref{fig:DAG example} at $x=(-2,1)$, for either selection
    $\kappa\in\mathcal M(x)$; at $h_2$, which has no inhibitory
    contribution, the rule falls back to LRP$_{z^+}$.}
    \label{fig:lrp-ab-example}
\end{figure}

Case (iii) requires every node to mix excitatory and inhibitory
contributions, an assumption the LRP$_{\alpha,\beta}$ literature leaves
implicit; without it, the classical rule does not conserve relevance.
Where $\tilde z^-_v=0$, the inhibitory term of (iii) is undefined;
implementations drop it, keeping the reference
$\rho_{u,v}=\alpha\,z^+_{u,v}/\tilde z^+_v$ and the classical flow
$\pi=\rho\mu$ without normalizing it: its kernel sums to $\alpha$ at $v$, so that $v$ receives $\alpha\mu_v$ and
passes on $\mu_v$: the share $\beta\mu_v$ is created or destroyed, and
the flow leaves $\mathcal A_o(x)$. Symmetrically, where
$\tilde z^+_v=0$, the kernel sums to $\beta$ and the share $\alpha\mu_v$
is created or destroyed. Being proportional to $K=\rho/\tilde\rho$ at
$v$, this flow still has $\mathcal H(\pi\|K)=0$; the minimization over
$\mathcal A_o(x)$ discards it and selects LRPN$_{\alpha,\beta}$ with
$(\alpha_v,\beta_v)=(\alpha,\beta)$, whose reference at such a node is
the non-empty channel normalized, $z^+_{u,v}/\tilde z^+_v$ (resp.\
$z^-_{u,v}/\tilde z^-_v$): it sums to $1$ and does not depend on
$(\alpha,\beta)$. Relevance is then conserved,
and LRPN$_{\alpha,\beta}$ falls back to LRP$_{z^+}$ (resp.\ to its inhibitory counterpart)
at such nodes. In
Figure~\ref{fig:lrp-ab-example}, at $x=(-2,1)$, $h_2$ receives only the
contributions $2$, $1$, $1$: the classical rule LRP$_{2,-1}$ sends $24$ to
its parents and the input total
becomes $18$, whereas LRPN$_{2,-1}$ splits $\mu_{h_2}=12$ as $6$, $3$, $3$, and the
inputs receive $0$, $3$, $3$, adding up to $x_{o_1}=6$.

At a ReLU kink the node values do not branch, so that the relevance is a
single vector; at a tie of a max node they do not branch either, but the
winning edge, hence the reference, does. The behaviour of LRP$_0$,
LRP$_{z^+}$ and LRPN$_{\alpha,\beta}$ near a kink and near a tie is given by
Corollary~\ref{thm:RP0ReLU}(iii) and (iv).

In the following, we fix a single active output $o\in\mathcal O$ and
omit it from the superscripts for readability; the dependence on $o$
remains explicit through $\mathcal P(u,o)$ wherever it appears.

\begin{corollary}
    \label{thm:RP0ReLU}
    Let $\mathcal N_{\mathcal F}(\mathcal G)$ be a feed-forward Filippov
    neural network whose activations are all linear, ReLU or max, a max
    node $v$ taking the value $x_v=\max_{u:u\to v}w_{u,v}x_u$,
    $x\in\mathbb R^d$ an input at which the node values do not branch (see
    Theorem~\ref{thm:ffnn-output}(iv)), $o\in\mathcal O$ an active output,
    and $\kappa\in\mathcal M(x)$, which picks at every max node $v$ a
    predecessor $\kappa_v$ with $z_{\kappa_v,v}=x_v$. With
    $\tilde z_v:=\sum_{u:u\to v}w_{u,v}x_u$ over all the edges of
    $\mathcal G$ entering $v$, let the gate $\eta_{u,v}(x)$ of an edge be
    $1$ if $v$ is linear, ReLU with $\tilde z_v>0$, or max with
    $u=\kappa_v$, and $0$ otherwise, $W_u(x)$ the sum over the paths of
    $\mathcal P(u,o)$ of the products of $w_{a,b}\,\eta_{a,b}(x)$ along
    them, and $W^v_u(x)$ the same sum over $\mathcal P(u,v)$ without the
    gate of the last edge. Let $\rho$ be the
    reference of LRP$_0$, LRP$_{z^+}$ or LRPN$_{\alpha,\beta}$, with weights
    continuous in $x$ (see Corollary~\ref{cor:lrp-references}(i), (ii),
    (iv)), on the edges entering the linear and ReLU nodes, and
    $\rho_{u,v}:=z_{u,v}$ if $u=\kappa_v$, $0$ otherwise, on the edges
    entering a max node $v$, at inputs where the hypotheses of that rule
    hold. A point $\bar x$ is a \emph{simple switching point} of a node $v$
    if $\tilde z_r(\bar x)\neq0$ at every linear or ReLU hidden node
    $r\neq v$, and every max node $r\neq v$ has at $\bar x$ a single
    maximizer, or the maximum $0$ attained only at ReLU nodes $r'$ with
    $\tilde z_{r'}(\bar x)\neq0$.

    \smallskip\noindent
    (i) On $V(x)\setminus(\mathcal O\setminus\{o\})$, $\mu_u=x_u\,W^\rho_u$,
    where $W^\rho_o=1$ and
    $W^\rho_u=\sum_{v\in H(x)\cup\{o\}:\,u\to v}\omega_{u,v}\,s_v\,W^\rho_v$,
    with $\omega_{u,v}:=\rho_{u,v}/x_u$ and $s_v:=x_v/\tilde\rho_v$. For LRP$_0$,
    if $\tilde z_v\neq0$ at every linear hidden node $v$, $W^\rho=W$, and
    $W_u(x)=\partial x_o/\partial x_u(x)$ if $x\notin\mathcal D$
    (\emph{Gradient$\times$Input}).

    \smallskip\noindent
    (ii) On an open set of inputs where the signs of the node values and of
    the contributions are constant, and so is $\kappa$, $\mu$ is
    continuous, and, for LRP$_0$ with $\tilde z_v\neq0$ at every linear
    hidden node $v$, $\mu_u=x_u\,W_u$ with $W$ constant. If the hidden nodes
    are ReLU, or max of non-negative value, and $x_o>0$, LRP$_{z^+}$
    satisfies $0\le\mu_u\le|x_u|\,\bar W_u$, $\bar W_u$ being the same sum
    as $W_u$ with $|w_{a,b}|$ in place of $w_{a,b}\,\eta_{a,b}$.

    \smallskip\noindent
    (iii) Let $\bar x$ be a simple switching point of a ReLU node $v$ with
    $\tilde z_v(\bar x)=0$, $\tilde z_v$ taking both signs near $\bar x$,
    and $v$ the maximizer of no max node, and let $\mu^\mp$ be the limits
    of $\mu(x)$ as $x\to\bar x$ with $\tilde z_v(x)<0$, resp.\
    $\tilde z_v(x)>0$. Then $\mu^-=\mu(\bar x)$, and
    $\Delta:=\mu^+-\mu(\bar x)$ satisfies $\sum_{u\in\mathcal I}\Delta_u=0$,
    with: (a) for LRP$_0$, $\Delta_u=\bar x_u\,W^v_u(\bar x)\,W_v(\bar x)$;
    (b) for LRPN$_{\alpha,\beta}$, if $\bar x_i\neq0$ at every input $i$,
    $\Delta=0$ unless some successor $w$ of $v$ other than a max node has
    at $\bar x$ no contribution of the sign of $w_{v,w}$, and $\Delta=0$
    for LRP$_{z^+}$.

    \smallskip\noindent
    (iv) Let $\bar x$ be a simple switching point of a max node $v$ whose
    maximum $x_v(\bar x)\neq0$ is attained at exactly two predecessors
    $u$, $u'$, with $z_{u,v}-z_{u',v}$ taking both signs near $\bar x$; let
    $\kappa,\kappa'\in\mathcal M(\bar x)$ pick $u$, resp.\ $u'$, at $v$ and
    the same predecessors elsewhere, and $R^{(a)}_i$ be the sum over
    $\mathcal P(i,a)$ of the products of the kernels of the rule along
    $\kappa$ at $\bar x$. Then $\mu(x)$ tends to $\mu^{\kappa}(\bar x)$ as
    $x\to\bar x$ with $z_{u,v}(x)>z_{u',v}(x)$, to $\mu^{\kappa'}(\bar x)$
    with $z_{u',v}(x)>z_{u,v}(x)$, and $\Delta:=\mu^{\kappa'}(\bar
    x)-\mu^{\kappa}(\bar x)$ satisfies $\sum_{i\in\mathcal I}\Delta_i=0$, with:
    (a) for LRP$_0$, $\Delta_i=\bar x_i\,W_v(\bar x)\bigl(w_{u',v}\hat
    W^{u'}_i(\bar x)-w_{u,v}\hat W^{u}_i(\bar x)\bigr)$, $\hat W^a_i$ being
    the sum over $\mathcal P(i,a)$ of the products of $w\eta$ along
    $\kappa$; (b) for LRP$_{z^+}$ and LRPN$_{\alpha,\beta}$, if
    $\bar x_i\neq0$ at every input $i$,
    $\Delta_i=\mu_v(\bar x)\bigl(R^{(u')}_i-R^{(u)}_i\bigr)$.
\end{corollary}
Proof of the corollary is provided in Appendix~\ref{proof:thm:RP0ReLU}.

For LRP$_0$, the recursion of (i) is the adjoint recursion
\eqref{eq:adjoint-recursion}, the gates being its branch Jacobians; the
hypothesis on linear nodes is sharp, a linear node with $\tilde z_v=0$
leaving $G(x)$ but keeping the gate $1$. For the other rules, the gate
becomes the \emph{secant} ratio $s_v$ -- for LRP$_{z^+}$, the share
$\tilde z_v/\tilde z^+_v\in(0,1]$ of the excitatory input of $v$ that
survives the inhibition --, so that $W^\rho$ moves with $x$ inside a
region of constant pattern. The relevance of an input stays at most
proportional to its value for LRP$_{z^+}$, whereas LRPN$_{\alpha,\beta}$
only bounds the absolute mass of the inputs by $|x_o|\,c^L$, with
$c=\max_v(|\alpha_v|+|\beta_v|)$ and $L$ the length of the longest path:
the amplification met in Section~\ref{sec:app-images}.

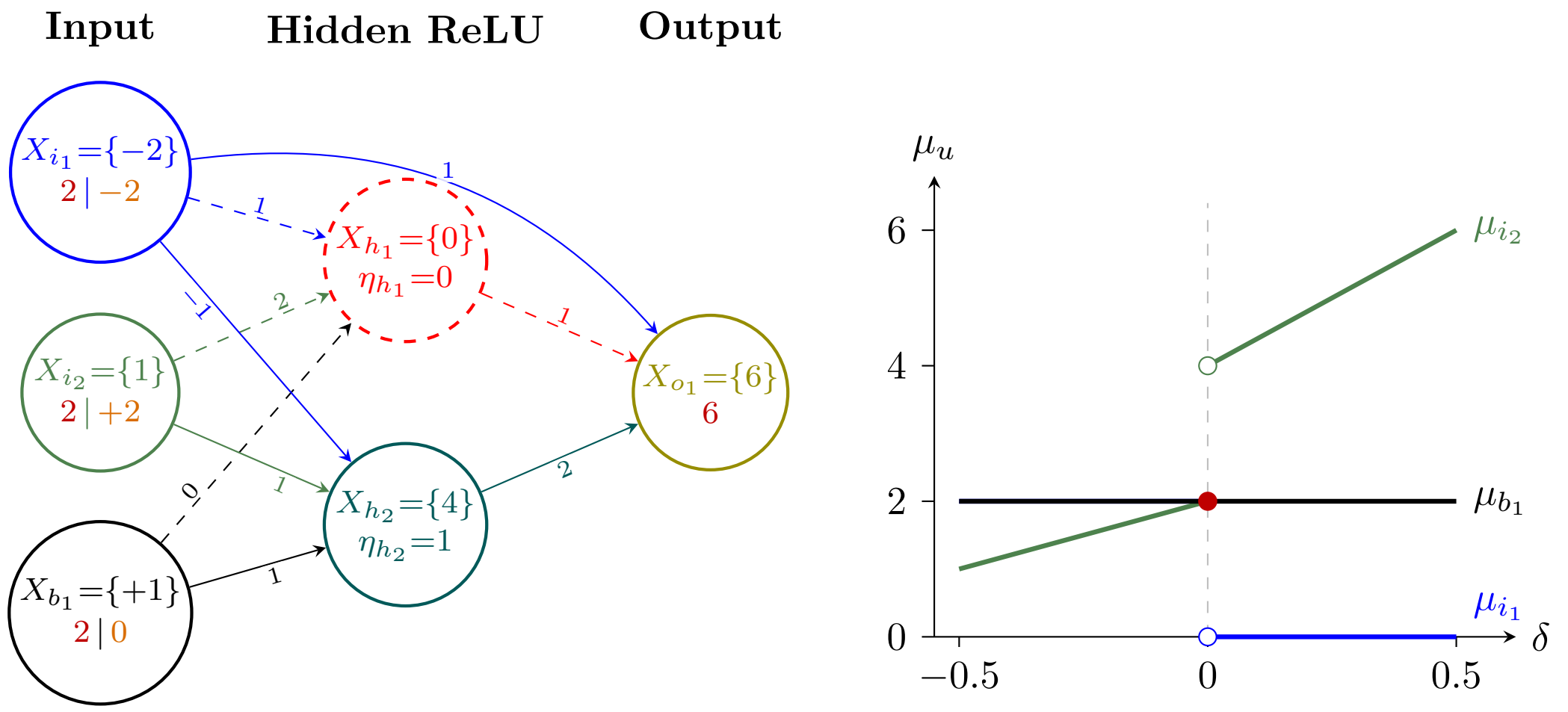
\begin{figure}[H]
    \centering
    \begin{tikzpicture}[
        x=2.7cm,
        y=1.95cm,
        >=stealth,
        neuron/.style={circle,draw,font=\scriptsize,inner sep=0.5pt,minimum size=0.8cm,thick},
        wl/.style={inner sep=1pt,font=\scriptsize,fill=white,fill opacity=0.85,text opacity=1},
        every node/.style={font=\small}
    ]
    \node[neuron,blue] (x1) at (0,2)
        {\shortstack{$X_{i_1}{=}\{-2\}$\\[-1pt]{\color{red!75!black}$2$}\,$|$\,{\color{orange!85!black}$-2$}}};
    \node[neuron,green!30!black!70] (x2) at (0,1)
        {\shortstack{$X_{i_2}{=}\{1\}$\\[-1pt]{\color{red!75!black}$2$}\,$|$\,{\color{orange!85!black}$+2$}}};
    \node[neuron] (b1) at (0,0)
        {\shortstack{$X_{b_1}{=}\{+1\}$\\[-1pt]{\color{red!75!black}$2$}\,$|$\,{\color{orange!85!black}$0$}}};
    \node[neuron,red,dashed] (h1) at (1,1.6)
        {\shortstack{$X_{h_1}{=}\{0\}$\\[-1pt]$\eta_{h_1}{=}0$}};
    \node[neuron,teal!70!black] (h2) at (1,0.4)
        {\shortstack{$X_{h_2}{=}\{4\}$\\[-1pt]$\eta_{h_2}{=}1$}};
    \node[neuron,olive] (y1) at (2,1)
        {\shortstack{$X_{o_1}{=}\{6\}$\\[-1pt]{\color{red!75!black}$6$}}};
    \node at (0,2.65) {\textbf{Input}};
    \node at (1,2.65) {\textbf{Hidden ReLU}};
    \node at (2,2.65) {\textbf{Output}};

    \draw[->,blue,dashed] (x1) -- node[wl,above,sloped,pos=0.5,text=blue] {$\scriptstyle 1$} (h1);
    \draw[->,green!30!black!70,dashed] (x2) -- node[wl,above,sloped,pos=0.72,text=green!30!black!70] {$\scriptstyle 2$} (h1);
    \draw[->,dashed] (b1) -- node[wl,above,sloped,pos=0.2] {$\scriptstyle 0$} (h1);
    \draw[->,blue] (x1) -- node[wl,below,sloped,pos=0.24,text=blue] {$\scriptstyle -1$} (h2);
    \draw[->,green!30!black!70] (x2) -- node[wl,below,sloped,pos=0.72,text=green!30!black!70] {$\scriptstyle 1$} (h2);
    \draw[->] (b1) -- node[wl,below,sloped,pos=0.6] {$\scriptstyle 1$} (h2);
    \draw[->,red,dashed] (h1) -- node[wl,above,sloped,pos=0.5,text=red] {$\scriptstyle 1$} (y1);
    \draw[->,teal!70!black] (h2) -- node[wl,below,sloped,pos=0.5,text=teal!70!black] {$\scriptstyle 2$} (y1);
    \draw[->,blue] (x1) to[bend left=28] node[wl,above,pos=0.5,text=blue] {$\scriptstyle 1$} (y1);
    \end{tikzpicture}
    \hspace{0.4cm}
    \begin{tikzpicture}[x=4.4cm,y=0.6cm,>=stealth,every node/.style={font=\small}]
    \draw[->] (-0.55,0) -- (0.62,0) node[right] {$\delta$};
    \draw[->] (-0.55,0) -- (-0.55,6.8) node[above] {$\mu_u$};
    \foreach \t in {-0.5,0,0.5} \draw (\t,0) -- (\t,-0.15) node[below] {$\t$};
    \foreach \t in {0,2,4,6} \draw (-0.55,\t) -- (-0.575,\t) node[left] {$\t$};
    \draw[gray!50,dashed] (0,0) -- (0,6.4);
    \draw[very thick,blue] (-0.5,2) -- (0,2);
    \draw[very thick,blue] (0,0) -- (0.5,0) node[above right] {$\mu_{i_1}$};
    \draw[very thick,green!30!black!70] (-0.5,1) -- (0,2);
    \draw[very thick,green!30!black!70] (0,4) -- (0.5,6) node[right] {$\mu_{i_2}$};
    \draw[very thick] (-0.5,2) -- (0.5,2) node[right] {$\mu_{b_1}$};
    \fill[red!75!black] (0,2) circle (2.4pt);
    \draw[blue,fill=white] (0,0) circle (2.2pt);
    \draw[green!30!black!70,fill=white] (0,4) circle (2.2pt);
    \end{tikzpicture}
    \caption{Corollary~\ref{thm:RP0ReLU}(iii)(a) at the switching point
    $\bar x=(-2,1)$ of Figure~\ref{fig:DAG example}, for $o_1$. Left: the
    weights $w_{u,v}$ of the full graph; the kink node $h_1$ and its
    edges (dashed) lie outside $G(\bar x)$. At each input, the LRP$_0$
    relevance $\mu_u(\bar x)$ (red) and the jump $\Delta_u$ (orange) met when
    $h_1$ turns on. Right: the input relevances along
    $x=(-2,1+\delta)$; at $\delta=0$, they equal the limit from
    $\delta<0$ (red dot), and jump by $\Delta$ for $\delta>0$ (open dots).}
    \label{fig:kink-relevance}
\end{figure}

\begin{figure}[H]
    \centering
    \begin{minipage}[t]{0.325\textwidth}
    \centering
    \begin{tikzpicture}[x=1.95cm,y=1.05cm,>=stealth,nn/.style={circle,draw,thick,minimum size=0.62cm,inner sep=0pt,font=\scriptsize},wl/.style={inner sep=0.5pt,font=\scriptsize,fill=white,fill opacity=0.85,text opacity=1}]
    \node[nn,blue] (i1) at (0,2) {$i_1$};
    \node[nn,green!30!black!70] (i2) at (0,1) {$i_2$};
    \node[nn] (b1) at (0,0) {$b_1$};
    \node[nn,red,dashed] (h1) at (1,1.6) {$h_1$};
    \node[nn,teal!70!black] (h2) at (1,0.4) {$h_2$};
    \node[nn,olive] (o) at (2,1) {$o_1$};
    \draw[->,blue] (i1) -- node[wl,above,sloped,pos=0.5,text=blue] {$\scriptstyle 1$} (h1);
    \draw[->,green!30!black!70] (i2) -- node[wl,above,sloped,pos=0.7,text=green!30!black!70] {$\scriptstyle 2$} (h1);
    \draw[->] (b1) -- node[wl,above,sloped,pos=0.25] {$\scriptstyle 0$} (h1);
    \draw[->,blue] (i1) -- node[wl,below,sloped,pos=0.25,text=blue] {$\scriptstyle -1$} (h2);
    \draw[->,green!30!black!70] (i2) -- node[wl,below,sloped,pos=0.7,text=green!30!black!70] {$\scriptstyle 1$} (h2);
    \draw[->] (b1) -- node[wl,below,sloped,pos=0.6] {$\scriptstyle 1$} (h2);
    \draw[->,red,dashed] (h1) -- node[wl,above,sloped,pos=0.5,text=red] {$\scriptstyle 1$} (o);
    \draw[->,teal!70!black] (h2) -- node[wl,below,sloped,pos=0.5,text=teal!70!black] {$\scriptstyle 2$} (o);
    \draw[->,blue] (i1) to[bend left=30] node[wl,above,pos=0.5,text=blue] {$\scriptstyle 1$} (o);
    \end{tikzpicture}

    \smallskip
    \begin{tikzpicture}[x=3.6cm,y=0.42cm,>=stealth,every node/.style={font=\small}]
    \draw[->] (-0.55,-4) -- (0.62,-4) node[right] {$\delta$};
    \draw[->] (-0.55,-4) -- (-0.55,8.6) node[above] {$\mu_u$};
    \foreach \t in {-0.5,0,0.5} \draw (\t,-4) -- (\t,-4.151) node[below] {$\scriptstyle\t$};
    \draw (-0.55,-3) -- (-0.575,-3) node[left] {$\scriptstyle -3$}; \draw (-0.55,0) -- (-0.575,0) node[left] {$\scriptstyle 0$}; \draw (-0.55,3) -- (-0.575,3) node[left] {$\scriptstyle 3$}; \draw (-0.55,6) -- (-0.575,6) node[left] {$\scriptstyle 6$};
    \draw[gray!50,dashed] (0,-4) -- (0,8.6-0.4);
    \draw[very thick,blue] plot coordinates {(-0.50,2.857) (-0.45,2.873) (-0.40,2.889) (-0.35,2.904) (-0.30,2.919) (-0.25,2.933) (-0.20,2.947) (-0.15,2.961) (-0.10,2.974) (-0.05,2.987) (0.00,3.000)};
    \draw[very thick,blue] plot coordinates {(0.00,3.000) (0.05,3.024) (0.10,3.048) (0.15,3.070) (0.20,3.091) (0.25,3.111) (0.30,3.130) (0.35,3.149) (0.40,3.167) (0.45,3.184) (0.50,3.200)};
    \draw[very thick,green!30!black!70] plot coordinates {(-0.50,0.714) (-0.45,0.790) (-0.40,0.867) (-0.35,0.944) (-0.30,1.022) (-0.25,1.100) (-0.20,1.179) (-0.15,1.258) (-0.10,1.338) (-0.05,1.419) (0.00,1.500)};
    \draw[very thick,green!30!black!70] plot coordinates {(0.00,1.500) (0.05,1.663) (0.10,1.829) (0.15,1.995) (0.20,2.164) (0.25,2.333) (0.30,2.504) (0.35,2.677) (0.40,2.850) (0.45,3.024) (0.50,3.200)};
    \draw[very thick,black] plot coordinates {(-0.50,1.429) (-0.45,1.437) (-0.40,1.444) (-0.35,1.452) (-0.30,1.459) (-0.25,1.467) (-0.20,1.474) (-0.15,1.481) (-0.10,1.487) (-0.05,1.494) (0.00,1.500)};
    \draw[very thick,black] plot coordinates {(0.00,1.500) (0.05,1.512) (0.10,1.524) (0.15,1.535) (0.20,1.545) (0.25,1.556) (0.30,1.565) (0.35,1.574) (0.40,1.583) (0.45,1.592) (0.50,1.600)};
    \draw[very thick,dashed,blue] plot coordinates {(-0.50,0.714) (-0.45,0.646) (-0.40,0.578) (-0.35,0.508) (-0.30,0.438) (-0.25,0.367) (-0.20,0.295) (-0.15,0.222) (-0.10,0.149) (-0.05,0.075) (0.00,0.000)};
    \draw[very thick,dashed,blue] plot coordinates {(0.00,0.000) (0.05,-0.302) (0.10,-0.610) (0.15,-0.921) (0.20,-1.236) (0.25,-1.556) (0.30,-1.878) (0.35,-2.204) (0.40,-2.533) (0.45,-2.865) (0.50,-3.200)};
    \draw[very thick,dashed,green!30!black!70] plot coordinates {(-0.50,1.429) (-0.45,1.580) (-0.40,1.733) (-0.35,1.888) (-0.30,2.043) (-0.25,2.200) (-0.20,2.358) (-0.15,2.517) (-0.10,2.677) (-0.05,2.838) (0.00,3.000)};
    \draw[very thick,dashed,green!30!black!70] plot coordinates {(0.00,3.000) (0.05,3.478) (0.10,3.962) (0.15,4.451) (0.20,4.945) (0.25,5.444) (0.30,5.948) (0.35,6.455) (0.40,6.967) (0.45,7.482) (0.50,8.000)};
    \draw[very thick,dashed,black] plot coordinates {(-0.50,2.857) (-0.45,2.873) (-0.40,2.889) (-0.35,2.904) (-0.30,2.919) (-0.25,2.933) (-0.20,2.947) (-0.15,2.961) (-0.10,2.974) (-0.05,2.987) (0.00,3.000)};
    \draw[very thick,dashed,black] plot coordinates {(0.00,3.000) (0.05,3.024) (0.10,3.048) (0.15,3.070) (0.20,3.091) (0.25,3.111) (0.30,3.130) (0.35,3.149) (0.40,3.167) (0.45,3.184) (0.50,3.200)};
    \fill[blue] (0,3.000) circle (1.6pt);
    \fill[green!30!black!70] (0,1.500) circle (1.6pt);
    \fill[black] (0,1.500) circle (1.6pt);
    \fill[blue] (0,0.000) circle (1.6pt);
    \fill[green!30!black!70] (0,3.000) circle (1.6pt);
    \fill[black] (0,3.000) circle (1.6pt);
    \end{tikzpicture}

    {\small (a) (iii)(b), generic case}
    \end{minipage}\hfill
    \begin{minipage}[t]{0.325\textwidth}
    \centering
    \begin{tikzpicture}[x=1.95cm,y=1.05cm,>=stealth,nn/.style={circle,draw,thick,minimum size=0.62cm,inner sep=0pt,font=\scriptsize},wl/.style={inner sep=0.5pt,font=\scriptsize,fill=white,fill opacity=0.85,text opacity=1}]
    \node[nn,blue] (i1) at (0,2) {$i_1$};
    \node[nn,green!30!black!70] (i2) at (0,1) {$i_2$};
    \node[nn] (b1) at (0,0) {$b_1$};
    \node[nn,red,dashed] (h1) at (1,1.6) {$h_1$};
    \node[nn,teal!70!black] (h2) at (1,0.4) {$h_2$};
    \node[nn,olive] (o) at (2,1) {$o_1$};
    \draw[->,blue] (i1) -- node[wl,above,sloped,pos=0.5,text=blue] {$\scriptstyle 1$} (h1);
    \draw[->,green!30!black!70] (i2) -- node[wl,above,sloped,pos=0.7,text=green!30!black!70] {$\scriptstyle 2$} (h1);
    \draw[->] (b1) -- node[wl,above,sloped,pos=0.25] {$\scriptstyle 0$} (h1);
    \draw[->,blue] (i1) -- node[wl,below,sloped,pos=0.25,text=blue] {$\scriptstyle -1$} (h2);
    \draw[->,green!30!black!70] (i2) -- node[wl,below,sloped,pos=0.7,text=green!30!black!70] {$\scriptstyle 1$} (h2);
    \draw[->] (b1) -- node[wl,below,sloped,pos=0.6] {$\scriptstyle 1$} (h2);
    \draw[->,red,dashed,line width=1.2pt] (h1) -- node[wl,above,sloped,pos=0.5,text=red] {$\scriptstyle -1$} (o);
    \draw[->,teal!70!black] (h2) -- node[wl,below,sloped,pos=0.5,text=teal!70!black] {$\scriptstyle 2$} (o);
    \draw[->,blue,line width=1.2pt] (i1) to[bend left=30] node[wl,above,pos=0.5,text=blue] {$\scriptstyle -1$} (o);
    \end{tikzpicture}

    \smallskip
    \begin{tikzpicture}[x=3.6cm,y=0.125cm,>=stealth,every node/.style={font=\small}]
    \draw[->] (-0.55,-18) -- (0.62,-18) node[right] {$\delta$};
    \draw[->] (-0.55,-18) -- (-0.55,24) node[above] {$\mu_u$};
    \foreach \t in {-0.5,0,0.5} \draw (\t,-18) -- (\t,-18.504) node[below] {$\scriptstyle\t$};
    \draw (-0.55,-15) -- (-0.575,-15) node[left] {$\scriptstyle -15$}; \draw (-0.55,0) -- (-0.575,0) node[left] {$\scriptstyle 0$}; \draw (-0.55,10) -- (-0.575,10) node[left] {$\scriptstyle 10$}; \draw (-0.55,20) -- (-0.575,20) node[left] {$\scriptstyle 20$};
    \draw[gray!50,dashed] (0,-18) -- (0,24-0.4);
    \draw[very thick,blue] plot coordinates {(-0.50,6.000) (-0.45,6.000) (-0.40,6.000) (-0.35,6.000) (-0.30,6.000) (-0.25,6.000) (-0.20,6.000) (-0.15,6.000) (-0.10,6.000) (-0.05,6.000) (0.00,6.000)};
    \draw[very thick,blue] plot coordinates {(0.00,6.000) (0.05,5.941) (0.10,5.882) (0.15,5.825) (0.20,5.769) (0.25,5.714) (0.30,5.660) (0.35,5.607) (0.40,5.556) (0.45,5.505) (0.50,5.455)};
    \draw[very thick,green!30!black!70] plot coordinates {(-0.50,1.000) (-0.45,1.100) (-0.40,1.200) (-0.35,1.300) (-0.30,1.400) (-0.25,1.500) (-0.20,1.600) (-0.15,1.700) (-0.10,1.800) (-0.05,1.900) (0.00,2.000)};
    \draw[very thick,green!30!black!70] plot coordinates {(0.00,2.000) (0.05,2.079) (0.10,2.157) (0.15,2.233) (0.20,2.308) (0.25,2.381) (0.30,2.453) (0.35,2.523) (0.40,2.593) (0.45,2.661) (0.50,2.727)};
    \draw[very thick,black] plot coordinates {(-0.50,2.000) (-0.45,2.000) (-0.40,2.000) (-0.35,2.000) (-0.30,2.000) (-0.25,2.000) (-0.20,2.000) (-0.15,2.000) (-0.10,2.000) (-0.05,2.000) (0.00,2.000)};
    \draw[very thick,black] plot coordinates {(0.00,2.000) (0.05,1.980) (0.10,1.961) (0.15,1.942) (0.20,1.923) (0.25,1.905) (0.30,1.887) (0.35,1.869) (0.40,1.852) (0.45,1.835) (0.50,1.818)};
    \draw[very thick,dashed,blue] plot coordinates {(-0.50,6.000) (-0.45,6.000) (-0.40,6.000) (-0.35,6.000) (-0.30,6.000) (-0.25,6.000) (-0.20,6.000) (-0.15,6.000) (-0.10,6.000) (-0.05,6.000) (0.00,6.000)};
    \draw[very thick,dashed,blue] plot coordinates {(0.00,22.000) (0.05,21.881) (0.10,21.765) (0.15,21.650) (0.20,21.538) (0.25,21.429) (0.30,21.321) (0.35,21.215) (0.40,21.111) (0.45,21.009) (0.50,20.909)};
    \draw[very thick,dashed,green!30!black!70] plot coordinates {(-0.50,1.000) (-0.45,1.100) (-0.40,1.200) (-0.35,1.300) (-0.30,1.400) (-0.25,1.500) (-0.20,1.600) (-0.15,1.700) (-0.10,1.800) (-0.05,1.900) (0.00,2.000)};
    \draw[very thick,dashed,green!30!black!70] plot coordinates {(0.00,-16.000) (0.05,-15.842) (0.10,-15.686) (0.15,-15.534) (0.20,-15.385) (0.25,-15.238) (0.30,-15.094) (0.35,-14.953) (0.40,-14.815) (0.45,-14.679) (0.50,-14.545)};
    \draw[very thick,dashed,black] plot coordinates {(-0.50,2.000) (-0.45,2.000) (-0.40,2.000) (-0.35,2.000) (-0.30,2.000) (-0.25,2.000) (-0.20,2.000) (-0.15,2.000) (-0.10,2.000) (-0.05,2.000) (0.00,2.000)};
    \draw[very thick,dashed,black] plot coordinates {(0.00,4.000) (0.05,3.960) (0.10,3.922) (0.15,3.883) (0.20,3.846) (0.25,3.810) (0.30,3.774) (0.35,3.738) (0.40,3.704) (0.45,3.670) (0.50,3.636)};
    \fill[blue] (0,6.000) circle (1.6pt);
    \fill[green!30!black!70] (0,2.000) circle (1.6pt);
    \fill[black] (0,2.000) circle (1.6pt);
    \fill[blue] (0,6.000) circle (1.6pt);
    \draw[blue,fill=white] (0,22.000) circle (1.6pt);
    \fill[green!30!black!70] (0,2.000) circle (1.6pt);
    \draw[green!30!black!70,fill=white] (0,-16.000) circle (1.6pt);
    \fill[black] (0,2.000) circle (1.6pt);
    \draw[black,fill=white] (0,4.000) circle (1.6pt);
    \end{tikzpicture}

    {\small (b) (iii)(b), exceptional case}
    \end{minipage}\hfill
    \begin{minipage}[t]{0.325\textwidth}
    \centering
    \begin{tikzpicture}[x=1.95cm,y=1.05cm,>=stealth,nn/.style={circle,draw,thick,minimum size=0.62cm,inner sep=0pt,font=\scriptsize},wl/.style={inner sep=0.5pt,font=\scriptsize,fill=white,fill opacity=0.85,text opacity=1}]
    \node[nn,blue] (i1) at (0,1.6) {$i_1$};
    \node[nn,green!30!black!70] (i2) at (0,0.4) {$i_2$};
    \node[nn,purple] (v) at (1,1) {$v$};
    \node[nn,olive] (o) at (2,1) {$o$};
    \node[font=\scriptsize,purple] at (1,1.55) {$\max$};
    \draw[->,blue] (i1) -- node[wl,above,sloped,text=blue] {$\scriptstyle 1$} (v);
    \draw[->,green!30!black!70] (i2) -- node[wl,below,sloped,text=green!30!black!70] {$\scriptstyle 1$} (v);
    \draw[->,purple] (v) -- node[wl,above,text=purple] {$\scriptstyle 1$} (o);
    \end{tikzpicture}

    \smallskip
    \begin{tikzpicture}[x=3.6cm,y=3.0cm,>=stealth,every node/.style={font=\small}]
    \draw[->] (-0.55,0) -- (0.62,0) node[right] {$\delta$};
    \draw[->] (-0.55,0) -- (-0.55,1.75) node[above] {$\mu_u$};
    \foreach \t in {-0.5,0,0.5} \draw (\t,0) -- (\t,-0.021) node[below] {$\scriptstyle\t$};
    \draw (-0.55,0) -- (-0.575,0) node[left] {$\scriptstyle 0$}; \draw (-0.55,1) -- (-0.575,1) node[left] {$\scriptstyle 1$};
    \draw[gray!50,dashed] (0,0) -- (0,1.75-0.4);
    \draw[very thick,blue] (-0.5,1) -- (0,1); \draw[very thick,blue] (0,0) -- (0.5,0);
    \draw[very thick,green!30!black!70] (-0.5,0) -- (0,0); \draw[very thick,green!30!black!70] (0,1) -- (0.5,1.5);
    \draw[line width=3pt,blue,opacity=0.3] (-0.012,0) -- (-0.012,1);
    \draw[line width=3pt,green!30!black!70,opacity=0.3] (0.012,0) -- (0.012,1);
    \end{tikzpicture}

    {\small (c) (iv), change of winner}
    \end{minipage}
    \caption{Corollary~\ref{thm:RP0ReLU}(iii)(b) and (iv): input relevances
    $\mu_{i_1}$ (blue), $\mu_{i_2}$ (green) and $\mu_{b_1}$ (black) along a line
    through a switching point, for LRP$_{z^+}$ (solid) and LRPN$_{2,-1}$
    (dashed); filled dots: the relevance at $\delta=0$, open dots: the
    limit from $\delta>0$ where it differs. (a) The network of
    Figure~\ref{fig:kink-relevance}, along $x=(-2,1+\delta)$: where LRP$_0$
    jumps, both rules are continuous, $o_1$ already having contributions of
    both signs ($\mu_{i_1}$ for LRP$_{z^+}$ and $\mu_{b_1}$ for LRPN$_{2,-1}$
    coincide). (b) The same network with $w_{h_1,o_1}=w_{i_1,o_1}=-1$ (thick):
    at $\bar x=(-2,1)$, $o_1$ has no negative contribution, and $h_1$ brings
    the first one when it turns on; LRPN$_{2,-1}$ jumps from $(6,2,2)$ to
    $(22,-16,4)$, LRP$_{z^+}$ stays continuous, and the two rules coincide
    for $\delta<0$. (c) $x_o=x_v=\max(x_{i_1},x_{i_2})$
    along $x=(1,1+\delta)$: every rule jumps by $\Delta=(-1,1)$, and at
    $\delta=0$ the relevance is the segment between the two selections
    (shaded).}
    \label{fig:kink-cases}
\end{figure}
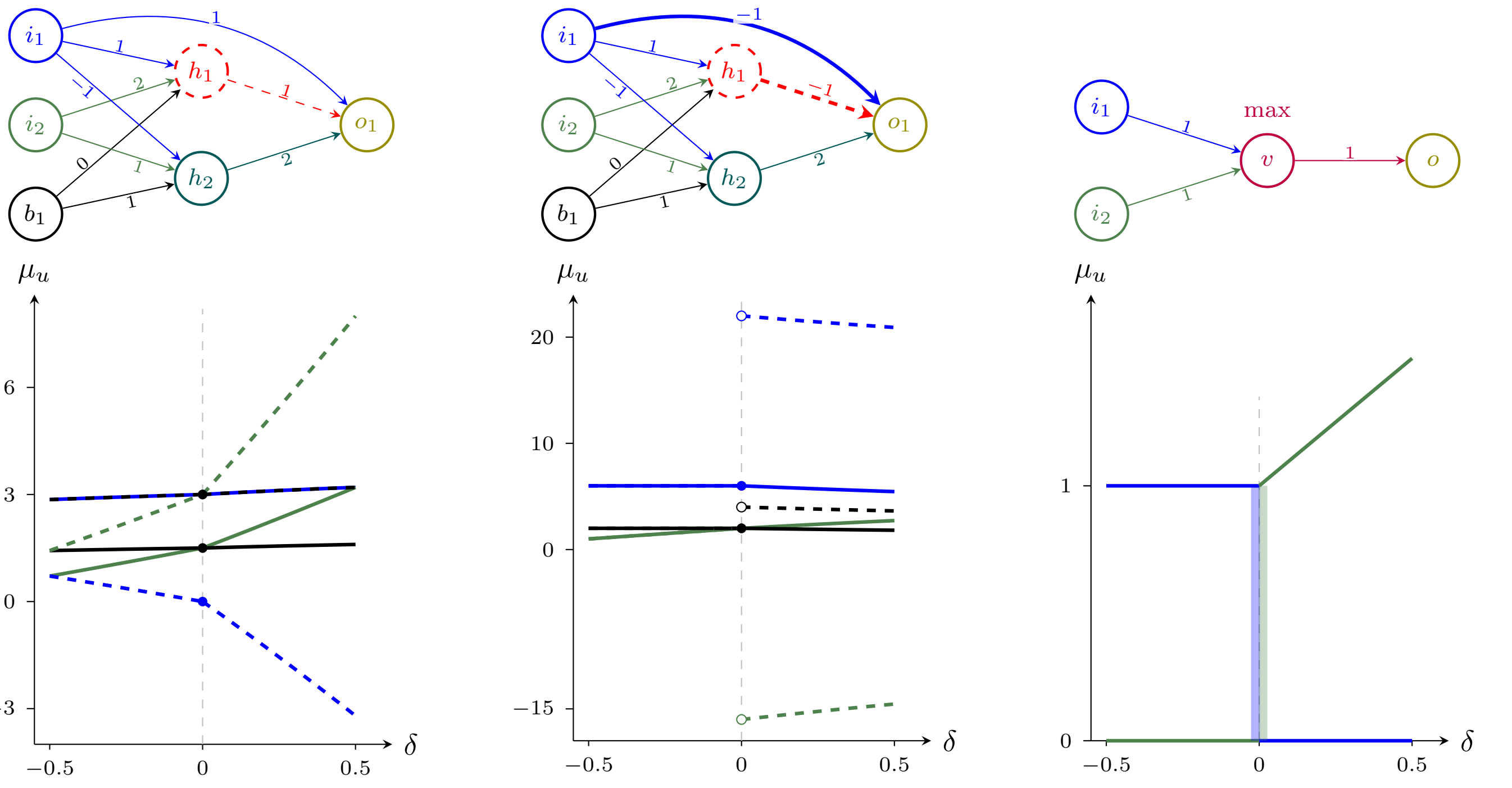

Part (iii) locates the jumps. For LRP$_0$, from the active side, $v$ has
a value $x_v\to0$ but splits $\mu_v=x_vW_v$ in the proportions
$z_{u,v}/x_v$, a product $0\cdot\infty$ of limit $z_{u,v}W_v$: the size
of the jump is set by the contributions that cancel at $v$, not by the
perturbation. For LRP$_{z^+}$ and LRPN$_{\alpha,\beta}$, the kernel at
$v$ stays bounded, and what $v$ receives from a successor $w$ vanishes
with $x_v$, unless $v$ brings to $w$ the first contribution of its sign:
that channel then holds $v$ alone and sends it a fixed share of $\mu_w$,
$\alpha_w$ or $\beta_w$, which is $\beta_w=0$ for LRP$_{z^+}$. On the
running example at $\bar x=(-2,1)$, LRP$_0$ jumps by $\Delta=(-2,2,0)$
(Figure~\ref{fig:kink-relevance}), whereas LRP$_{z^+}$ and LRPN$_{2,-1}$
are continuous (Figure~\ref{fig:kink-cases}(a)), both successors of
$h_1$ already having contributions of both signs; with
$w_{h_1,o_1}=w_{i_1,o_1}=-1$, $o_1$ has no negative contribution at
$\bar x$, and LRPN$_{2,-1}$ jumps by $\Delta=(16,-18,2)$
(Figure~\ref{fig:kink-cases}(b)).

Part (iv) is the counterpart of (iii) at a max node, with the opposite
outcome in general. Whatever the rule, the kernel of a max node is $1$ on its winning
edge and $0$ on the others, so that at a change of winner the whole
relevance $\mu_v$ of the node leaves the paths through $u$ for the paths
through $u'$; $\Delta=0$ if and only if the node carries no relevance,
$\mu_v(\bar x)=0$, or both predecessors pass a unit of relevance on to the
inputs in the same way, $R^{(u)}=R^{(u')}$. At $\bar x$ itself, the
relevance is the segment between $\mu^\kappa(\bar x)$ and
$\mu^{\kappa'}(\bar x)$ (Theorem~\ref{thm:RP0}). For $x_o=x_v$ and
$x_v=\max(x_{i_1},x_{i_2})$, at $\bar x_{i_1}=\bar x_{i_2}=a\neq0$, every
rule gives $\mu^\kappa=(a,0)$, $\mu^{\kappa'}=(0,a)$ and $\Delta=(-a,a)$
(Figure~\ref{fig:kink-cases}(c)). The
second alternative at the other max nodes, in (iii) and (iv), is a
max-pooling window whose ReLU inputs are all inactive, as in VGG-16: its
maximum $0$ is attained several times, but the window stays inactive near
$\bar x$, and its winner does not matter.

\subsection{Sinkhorn Relevance Propagation}
\label{sec:SRP}

The minimizer of Theorem~\ref{thm:RP0} rescales the reference at the head
of each edge, $\pi_{u,v}=\rho_{u,v}\,\mu_v/\tilde\rho_v$: it compares
transition laws, node by node. Comparing instead the flow itself with
$\rho$, edge by edge, gives another problem, whose solution rescales
$\rho$ at both ends of each edge, through one potential per node,
computed by Sinkhorn iterations \citep{Cuturi2013}. A reference of either
sign is split into its sign parts, $\rho=\rho^+-\rho^-$, and we look for
two non-negative flows whose difference is conservative. The positive
part carries mass along its edge, the negative part against it: let
$D_\rho(x)$ have an arc $u\to v$ for every edge $(u,v)\in E(x)$ with
$\rho_{u,v}>0$ and an arc $v\to u$ for every edge with $\rho_{u,v}<0$;
let $\bar D_\rho(x)$ merge the inputs into a single node $S$ and add the
arc $o\to S$; let $E^o_\rho(x)$ be the edges whose arc lies in the
strongly connected component of $S$ in $\bar D_\rho(x)$, and
$V^o_\rho(x)$ the nodes of that component other than $S$, together with
the inputs. For $\rho\ge0$, $E^o_\rho(x)$ consists of the edges lying on
a path of positive reference from $\mathcal I(x)$ to $o$.

\begin{theorem}
    \label{thm:srp}
    Let $x$ be an input at which the node values do not branch,
    $o\in\mathcal O$ an active output, $y>0$,
    $\mathcal A_o(x;y)$ the admissible set of
    Definition~\ref{def:admissible-set} with $x_o$ replaced by $y$, and
    $\rho\in\mathbb R^{E(x)}$ vanishing off $E^o_\rho(x)$ and on the edges
    entering the outputs other than $o$, with $o\in V^o_\rho(x)$. Then:

    \smallskip\noindent
    (i) The problem
    \begin{equation}
        \label{eq:srp-problem}
        \min_{\substack{\pi^\pm\ge0\\ \pi^+-\pi^-\in\mathcal A_o(x;y)}}\
        \mathrm{KL}(\pi^+\,\|\,\rho^+)+\mathrm{KL}(\pi^-\,\|\,\rho^-),
        \qquad
        \mathrm{KL}(\pi\,\|\,\rho):=\sum_{(u,v)\in E(x)}\Bigl(\pi_{u,v}\log\frac{\pi_{u,v}}{\rho_{u,v}}
        -\pi_{u,v}+\rho_{u,v}\Bigr),
    \end{equation}
    has a unique minimizer
    $(\pi^{\mathrm S+},\pi^{\mathrm S-})$, and $\pi^{\mathrm S\pm}>0$
    exactly where $\rho^\pm>0$.

    \smallskip\noindent
    (ii) There is a unique family of \emph{potentials}
    $(\varphi_v)_{v\in V^o_\rho(x)}$, with $\varphi_v>0$ and $\varphi_i=1$
    at the inputs, such that
    \begin{equation}
        \label{eq:srp-form}
        \pi^{\mathrm S+}_{u,v}=\rho^+_{u,v}\,\frac{\varphi_v}{\varphi_u},
        \qquad
        \pi^{\mathrm S-}_{u,v}=\rho^-_{u,v}\,\frac{\varphi_u}{\varphi_v},
        \qquad (u,v)\in E^o_\rho(x).
    \end{equation}
    In particular $\prod_{e\in p}\pi^{\mathrm S+}_e/\rho_e=\varphi_v$ for
    every path $p$ of positive reference from an input to $v$.

    \smallskip\noindent
    (iii) Conversely, a pair of the form \eqref{eq:srp-form} is the
    minimizer if and only if the potentials solve
    \begin{equation}
        \label{eq:srp-fixed-point}
        \varphi_v^2\Bigl(\sum_{u}\frac{\rho^+_{u,v}}{\varphi_u}+\sum_{w}\frac{\rho^-_{v,w}}{\varphi_w}\Bigr)
        =\sum_{w}\rho^+_{v,w}\varphi_w+\sum_{u}\rho^-_{u,v}\varphi_u
        \quad(v\in V^o_\rho(x)\cap H(x)),
        \qquad
        \varphi_o\sum_u\frac{\rho^+_{u,o}}{\varphi_u}-\frac1{\varphi_o}\sum_u\rho^-_{u,o}\varphi_u=y ,
    \end{equation}
    the sums running over $E^o_\rho(x)$.

    \smallskip\noindent
    (iv) The \emph{Sinkhorn relevances}
    $\mu^{\mathrm S}_u:=\sum_{w:\,u\to w}\bigl(\pi^{\mathrm S+}_{u,w}-\pi^{\mathrm S-}_{u,w}\bigr)$,
    $u\notin\mathcal O$, satisfy $\sum_{u\in\mathcal I(x)}\mu^{\mathrm S}_u=y$.

    \smallskip\noindent
    (v) Order $V^o_\rho(x)\setminus\mathcal I=\{v_1,\dots,v_m\}$, take
    $\varphi^{(0)}>0$ with $\varphi^{(0)}_i=1$ at the inputs, and obtain
    $\varphi^{(t+1)}$ from $\varphi^{(t)}$ by giving, for $j=1,\dots,m$ in
    turn, to $\varphi_{v_j}$ the positive value for which
    \eqref{eq:srp-fixed-point} holds at $v_j$, the other potentials being
    fixed at their current values. Then $\varphi^{(t)}$ converges to the
    potentials of (ii).
\end{theorem}
Proof of the theorem is provided in Appendix~\ref{proof:thm:srp}.
By default, SRP takes $\rho=z^+$ and $y=x_o$ for $x_o>0$, so that
$\pi^{\mathrm S-}=0$.

Problem \eqref{eq:srp-problem} is an entropically regularized
minimum-cost flow \citep{Peyre2019}. Writing $\rho^\pm_e=e^{-c_e/\varepsilon}$, the
relative entropy is, up to a constant,
$\varepsilon^{-1}\sum_ec_e\pi_e+\sum_e(\pi_e\log\pi_e-\pi_e)$: a linear transport
cost $c$ plus an entropy, minimized over the flows that are conservative at
every hidden node and bring $y$ to $o$. Entropic optimal transport
\citep{Cuturi2013} prescribes instead the two marginals of a bipartite
graph, and its solution
$\pi_{i,j}=a_iK_{i,j}b_j$ rescales the reference by one factor per node, the
factors being computed by alternate rescalings, the Sinkhorn iterations.
Theorem~\ref{thm:srp} has the same structure on a DAG, with free inputs:
every hidden node and $o$ carry a constraint,
hence a potential $\varphi_v$, and, for $\rho\ge0$,
$\pi^{\mathrm S}_{u,v}=\rho_{u,v}\varphi_v/\varphi_u$
rescales each edge at its two ends; the sweeps of (v) are the Sinkhorn
iterations. The Schrödinger bridge of Section~\ref{sec:region} has the
same two-ended form, on paths.

The potentials are the exponentials of the Lagrange multipliers of the
conservation constraints, and each update of (v) maximizes the dual
function of \eqref{eq:srp-problem} in one coordinate. For $\rho\ge0$, it
reads
$\varphi_v=\bigl(\sum_w\rho_{v,w}\varphi_w/\sum_u\rho_{u,v}/\varphi_u\bigr)^{1/2}$
at a hidden node and $\varphi_o=y/\sum_u\rho_{u,o}/\varphi_u$. One sweep costs
$O(|E^o_\rho(x)|)$, like a pass of LRP, and the observed convergence is
geometric, as for the Sinkhorn algorithm \citep{Cuturi2013,Peyre2019}.

\begin{proposition}
    \label{prop:srp-lrp}
    Under the assumptions of Theorem~\ref{thm:srp}, with $\rho\ge0$,
    write $\pi^{\mathrm S}:=\pi^{\mathrm S+}$:
    (i) if the restriction of $\rho$ to $E^o_\rho(x)$ lies in
    $\mathcal A_o(x;y)$, then $\pi^{\mathrm S}$ is this restriction and all
    potentials equal $1$;
    (ii) if every edge of $E^o_\rho(x)$ enters $o$, then
    \[
    \pi^{\mathrm S}_{u,o}=\frac{y\,\rho_{u,o}}
    {\sum_{(r,o)\in E^o_\rho(x)}\rho_{r,o}},
    \]
    which is the flow of Theorem~\ref{thm:RP0} for the same reference,
    with $x_o$ replaced by $y$;
    (iii) in general $\pi^{\mathrm S}$ differs from that flow: for one input
    $i$, two hidden nodes $h_1,h_2$ and the output $o$, with
    $E^o_\rho(x)=\{(i,h_1),(i,h_2),(h_1,o),(h_2,o)\}$,
    \[
    \frac{\pi^{\mathrm S}_{h_1,o}}{\pi^{\mathrm S}_{h_2,o}}
    =\Bigl(\frac{\rho_{i,h_1}\,\rho_{h_1,o}}{\rho_{i,h_2}\,\rho_{h_2,o}}\Bigr)^{1/2},
    \qquad\text{whereas}\qquad
    \frac{\pi_{h_1,o}}{\pi_{h_2,o}}=\frac{\rho_{h_1,o}}{\rho_{h_2,o}}
    \ \text{ for Theorem~\ref{thm:RP0}}.
    \]
\end{proposition}
Proof of the proposition is provided in Appendix~\ref{proof:prop:srp-lrp}.

\begin{figure}[ht]
    \centering
    \tikzset{
        srpfig/.style={x=1.3cm, y=1.3cm, >=stealth, every node/.style={font=\small}},
        neuron/.style={circle,draw,minimum size=0.8cm,thick,inner sep=0pt},
        flow/.style={->,red!75!black},
        lab/.style={font=\scriptsize,inner sep=1pt},
        ann/.style={font=\scriptsize,inner sep=1pt}
    }
    \newcommand{\srpnodes}{%
        \node[neuron,blue] (i) at (0,0) {$i$};
        \node[neuron,green!30!black!70] (h1) at (1.6,1) {$h_1$};
        \node[neuron,green!30!black!70] (h2) at (1.6,-1) {$h_2$};
        \node[neuron,olive] (o) at (3.2,0) {$o$};}
    \begin{minipage}[t]{0.32\textwidth}
        \centering
        \begin{tikzpicture}[srpfig]
        \srpnodes
        \draw[flow,line width=0.6pt] (i) -- node[lab,above,sloped] {$1$} (h1);
        \draw[flow,line width=0.6pt] (i) -- node[lab,below,sloped] {$1$} (h2);
        \draw[flow,line width=2.4pt] (h1) -- node[lab,above,sloped] {$4$} (o);
        \draw[flow,line width=0.6pt] (h2) -- node[lab,below,sloped] {$1$} (o);
        \node[ann,above=2pt of i] {$X_i\!=\!\{1\}$};
        \node[ann,above=1pt of h1] {$X_{h_1}\!=\!\{1\}$};
        \node[ann,below=1pt of h2] {$X_{h_2}\!=\!\{1\}$};
        \node[ann,below=6pt of o] {$X_o\!=\!\{5\}$};
        \end{tikzpicture}

        \small\textbf{Forward reference}\\[1pt]
        \scriptsize edges: $\rho_{u,v}=(w_{u,v}x_u)^+$; nodes: $X_v(x)$\\
        inflow $1$, outflow $4$ at $h_1$
    \end{minipage}\hfill
    \begin{minipage}[t]{0.32\textwidth}
        \centering
        \begin{tikzpicture}[srpfig]
        \srpnodes
        \draw[flow,line width=2.4pt] (i) -- node[lab,above,sloped] {$4$} (h1);
        \draw[flow,line width=0.6pt] (i) -- node[lab,below,sloped] {$1$} (h2);
        \draw[flow,line width=2.4pt] (h1) -- node[lab,above,sloped] {$4$} (o);
        \draw[flow,line width=0.6pt] (h2) -- node[lab,below,sloped] {$1$} (o);
        \node[ann,above=2pt of i] {$\mu_i\!=\!5$};
        \node[ann,above=1pt of h1] {$\mu_{h_1}\!=\!4$};
        \node[ann,below=1pt of h2] {$\mu_{h_2}\!=\!1$};
        \node[ann,below=6pt of o] {$\mu_o\!=\!5$};
        \end{tikzpicture}

        \small\textbf{LRP} (Theorem~\ref{thm:RP0})\\[1pt]
        \scriptsize edges: $\pi_{u,v}=K_{u,v}\,\mu_v$; nodes: $\mu_v$\\
        $K_{u,v}=\rho_{u,v}/\tilde\rho_v$: $K_{h_1,o}=4/5$, $K_{h_2,o}=1/5$
    \end{minipage}\hfill
    \begin{minipage}[t]{0.32\textwidth}
        \centering
        \begin{tikzpicture}[srpfig]
        \srpnodes
        \draw[flow,line width=2.0pt] (i) -- node[lab,above,sloped] {$10/3$} (h1);
        \draw[flow,line width=1.0pt] (i) -- node[lab,below,sloped] {$5/3$} (h2);
        \draw[flow,line width=2.0pt] (h1) -- node[lab,above,sloped] {$10/3$} (o);
        \draw[flow,line width=1.0pt] (h2) -- node[lab,below,sloped] {$5/3$} (o);
        \node[ann,above=2pt of i] {$\varphi_i\!=\!1$};
        \node[ann,above=1pt of h1] {$\varphi_{h_1}\!=\!10/3$};
        \node[ann,below=1pt of h2] {$\varphi_{h_2}\!=\!5/3$};
        \node[ann,below=6pt of o] {$\varphi_o\!=\!25/9$};
        \end{tikzpicture}

        \small\textbf{SRP} (Theorem~\ref{thm:srp})\\[1pt]
        \scriptsize edges: $\pi^{\mathrm S}_{u,v}=\rho_{u,v}\,\varphi_v/\varphi_u$; nodes: $\varphi_v$\\
        $\mu^{\mathrm S}_{h_1}=10/3$, $\mu^{\mathrm S}_{h_2}=5/3$, $\mu^{\mathrm S}_i=5$
    \end{minipage}
    \caption{The example of Proposition~\ref{prop:srp-lrp}(iii)
    ($w_{i,h_1}=w_{i,h_2}=1$, $w_{h_1,o}=4$, $w_{h_2,o}=1$, $x_i=1$), edge
    widths proportional to the values shown: the forward reference (not
    conservative at $h_1$), LRP and SRP, both conserving $x_o=5$.}
    \label{fig:srp-lrp}
\end{figure}
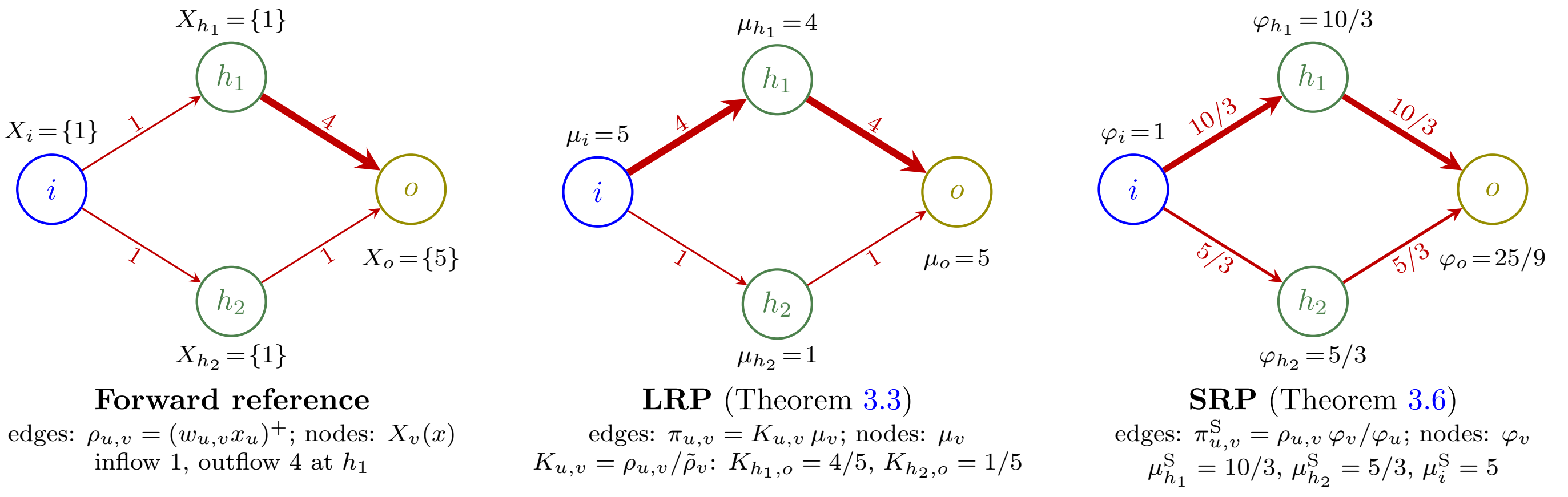

LRP allocates greedily, top-down, splitting the relevance of each node
in proportion to the contributions of its parents; SRP rescales every
path from an input to $o$ by the same factor $\varphi_o$
(Theorem~\ref{thm:srp}(ii)). With $\rho=(1,1,4,1)$ and $y=5$ in
Proposition~\ref{prop:srp-lrp}(iii) (Figure~\ref{fig:srp-lrp}), LRP
sends $4$ through $h_1$, following the last layer only, and SRP $10/3$,
balancing both layers. SRP needs no non-degeneracy condition: for
$\rho\ge0$, a node with $\tilde\rho_v=0$ has no arc entering it and lies
outside $V^o_\rho(x)$. For $x_o<0$ and a reference whose reflection
$-\rho$ satisfies the assumptions of Theorem~\ref{thm:srp}, such as
$\rho^\lambda$ below, which changes sign with $x_o$, we take the
minimizer for $(-\rho,|x_o|)$, on the support $E^o_{-\rho}(x)$, with its
sign changed, and the zero flow for $x_o=0$: exchanging the two sign
parts maps \eqref{eq:srp-problem} for $(\rho,y)$ onto
\eqref{eq:srp-problem} for $(-\rho,-y)$, with flows of opposite sign.
The reflection of $z^+\ge0$ has no path from the inputs to $o$; for
$x_o<0$, SRP with $\rho=z^+$ is then defined as the flow of
$(z^+,|x_o|)$ with its sign changed, a convention rather than a
minimizer for $y=x_o$.

With a signed reference, a negative input relevance
$\mu^{\mathrm S}_i<0$ is mass received back along negative edges --
counter-evidence; restricting $\rho$ to $E^o_\rho(x)$ and to the edges
not entering the other outputs is then a choice of the model, which
keeps the problem on the output explained. The reference must also carry
the scale of the output: the relative entropy keeps the flows close to
the reference in mass, and $z$, whose absolute values add up to the whole
forward activity of the network, is far larger than $x_o$. SRP also
depends on the scale of the hidden units: multiplying a hidden unit by
$c>0$ and its outgoing weights by $1/c$ leaves the network function
unchanged, but multiplies by $c$ the reference on the edges entering the
unit. In Figure~\ref{fig:srp-lrp}, with $h_1$ scaled by $9$, the ratio of
Proposition~\ref{prop:srp-lrp}(iii) becomes $6$ and SRP sends $30/7$
through $h_1$ instead of $10/3$, whereas LRP$_{z^+}$, which normalizes at
the head of each edge, still sends $4$: the Sinkhorn optimum is relative
to the computational representation of the network.

\begin{proposition}
    \label{prop:srp-mass}
    Under the assumptions of Theorem~\ref{thm:srp}, if the positive part
    $\rho^+$ lies in $\mathcal A_o(x;y)$, the Sinkhorn flows have total mass
    $\sum_{(u,v)}\bigl(\pi^{\mathrm S+}_{u,v}+\pi^{\mathrm S-}_{u,v}\bigr)\le e\,R$,
    where $R:=\sum_{(u,v)}|\rho_{u,v}|$.
\end{proposition}
Proof of the proposition is provided in Appendix~\ref{proof:prop:srp-mass}.

The LRP$_{z^+}$ flow $\pi^{z^+}$ of
Corollary~\ref{cor:lrp-references}(ii), of node relevances $\mu^+$, is
conservative at the scale of $x_o$; adding to it the inhibitory
contributions at the same scale gives the reference
\begin{equation}
    \label{eq:srp-lambda}
    \rho^{\lambda}_{u,v}:=\frac{\mu^+_v}{\tilde z^+_v}\bigl(z^+_{u,v}-\lambda\,z^-_{u,v}\bigr),
    \qquad \lambda\ge0 ,
\end{equation}
on the edges entering $H(x)\cup\{o\}$, and $\rho^\lambda_{u,v}:=0$ on the
edges entering the other outputs. Its positive part is $\pi^{z^+}$ for
$x_o>0$; for $x_o<0$, it is minus
the reference of $|x_o|$. We write SRP$_\lambda$ for SRP with this
reference, on its support $E^o_\rho(x)$ and with $y=x_o$ for $x_o>0$, and
with the convention above for $x_o\le0$. SRP$_0$ is LRP$_{z^+}$
(Proposition~\ref{prop:srp-lrp}(i)); for $\lambda>0$, the inhibitory
contributions enter as counter-evidence, with a total mass at most $e$
times that of the reference. The weight $\lambda$ plays the part of
$\beta$ in LRP$_{\alpha,\beta}$, but the balance is global instead of
node by node.

For $\rho\ge0$, SRP is a rule of Theorem~\ref{thm:RP0}: by
\eqref{eq:srp-form}, the net inflow of $v$ is
$\mu^{\mathrm S}_v=\varphi_v\sum_u\rho_{u,v}/\varphi_u$, so that
$\pi^{\mathrm S}_{u,v}=\mu^{\mathrm S}_v\,(\rho_{u,v}/\varphi_u)/\sum_r(\rho_{r,v}/\varphi_r)$
is the flow of Theorem~\ref{thm:RP0} on $E^o_\rho(x)$, with boundary mass
$y$, for the reference $\rho_{u,v}/\varphi_u$. Every statement on the
non-negative references of Theorem~\ref{thm:RP0} then applies to SRP, the
relevance by region of Section~\ref{sec:region} included.

\subsection{Application: explaining an image classifier}
\label{sec:app-images}

We apply the rules to VGG-16 \citep{SimonyanZisserman2015} with its
ImageNet weights \citep{Russakovsky2015} (Appendix~\ref{sec:architectures}),
a layered Filippov network (Definition~\ref{def:layered-DAG}) with one
input node per pixel and colour channel and one bias node of value $1$
per convolutional, dense and output node: a conservative rule
distributes $x_o$, the logit of the predicted class, over the pixels and
the bias nodes (Proposition~\ref{thm:RelevanceConservation}). At a
max-pooling node, every rule takes the reference $x_u$ on the winning
edge and $0$ elsewhere. The rules are LRP$_0$, here
Gradient$\times$Input (Corollary~\ref{thm:RP0ReLU}(i)); LRP$_{z^+}$;
LRPN$_{2,-1}$; LRPN$_{\alpha(\bar x),\beta(1-\bar x)}$, the Hebbian
weights $\alpha_v=\bar x_v:=x_v/\max_wx_w$, the maximum running over the
layer of $v$ ($\bar x_v:=0$ if it vanishes), and $\beta_v=1-\bar x_v$; SRP with $\rho=z^+$; and SRP$_1$
(\eqref{eq:srp-lambda}, $\lambda=1$), whose over-relaxed sweeps stop at
a conservation defect of $1\%$ of $x_o$, the recursion of LRP for
$\rho_{u,v}/\varphi_u$ then conserving $x_o$ exactly.

On $\NIMG$ Imagenette photographs \citep{Howard2019Imagenette}, we
measure the quantities of Table~\ref{tab:rules} and faithfulness by
region perturbation \citep{Samek2017}: the
$16\times16$ patches $M$ are blurred in decreasing (MoRF) or increasing
(LeRF) order of their region relevance $\mu(M)=\sum_{i\in M}\mu_i$
(Section~\ref{sec:region}), and ABPC is the area under LeRF minus that
under MoRF. Localization is the relevance mass accuracy
\citep{Arras2022}, the share of $\sum_i|\mu_i|$ inside the animal, on
$\NPET$ Oxford-IIIT Pet photographs \citep{Parkhi2012}; class
sensitivity, the Spearman correlation $\rho_2$ between the absolute maps
of the first two classes, on $\NCLS$ more photographs
\citep{Gu2018,Sixt2020}.

\begin{table}[t]
    \centering
    \caption{The rules on VGG-16. On $\NIMG$ Imagenette photographs:
    largest relative conservation error over the pixels and the bias
    nodes, mean share of $x_o$ absorbed by the bias nodes, mean share of
    negative relevance in the pixel maps, median time per image (NVIDIA
    T1000), and faithfulness (ABPC, higher is better). On $\NPET$
    Oxford-IIIT Pet photographs: relevance mass accuracy. On $\NCLS$
    Imagenette photographs: Spearman correlation $\rho_2$ of the absolute
    maps of the predicted and of the second-ranked class; $1$ means the
    same ranking of the pixels for both classes.}
    \label{tab:rules}
    \small
    \begin{tabular}{l cccc ccc}
        \toprule
        rule & cons.\ error & bias (\%) & neg.\ (\%) & time (s) & ABPC & mass & $\rho_2$ \\
        \midrule
        LRP$_0$ & $9\cdot10^{-7}$ & 92 & 50 & 0.11 & 0.053 & 0.736 & 0.76 \\
LRP$_{z^+}$ & $2\cdot10^{-7}$ & 16 & 0 & 0.16 & 0.431 & 0.816 & 1.00 \\
LRPN$_{2,-1}$ & $6\cdot10^{-7}$ & 32 & 20 & 0.26 & 0.431 & 0.855 & 1.00 \\
LRPN$_{\alpha(\bar x),\beta(1-\bar x)}$ & $4\cdot10^{-7}$ & 10 & 0 & 0.24 & 0.419 & 0.785 & 1.00 \\
SRP & $3\cdot10^{-7}$ & 9 & 0 & 22 & 0.329 & 0.642 & 1.00 \\
SRP$_1$ & $1\cdot10^{-6}$ & 40 & 2 & 93 & 0.427 & 0.820 & 1.00 \\
\midrule
random & -- & -- & -- & -- & -0.001 & 0.577 & -- \\
        \bottomrule
    \end{tabular}
\end{table}

\begin{figure}[t]
    \centering
    \includegraphics[width=0.92\linewidth]{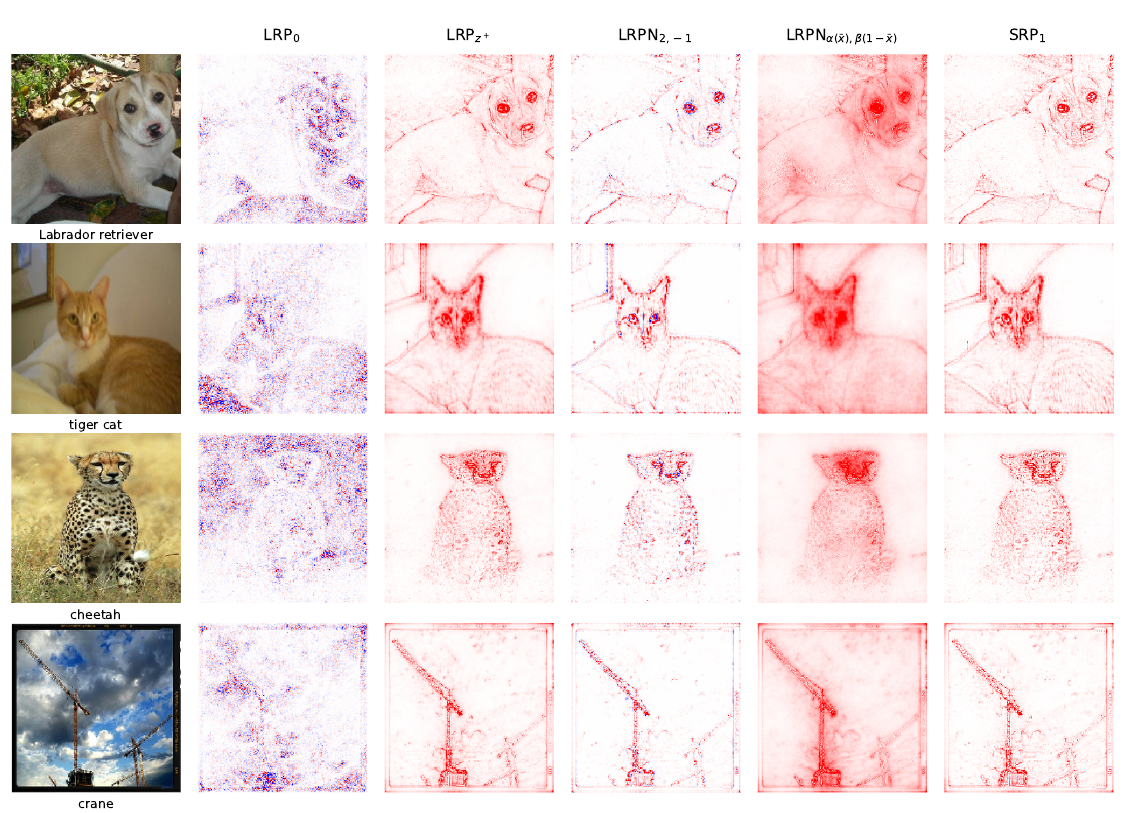}
    \caption{Pixel maps for the class predicted by VGG-16 (below each
    photograph): LRP$_0$, LRP$_{z^+}$, LRPN$_{2,-1}$,
    LRPN$_{\alpha,\beta}$ with $\alpha_v=\bar x_v$ (defined in the text),
    and SRP$_1$. Red: positive relevance, blue:
    negative; each map is scaled to the $99.5$th percentile of its
    $|\mu_i|$.}
    \label{fig:photos-rules}
\end{figure}

\begin{figure}[t]
    \centering
    \includegraphics[width=\linewidth]{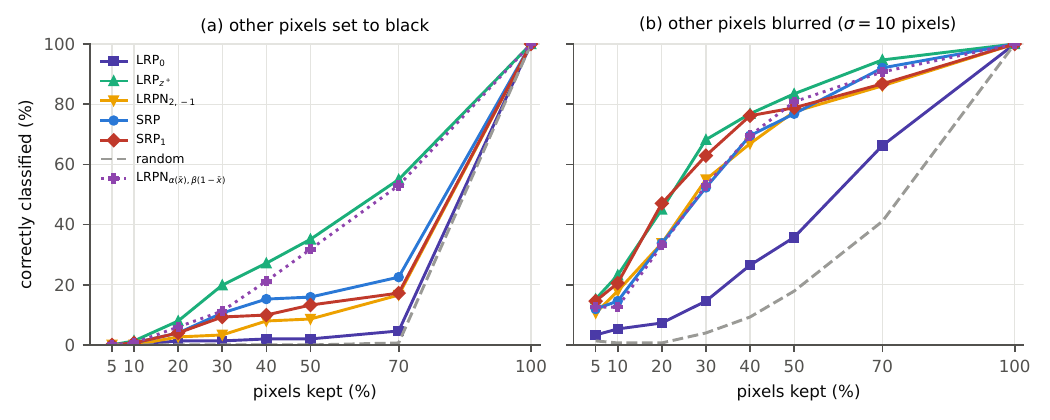}
    \caption{Share of the $\NKEEP$ Imagenette photographs (of the
    $\NIMG$) that VGG-16 classifies correctly, still classified in their
    true class when only the $k\%$ pixels of largest $|\mu_i|$ are kept,
    for the five rules and a random map (dotted:
    LRPN$_{\alpha(\bar x),\beta(1-\bar x)}$, defined in the text), the
    other pixels being set to
    black (a) or replaced by a blurred copy of the image (b); at $k=100$,
    every curve is at $100\%$ by construction.}
    \label{fig:keep}
\end{figure}

The rules conserve $x_o$ to single-precision round-off
(Table~\ref{tab:rules}). The LRP$_0$ map is half
negative, alternating from pixel to pixel
(Figure~\ref{fig:photos-rules}): its values of both signs compensate,
leaving $\BIASlrpzero\%$ of $x_o$ to the bias nodes, and largely cancel
within a patch; its ABPC is barely above that of a random map.
LRP$_{z^+}$ and LRPN$_{2,-1}$ have the highest ABPC, with no clear
paired difference in this sample ($\DABPClrpab\pm\DABPCSElrpab$),
LRPN$_{2,-1}$ localizing best;
LRPN$_{\alpha(\bar x),\beta(1-\bar x)}$, non-negative and more diffuse,
is nearly as faithful and leaves less of $x_o$ to the bias nodes. SRP
follows the edges and textures of the lower layers (rank correlation
$\CORRlrpzpsrp$ with LRP$_{z^+}$), is less faithful and localizes no
better than the area of the animal ($\AREA\%$); SRP$_1$ comes close to
the ABPC of LRP$_{z^+}$, with a small but systematic deficit (paired
difference $\DABPCsrpl\pm\DABPCSEsrpl$), at $\TIMESRPLMED$~s per image. ABPC and keeping pixels
(Figure~\ref{fig:keep}) reward the maps that cover the object, and every
rule but LRP$_0$ ranks the pixels almost as for the second class
($\rho_2$): the class enters only through the output layer, through its
weights $w^+_{u,o}$ and its bias for a non-negative reference and through
its weights of both signs for LRPN$_{2,-1}$ and SRP$_1$, whereas LRP$_0$
depends on it through its costate.

On VGG-16, a Gaussian noise of standard deviation $10^{-5}$ on $\KINKN$
photographs changes $x_o$ by $\KINKdxo$ (median) but switches
$\KINKgates$ of the $\KINKunits$ million ReLU gates and the winners of
$\KINKpools$ max-pooling windows, many events at once rather than the
isolated ones of Corollary~\ref{thm:RP0ReLU}. The LRP$_0$ map moves by
$\KINKlrpzero\%$ in $\ell^1$ norm (median), those of LRP$_{z^+}$,
LRPN$_{2,-1}$ and LRPN$_{\alpha(\bar x),\beta(1-\bar x)}$ by
$\KINKlrpzp\%$, $\KINKlrpab\%$ and $\KINKxvn\%$: consistent with the
jumps of LRP$_0$ at a kink (part (iii)(a)), the continuity of the other
rules there outside the case excluded in (iii)(b), and the jumps that a
change of winner may cause for every rule (part (iv)).

\begin{figure}[!tb]
    \centering
    \includegraphics[width=0.8\linewidth]{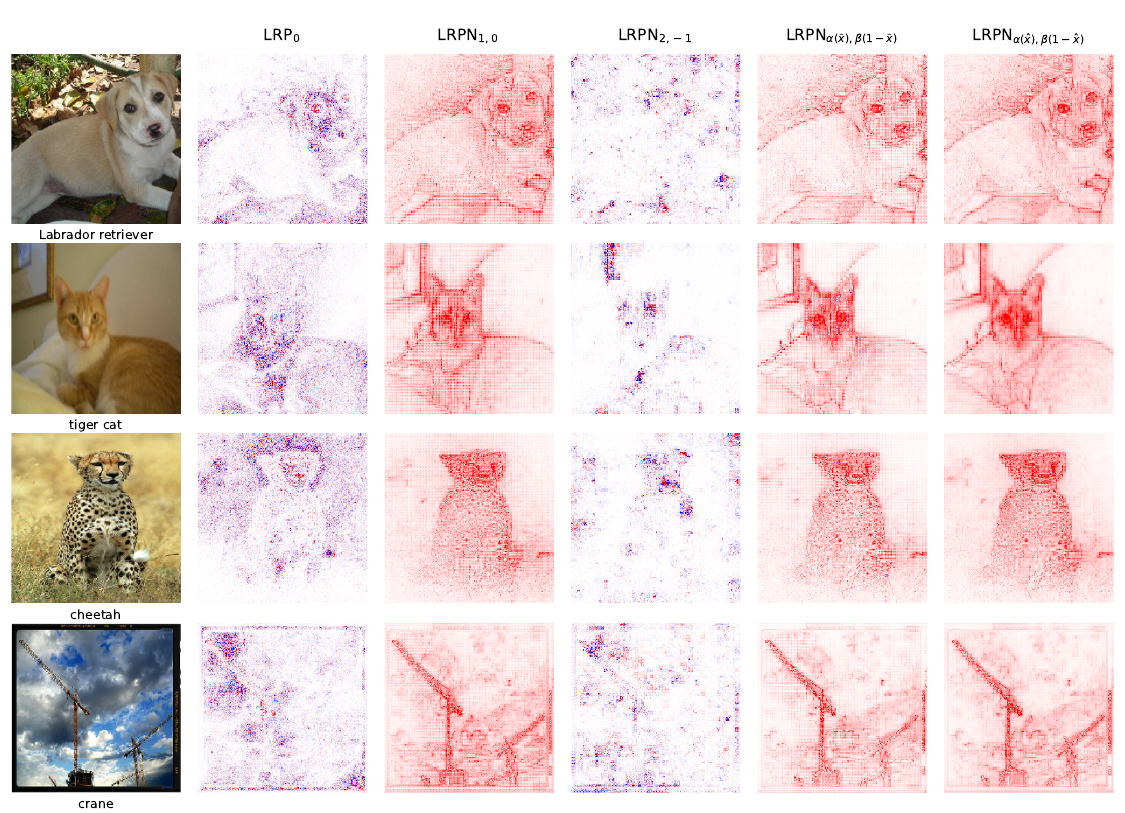}
    \caption{Pixel maps for the class predicted by ConvNeXt-Tiny (below
    each photograph), as in Figure~\ref{fig:photos-rules}: LRP$_0$,
    LRPN$_{1,0}$, LRPN$_{2,-1}$,
    LRPN$_{\alpha(\bar x),\beta(1-\bar x)}$ and
    LRPN$_{\alpha(\hat x),\beta(1-\hat x)}$ (defined in the text). Red:
    positive relevance, blue: negative; each map is scaled to the
    $99.5$th percentile of its $|\mu_i|$.}
    \label{fig:cnx-photos}
\end{figure}

\begin{figure}[!tb]
    \centering
    \includegraphics[width=\linewidth]{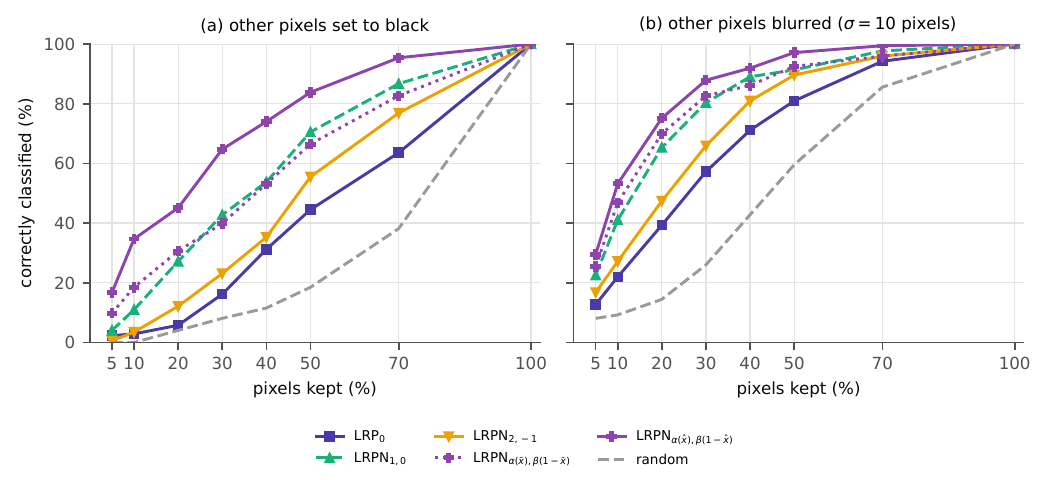}
    \caption{As Figure~\ref{fig:keep}, for ConvNeXt-Tiny and the rules of
    Figure~\ref{fig:cnx-photos}, with a random map: share of the
    $\CNXNkeep$ Imagenette photographs (of the $\CNXN$) that it classifies
    correctly, still classified in their true class when only the $k\%$
    pixels of largest $|\mu_i|$ are kept, the other pixels being set to
    black (a) or replaced by a blurred copy of the image (b).}
    \label{fig:cnx-keep}
\end{figure}

In ConvNeXt-Tiny \citep{Liu2022ConvNeXt} (Appendix~\ref{sec:architectures}),
GELU activations, layer normalizations and residual sums give the hidden
values both signs. We read it at $x$ as a network of linear and GELU
nodes, each layer normalization being linear with its standard deviation
held constant, and a GELU node splitting its relevance as a ReLU node
does. On the same $\CNXN$ photographs (Figures~\ref{fig:cnx-photos}
and~\ref{fig:cnx-keep}), LRP$_{z^+}$ loses $\CNXlostzp\%$ of $x_o$
(median) at the residual sums whose two contributions are negative,
where $\tilde z^+_v=0$ (Corollary~\ref{cor:lrp-references}(ii));
LRPN$_{1,0}$, the same rule with the fallback of (iv), conserves it, as
the other rules do. LRP$_0$ is no longer Gradient$\times$Input
(Spearman correlation $\CNXspearman$): extending the algebra of
Corollary~\ref{thm:RP0ReLU}(i) to GELU nodes, its factors $s_v$ become
$\mathrm{GELU}(\tilde z_v)/\tilde z_v$, and the standard deviations of
the layer normalizations, held constant, do not move with $x$ as they do
in the gradient; LRP$_0$ is the more faithful of the
two (ABPC $\CNXabpclrpzero$ against $\CNXabpcgi$). LRPN$_{2,-1}$
amplifies the absolute mass to $\CNXabslrpab\,x_o$ (median).
LRPN$_{\alpha(\bar x),\beta(1-\bar x)}$ has the highest ABPC of the rules tested (ABPC
$\CNXabpcxvn$, against $\CNXabpclrpnonezero$ for LRPN$_{1,0}$), and its
$10\%$ most relevant pixels, on a blurred background, keep the class of
$\CNXkeepxvn\%$ of the photographs, against $\CNXkeeplrpnonezero\%$ for
LRPN$_{1,0}$; with $\hat x_v:=(x_v-\min_wx_w)/(\max_wx_w-\min_wx_w)$
($\hat x_v:=0$ on a constant layer) in
place of $\bar x_v$, which can now be negative, this share rises to
$\CNXkeepxvmm\%$.

\section{Relevance by Region}
\label{sec:transformers}

The flows of Section~\ref{sec:LRP} live on the edges of the active
graph; this section aggregates them over sets of nodes. For a
non-negative reference, a region of the input -- a set of pixels --
receives the relevance of the paths that end in it, and the hidden units
that carry it are read off one forward pass (Section~\ref{sec:region});
we apply this to VGG-16 on $\RPN$ annotated photographs of cats and dogs,
to read which hidden units carry their head, their body and the
background (Section~\ref{sec:app-flows}).

\subsection{Region relevance and Schrödinger bridges}
\label{sec:region}

For a non-negative reference $\rho$
and a non-negative flow $\pi$, \eqref{eq:cond-entropy} is, by the chain rule, the
path-space entropy of Schrödinger's problem
\citep{Benamou2000,Peyre2019} between the Markov measures on backward
paths with transitions $\pi_{u,v}/m_v(\pi)$ and $K_{u,v}$
(Appendix~\ref{proof:path-space}); a signed reference loses this
reading.

In image classification, a relevance map says which pixels support the
class, not through which hidden units. Concept relevance propagation
\citep{Achtibat2023} answers one direction of this question: it keeps,
at a hidden layer, the relevance of a chosen channel only, and
propagates it to the image, which shows where the evidence used by that
channel lies. The converse question -- which channels carry the
evidence of a given region; are the head and the collar of a dog
recognized by the same channels? -- is answered by the Markov structure
of the path measure, which treats both directions alike, and
conditioning it on a region is an entropic transport problem with both
marginals prescribed. Figure~\ref{fig:joint-mass} shows the two readings,
and LRP itself, on the running example.

\begin{figure}[!tb]
    \centering
    \begin{tikzpicture}[x=1.5cm,y=0.78cm,every node/.style={font=\small}]
    \fill[red!10] (0.45,0.42) rectangle (3.55,-0.42);
    \fill[teal!14] (1.55,0.8) rectangle (2.45,-2.42);
    \node[blue] at (1,1.05) {$i_1$};
    \node[green!30!black!70] at (2,1.05) {$i_2$};
    \node at (3,1.05) {$b_1$};
    \node[gray] at (4.3,1.05) {$\mu_v$};
    \node[anchor=east] at (0.35,0) {paths through $h_1$};
    \node[anchor=east] at (0.35,-1) {paths through $h_2$};
    \node[anchor=east] at (0.35,-2) {direct edge $i_1\to o_1$};
    \node[anchor=east,gray] at (0.35,-3.15) {$\mu_i$ (LRP)};
    \foreach \x/\v in {1/1,2/2,3/0} \node[font=\small\bfseries] at (\x,0) {$\v$};
    \foreach \x/\v in {1/0,2/1,3/1} \node at (\x,-1) {$\v$};
    \foreach \x/\v in {1/1,2/0,3/0} \node at (\x,-2) {$\v$};
    \draw[gray] (0.45,-2.6) -- (3.55,-2.6);
    \draw[gray] (3.75,0.42) -- (3.75,-2.42);
    \foreach \x/\v in {1/2,2/3,3/1} \node[gray] at (\x,-3.15) {$\v$};
    \foreach \y/\v in {0/3,-1/2,-2/1} \node[gray] at (4.3,\y) {$\v$};
    \node[gray] at (4.3,-3.15) {$6$};
    \node[anchor=west,red!70!black,font=\footnotesize,align=left] at (4.75,0.15)
        {row: CRP, where the evidence\\of $h_1$ lies};
    \node[anchor=west,teal!70!black,font=\footnotesize,align=left] at (4.75,-1.75)
        {column: Corollary~\ref{cor:region-relevance}, which units\\carry the evidence of $M=\{i_2\}$;\\the bridge rescales it by $x_{o_1}/\mu(M)=2$:\\$4$ through $h_1$, $2$ through $h_2$};
    \node[anchor=west,gray,font=\footnotesize,align=left] at (4.75,-3.15)
        {margin: the LRP relevances};
    \end{tikzpicture}
    \caption{The joint mass $\mu_vR^{(v)}_i$ of
    Corollary~\ref{cor:region-relevance}(i) on the running example at
    $x=(1,1)$, for $o_1$ and $\rho=z^+$, the paths being grouped by the
    hidden node they cross. LRP reads its column margins, concept
    relevance a row, and the region relevances a column.}
    \label{fig:joint-mass}
\end{figure}

\begin{corollary}
    \label{cor:region-relevance}
    Let $x\in\mathbb R^d$, $\kappa\in\mathcal M(x)$, $o\in\mathcal O$ an
    active output with $x^{(\kappa)}_o>0$, and
    $\rho\in\mathbb R^{E^{(\kappa)}(x)}_{\ge0}$ satisfying the assumptions
    of Theorem~\ref{thm:RP0}, of kernel $K$ extended by $0$ off
    $E^{(\kappa)}(x)$; write $\mu$ and $\pi^\ast$ for $\mu^{o,\kappa}$ and
    $\pi^{o,\kappa,\ast}$. For nodes $u,v$, let
    \[
        R^{(v)}_u:=\sum_{p\in\mathcal P(u,v)}\ \prod_{(a,b)\in p}K_{a,b},
    \]
    the relevance of $u$ when the recursion of Theorem~\ref{thm:RP0} is
    started from $v$ with unit mass, and, for
    $M\subset\mathcal I(x)$, $R^{(v)}(M):=\sum_{i\in M}R^{(v)}_i$ and
    $\mu(M):=\sum_{i\in M}\mu_i$.

    \smallskip\noindent
    (i) For every $v\in V^{(\kappa)}(x)\setminus(\mathcal O\setminus\{o\})$,
    $(R^{(v)}_i)_{i\in\mathcal I(x)}$ is a probability vector, and the
    paths of the backward Markov chain of kernel $K$ started at $o$ with
    mass $x^{(\kappa)}_o$ (Appendix~\ref{proof:path-space}) that pass
    through $v$ and end at $i$ carry the mass $\mu_v R^{(v)}_i$.

    \smallskip\noindent
    (ii) The \emph{region relevances} $\mu^M_v:=\mu_v\,R^{(v)}(M)$
    satisfy $\sum_{M\in\mathfrak M}\mu^M_v=\mu_v$ for every partition
    $\mathfrak M$ of $\mathcal I(x)$, and
    $\sum_{v\in C}\mu^M_v=\mu(M)$ for every cut $C$, i.e.\ every set of
    nodes met exactly once by each directed path of $G^{(\kappa)}(x)$
    from an input to $o$.

    \smallskip\noindent
    (iii) If $\mu(M)>0$, the problem
    \[
        \min\Bigl\{\mathcal H(\pi\,\|\,K):\ \pi\in\mathcal A^{(\kappa)}_o(x),\
        \pi\ge0,\ \sum_{v:\,i\to v}\pi_{i,v}=\frac{x^{(\kappa)}_o}{\mu(M)}\,
        \mu_i\,\mathbf 1_M(i)\ \ (i\in\mathcal I(x))\Bigr\}
    \]
    has the unique minimizer
    $\pi^M_{u,v}:=\bigl(x^{(\kappa)}_o/\mu(M)\bigr)\,\pi^\ast_{u,v}\,R^{(u)}(M)$,
    of node relevances $\bigl(x^{(\kappa)}_o/\mu(M)\bigr)\,\mu^M_u$, and
    its minimal value is $x^{(\kappa)}_o\log\bigl(x^{(\kappa)}_o/\mu(M)\bigr)$.
\end{corollary}
Proof of the corollary is provided in
Appendix~\ref{proof:cor:region-relevance}.

The region relevance $\mu^M_v$ is the relevance of $v$ for $o$ times the
share of the evidence of $v$ itself that lies in $M$. Over a partition
of the image into regions -- superpixels, or the parts of the object --
the $\mu^M_v$ form a table of hidden units against regions whose margins
are the relevances of Theorem~\ref{thm:RP0}: for the dog, the
row of a channel tells on which parts of the image it relies, and the
column of the head which channels carry its evidence. Both directions
read the joint mass $\mu_vR^{(v)}_i$ of (i): concept relevance by hidden
unit, the region relevances by region. Part (iii) is the
two-marginal problem whose one-marginal form is Theorem~\ref{thm:RP0}:
$\pi^M$ is the Schrödinger bridge, the most likely routing of relevance
given that it ends in $M$, and its potential $R^{(\cdot)}(M)$ follows
from one forward pass. With a signed reference, (ii) still holds by
linearity, without its probabilistic meaning.

The bridge rescales each edge at both ends,
$\pi^M_{u,v}=\bigl(x_o/\mu(M)\bigr)K_{u,v}\,\mu_v\,R^{(u)}(M)$: by $\mu_v$ at its head, as
LRP does, and by the potential $R^{(u)}(M)$ at its tail, the factor
$\bigl(x_o/\mu(M)\bigr)\mathbf 1_M$ of the endpoint propagated back along the paths.
Sinkhorn Relevance Propagation has the same two-ended form,
$\pi^{\mathrm S}_{u,v}=\rho_{u,v}\varphi_v/\varphi_u$
(Theorem~\ref{thm:srp}), but the two problems compare different objects:
the bridge compares path measures, with a constraint at the end of the
paths only; SRP compares flows on the edges, with a constraint at every
node, and its flow is not in general the flow of the chain of
transitions $K$ (Proposition~\ref{prop:srp-lrp}(iii)).

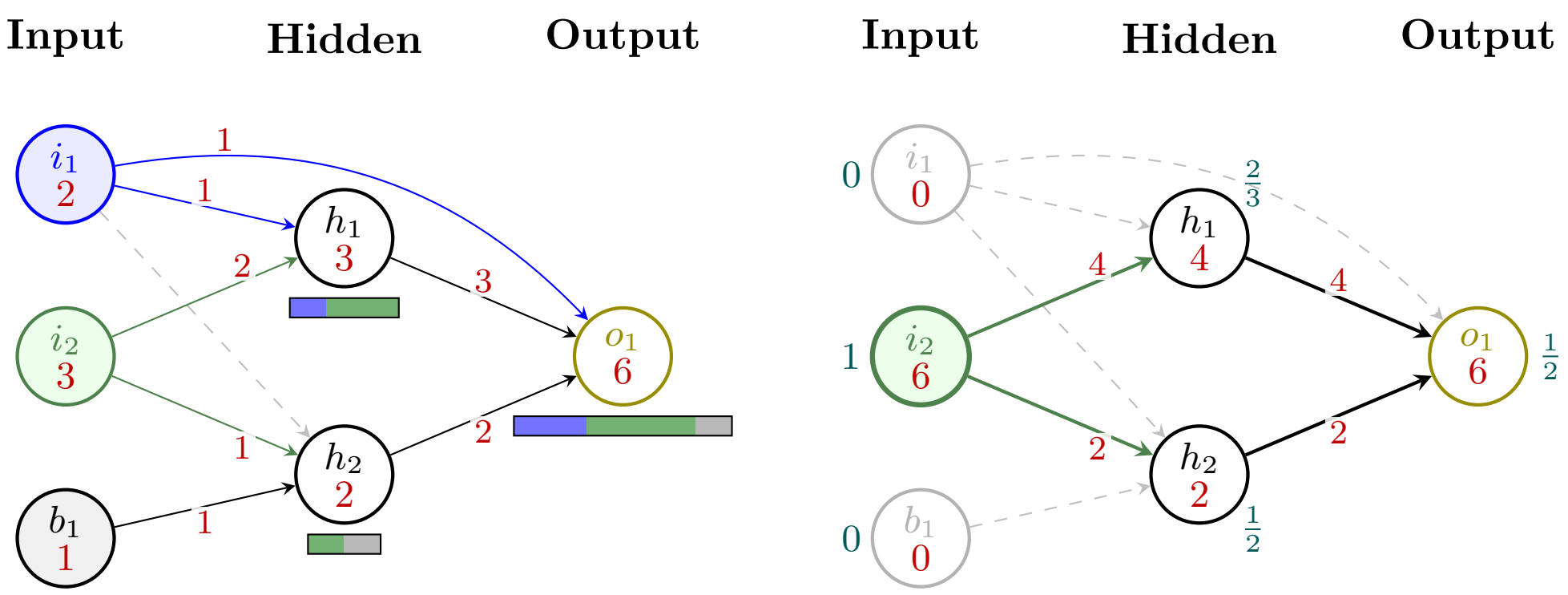
\begin{figure}[H]
    \centering
    \newcommand{\regionbar}[4]{%
        \path (#1.south) ++({-(#2+#3+#4)*0.15cm},-0.08cm) coordinate (rb0)
            ++({#2*0.3cm},0) coordinate (rb1) ++({#3*0.3cm},0) coordinate (rb2);
        \fill[blue!55] (rb0) rectangle ++({#2*0.3cm},-0.16cm);
        \fill[green!45!black!55] (rb1) rectangle ++({#3*0.3cm},-0.16cm);
        \fill[gray!55] (rb2) rectangle ++({#4*0.3cm},-0.16cm);
        \draw[thin] (rb0) rectangle ++({(#2+#3+#4)*0.3cm},-0.16cm);}
    \begin{tikzpicture}[
        x=2.3cm,
        y=1.5cm,
        >=stealth,
        neuron/.style={circle,draw,font=\footnotesize,inner sep=1pt,minimum size=0.8cm,thick},
        fl/.style={inner sep=1pt,font=\scriptsize,red!75!black,fill=white,fill opacity=0.85,text opacity=1},
        faint/.style={->,gray!50,dashed},
        every node/.style={font=\small}
    ]
    \node[neuron,blue,fill=blue!8] (x1) at (0,2)
        {\shortstack{$i_1$\\[-1pt]{\color{red!75!black}$2$}}};
    \node[neuron,green!30!black!70,fill=green!8] (x2) at (0,1)
        {\shortstack{$i_2$\\[-1pt]{\color{red!75!black}$3$}}};
    \node[neuron,fill=gray!12] (b1) at (0,0)
        {\shortstack{$b_1$\\[-1pt]{\color{red!75!black}$1$}}};
    \node[neuron] (h1) at (1,1.65)
        {\shortstack{$h_1$\\[-1pt]{\color{red!75!black}$3$}}};
    \node[neuron] (h2) at (1,0.35)
        {\shortstack{$h_2$\\[-1pt]{\color{red!75!black}$2$}}};
    \node[neuron,olive] (y1) at (2,1)
        {\shortstack{$o_1$\\[-1pt]{\color{red!75!black}$6$}}};
    \node at (0,2.75) {\textbf{Input}};
    \node at (1,2.75) {\textbf{Hidden}};
    \node at (2,2.75) {\textbf{Output}};

    \draw[faint] (x1) -- (h2);
    \draw[->,blue] (x1) -- node[fl,above,pos=0.5] {$1$} (h1);
    \draw[->,green!30!black!70] (x2) -- node[fl,above,pos=0.7] {$2$} (h1);
    \draw[->,green!30!black!70] (x2) -- node[fl,below,pos=0.7] {$1$} (h2);
    \draw[->] (b1) -- node[fl,below,pos=0.5] {$1$} (h2);
    \draw[->] (h1) -- node[fl,above,pos=0.5] {$3$} (y1);
    \draw[->] (h2) -- node[fl,below,pos=0.5] {$2$} (y1);
    \draw[->,blue] (x1) to[bend left=28] node[fl,above,pos=0.2] {$1$} (y1);

    \regionbar{h1}{1}{2}{0}
    \regionbar{h2}{0}{1}{1}
    \regionbar{y1}{2}{3}{1}
    \end{tikzpicture}
    \hspace{0.6cm}
    \begin{tikzpicture}[
        x=2.3cm,
        y=1.5cm,
        >=stealth,
        neuron/.style={circle,draw,font=\footnotesize,inner sep=1pt,minimum size=0.8cm,thick},
        fl/.style={inner sep=1pt,font=\scriptsize,red!75!black,fill=white,fill opacity=0.85,text opacity=1},
        pot/.style={font=\footnotesize,teal!70!black,inner sep=1pt},
        faint/.style={->,gray!50,dashed},
        every node/.style={font=\small}
    ]
    \node[neuron,gray!60] (x1) at (0,2)
        {\shortstack{$i_1$\\[-1pt]{\color{red!75!black}$0$}}};
    \node[neuron,green!30!black!70,fill=green!8,very thick] (x2) at (0,1)
        {\shortstack{$i_2$\\[-1pt]{\color{red!75!black}$6$}}};
    \node[neuron,gray!60] (b1) at (0,0)
        {\shortstack{$b_1$\\[-1pt]{\color{red!75!black}$0$}}};
    \node[neuron] (h1) at (1,1.65)
        {\shortstack{$h_1$\\[-1pt]{\color{red!75!black}$4$}}};
    \node[neuron] (h2) at (1,0.35)
        {\shortstack{$h_2$\\[-1pt]{\color{red!75!black}$2$}}};
    \node[neuron,olive] (y1) at (2,1)
        {\shortstack{$o_1$\\[-1pt]{\color{red!75!black}$6$}}};
    \node at (0,2.75) {\textbf{Input}};
    \node at (1,2.75) {\textbf{Hidden}};
    \node at (2,2.75) {\textbf{Output}};

    \node[pot,left=1pt of x1] {$0$};
    \node[pot,left=1pt of x2] {$1$};
    \node[pot,left=1pt of b1] {$0$};
    \node[pot,above right=-2pt and 0pt of h1] {$\tfrac23$};
    \node[pot,below right=-2pt and 0pt of h2] {$\tfrac12$};
    \node[pot,right=1pt of y1] {$\tfrac12$};

    \draw[faint] (x1) -- (h2);
    \draw[faint] (x1) -- (h1);
    \draw[faint] (b1) -- (h2);
    \draw[faint] (x1) to[bend left=28] (y1);
    \draw[->,green!30!black!70,thick] (x2) -- node[fl,above,pos=0.7] {$4$} (h1);
    \draw[->,green!30!black!70,thick] (x2) -- node[fl,below,pos=0.7] {$2$} (h2);
    \draw[->,thick] (h1) -- node[fl,above,pos=0.5] {$4$} (y1);
    \draw[->,thick] (h2) -- node[fl,below,pos=0.5] {$2$} (y1);
    \end{tikzpicture}
    \caption{Corollary~\ref{cor:region-relevance} on the running example
    at $x=(1,1)$, for $o_1$ and $\rho=z^+$ ($b_2$, $o_2$ omitted).
    Left: LRP$_{z^+}$ flows and relevances, and the region relevances
    $\mu^{M}_v$ of each node, for $M_1=\{i_1\}$ (blue), $M_2=\{i_2\}$
    (green) and $M_3=\{b_1,b_2\}$ (gray). Right: the bridge $\pi^M$
    for $M=M_2$, with the forward potentials $R^{(v)}(M)$ (teal).}
    \label{fig:region-relevance}
\end{figure}

In Figure~\ref{fig:region-relevance} ($x=(1,1)$, $\rho=z^+$, regions
$M_1=\{i_1\}$, $M_2=\{i_2\}$, $M_3=\{b_1,b_2\}$), $\mu_{h_1}=3$ splits as
$1+2+0$ and $\mu_{h_2}=2$ as $0+1+1$: $h_1$ carries the evidence of
$i_1$, $h_2$ that of the bias, and both share that of $i_2$.
Conditioning on $M=M_2$ sends $4$ through $h_1$ and $2$ through $h_2$,
the whole mass $x_{o_1}=6$ ending at $i_2$.

Figure~\ref{fig:schrodinger} makes the analogy explicit. In
Schrödinger's problem, particles follow a reference Brownian motion and
are observed with an unexpected final distribution; the most likely
evolution compatible with both marginals minimizes the relative entropy
to the reference on path space, and is the reference reweighted by a
function of the endpoint. Here the particles are the quanta of
relevance, the reference is the chain of kernel $K$, the initial
marginal is $x_o\delta_o$ and the final one $\nu$, and the reweighting is
the Doob transform by $h=R^{(\cdot)}(M)$.

\begin{figure}[H]
    \centering
    \begin{tikzpicture}[x=1.7cm,y=1.05cm,>=stealth,
        nd/.style={circle,draw,thick,minimum size=0.62cm,inner sep=0pt,font=\footnotesize},
        tp/.style={inner sep=1pt,font=\scriptsize,fill=white,fill opacity=0.85,text opacity=1},
        every node/.style={font=\small}]
    \node[font=\small\bfseries,align=center] at (1,3.05) {(a) reference $Q_K$:\\the LRP chain};
    \node[nd,olive] (o) at (0,1) {$o_1$};
    \node[nd] (h1) at (1,1.65) {$h_1$};
    \node[nd] (h2) at (1,0.35) {$h_2$};
    \node[nd,blue] (i1) at (2,2.2) {$i_1$};
    \node[nd,green!30!black!70] (i2) at (2,1) {$i_2$};
    \node[nd] (b1) at (2,-0.2) {$b_1$};
    \draw[line width=3pt,olive] (-0.42,0.7) -- (-0.42,1.3);
    \node[font=\scriptsize,olive,align=center] at (-0.45,0.22) {$P_0=x_o\delta_o$};
    \draw[->,red!75!black,line width=2.4pt] (o) -- node[tp,above,pos=0.5] {$\tfrac12$} (h1);
    \draw[->,red!75!black,line width=1.8pt] (o) -- node[tp,below,pos=0.5] {$\tfrac13$} (h2);
    \draw[->,red!75!black,line width=1.0pt] (h1) -- node[tp,above,pos=0.5] {$\tfrac13$} (i1);
    \draw[->,red!75!black,line width=1.8pt] (h1) -- node[tp,above,pos=0.62] {$\tfrac23$} (i2);
    \draw[->,red!75!black,line width=1.0pt] (h2) -- node[tp,below,pos=0.62] {$\tfrac12$} (i2);
    \draw[->,red!75!black,line width=1.0pt] (h2) -- node[tp,below,pos=0.5] {$\tfrac12$} (b1);
    \draw[->,red!75!black,line width=1.0pt] (o) to[bend left=38] node[tp,above,pos=0.35] {$\tfrac16$} (i1);
    \fill[blue!70] (2.32,2.06) rectangle (2.520,2.3400000000000003);
    \node[anchor=west,font=\scriptsize] at (2.560,2.2) {$\tfrac13$};
    \fill[green!30!black!70] (2.32,0.86) rectangle (2.620,1.1400000000000001);
    \node[anchor=west,font=\scriptsize] at (2.660,1) {$\tfrac12$};
    \fill[black!70] (2.32,-0.34) rectangle (2.420,-0.06);
    \node[anchor=west,font=\scriptsize] at (2.460,-0.2) {$\tfrac16$};
    \node[font=\scriptsize,align=center] at (2.62,-0.75) {$P_T=\mu$};
    \draw[->,gray] (0,-1.15) -- (2,-1.15) node[midway,below,font=\scriptsize] {backward time: output $\to$ inputs};

    \end{tikzpicture}
    \hspace{0.3cm}
    \begin{tikzpicture}[x=1.7cm,y=1.05cm,>=stealth,
        nd/.style={circle,draw,thick,minimum size=0.62cm,inner sep=0pt,font=\footnotesize},
        tp/.style={inner sep=1pt,font=\scriptsize,fill=white,fill opacity=0.85,text opacity=1},
        every node/.style={font=\small}]
    \node[font=\small\bfseries,align=center] at (1,3.05) {(b) bridge $P^M$, $M=\{i_2\}$:\\transitions $K_{u,v}\,h(u)/h(v)$};
    \node[nd,olive] (o) at (0,1) {$o_1$};
    \node[nd] (h1) at (1,1.65) {$h_1$};
    \node[nd] (h2) at (1,0.35) {$h_2$};
    \node[nd,blue] (i1) at (2,2.2) {$i_1$};
    \node[nd,green!30!black!70] (i2) at (2,1) {$i_2$};
    \node[nd] (b1) at (2,-0.2) {$b_1$};
    \draw[line width=3pt,olive] (-0.42,0.7) -- (-0.42,1.3);
    \node[font=\scriptsize,olive,align=center] at (-0.45,0.22) {$P_0=x_o\delta_o$};
    \draw[->,red!75!black,line width=2.8pt] (o) -- node[tp,above,pos=0.5] {$\tfrac23$} (h1);
    \draw[->,red!75!black,line width=1.8pt] (o) -- node[tp,below,pos=0.5] {$\tfrac13$} (h2);
    \draw[->,gray!45,dashed] (h1) -- node[tp,above,pos=0.5] {$0$} (i1);
    \draw[->,red!75!black,line width=2.8pt] (h1) -- node[tp,above,pos=0.62] {$1$} (i2);
    \draw[->,red!75!black,line width=1.8pt] (h2) -- node[tp,below,pos=0.62] {$1$} (i2);
    \draw[->,gray!45,dashed] (h2) -- node[tp,below,pos=0.5] {$0$} (b1);
    \draw[->,gray!45,dashed] (o) to[bend left=38] node[tp,above,pos=0.35] {$0$} (i1);
    \node[anchor=west,font=\scriptsize] at (2.360,2.2) {$0$};
    \fill[green!30!black!70] (2.32,0.86) rectangle (2.920,1.1400000000000001);
    \node[anchor=west,font=\scriptsize] at (2.960,1) {$1$};
    \node[anchor=west,font=\scriptsize] at (2.360,-0.2) {$0$};
    \node[font=\scriptsize,align=center] at (2.62,-0.75) {$P_T=\nu$ prescribed};
    \draw[->,gray] (0,-1.15) -- (2,-1.15) node[midway,below,font=\scriptsize] {backward time: output $\to$ inputs};
    \node[font=\scriptsize,teal!70!black] at (1,2.12) {$h=\tfrac23$};
    \node[font=\scriptsize,teal!70!black] at (1,-0.12) {$h=\tfrac12$};
    \node[font=\scriptsize,teal!70!black] at (-0.15,1.5) {$h=\tfrac12$};
    \end{tikzpicture}

    \smallskip
    {\small $P^M=\operatorname*{argmin}\bigl\{\mathrm{KL}(P\,\|\,Q_K):\ P_0=x_o\delta_o,\ P_T=\nu\bigr\}$,\quad
    $\dfrac{dP^M}{dQ_K}=\dfrac{x_o}{\mu(M)}\,\mathbf 1_M(\text{endpoint})$,\quad
    $\mathrm{KL}(P^M\|Q_K)=x_o\log\dfrac{x_o}{\mu(M)}=6\log 2$}
    \caption{Corollary~\ref{cor:region-relevance}(iii) as a Schrödinger
    problem, on the running example at $x=(1,1)$, for $o_1$ and
    $\rho=z^+$. Both path measures start from the Dirac mass $x_o\delta_o$;
    edges carry transition probabilities, with widths proportional to the
    flows, and bars give the endpoint masses divided by $x_o$. (a) The reference ends
    with the LRP relevances. (b) The bridge is forced to end in $M$, with
    the forward potentials $h=R^{(\cdot)}(M)$ (teal); it sends $4$ through
    $h_1$ and $2$ through $h_2$.}
    \label{fig:schrodinger}
\end{figure}

\subsection{Application: regions of an image}
\label{sec:app-flows}

We apply the relevance by region of Section~\ref{sec:region} to the
image classifier of Section~\ref{sec:app-images}, to read which hidden
units carry each part of an image.

The regions come from the Oxford-IIIT Pet dataset \citep{Parkhi2012},
whose photographs carry two annotations: a trimap, which separates the
animal, its boundary and the background, and a bounding box of the head
of the animal. They fix the regions at the resolution of the pixels: the
\emph{head} is the set of pixels of the animal (foreground and boundary
of the trimap) inside the head box, enlarged by $10\%$ of its size on
each side to take in the ears, which the annotated boxes leave out; the
\emph{body} is the set of the other pixels of
the animal, the \emph{background} the remaining pixels, and the bias
nodes form one more region. Photographs and annotations are resized and
cropped to $224\times224$ pixels in the same way. The trainval
photographs are shuffled once and examined in that order; a photograph
is kept if it has a head box, if VGG-16 predicts a cat or dog class,
with $x_o>0$, if the crop keeps at least $95\%$ of the head box, and if
the head and the body cover at least $2\%$ and $15\%$ of the image,
which leaves out the close-ups, where the animal outside the head box is
little more than the ears; $\RPN$ photographs are kept ($\RPncat$ cats
and $\RPndog$ dogs) out of $\RPseen$ examined, $\RPskipnobox$ having no
head box, $\RPskipnotpet$ not being predicted as a cat or a dog,
$\RPskipboxcropped$ having the head cut by the crop and
$\RPskipsmallregion$ too small a head or body.
Corollary~\ref{cor:region-relevance} applies to LRP$_{z^+}$ and to the
LRPN$_{\alpha(\bar x),\beta(1-\bar x)}$ of Section~\ref{sec:app-images}:
the hidden activations of VGG-16 being non-negative and $x_o>0$, its
weights $\bar x_v$ and $1-\bar x_v$ lie in $[0,1]$ and its reference is
non-negative, unlike that of LRPN$_{2,-1}$. For every region $M$ and
every node $v$ of a cut, the region relevance $\mu^M_v=\mu_vR^{(v)}(M)$
is the backward relevance times one forward potential, so that the
three regions of pixels cost three forward passes, batched. Every path
from a pixel to $o$ meets a whole convolutional layer exactly once,
which is all that the cut identity of
Corollary~\ref{cor:region-relevance}(ii) needs for a region of pixels
(the paths from the bias nodes of the upper layers do not end in it);
we use conv4\_3 and conv5\_3, summing $\mu^M_v$ over the positions of
each of their $512$ channels $c$ into $\mu^M_c$.

On each photograph and for each rule, we record the share of $x_o$
carried by each region at conv5\_3; the share of the pixel relevance
that falls in the head, and its \emph{density} relative to the body,
i.e.\ the share of the pixel relevance per unit area of the head divided
by that of the body; the number of channels of conv5\_3 and conv4\_3
that carry half of the evidence of the head; the \emph{purity} of the
ten channels that carry most of it, the mean share of their evidence
lying in the head; and the number of channels common to the ten leading
channels of the head and of the body. Table~\ref{tab:regions} gives
their medians and quartiles over the $\RPN$ photographs.

\begin{figure}[t]
    \centering
    \includegraphics[width=\linewidth]{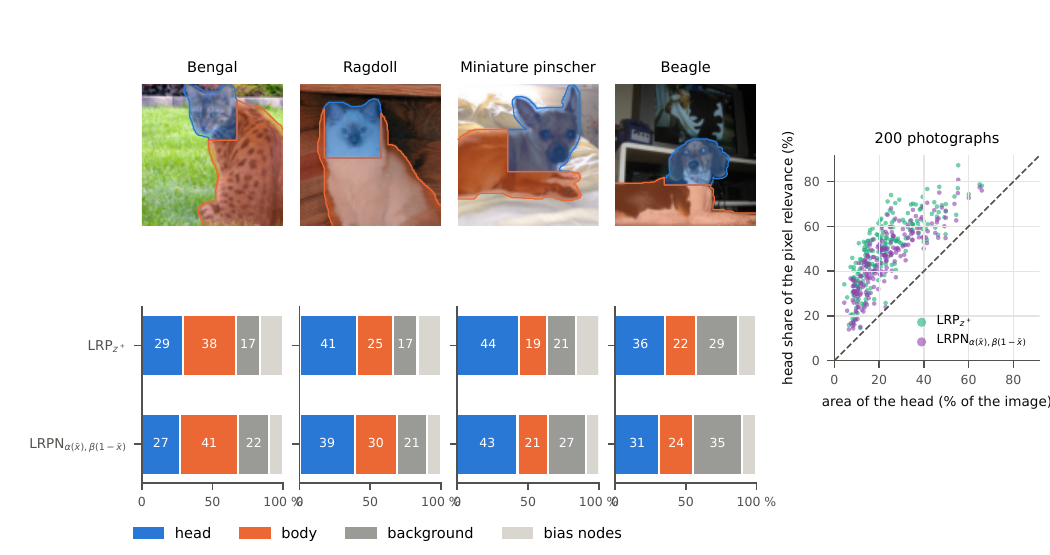}
    \caption{Relevance by region on VGG-16
    (Corollary~\ref{cor:region-relevance}), Oxford-IIIT Pet. Left, top:
    the regions given by the annotations of the dataset on the first two
    cats and the first two dogs of the sample: head (blue, the pixels of
    the animal inside the enlarged head box), body (orange, the other
    pixels of the animal) and background. Left, bottom: share of $x_o$
    carried by each region and by the bias nodes, in percent, for
    LRP$_{z^+}$ and LRPN$_{\alpha(\bar x),\beta(1-\bar x)}$, computed at
    conv5\_3. Right: share of the pixel relevance falling in the head
    against the area of the head, on the $\RPN$ photographs; dashed:
    equal shares.}
    \label{fig:region-shares}
\end{figure}

\begin{figure}[t]
    \centering
    \includegraphics[width=\linewidth]{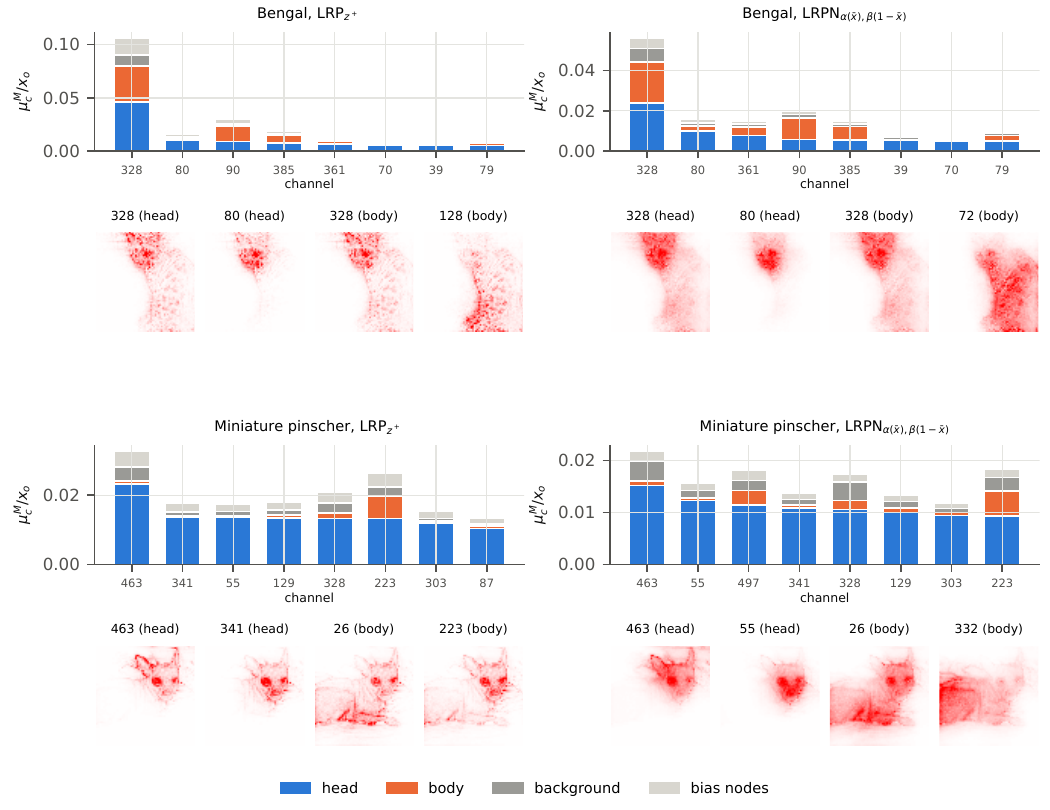}
    \caption{Which conv5\_3 channels carry the evidence of the head, for
    LRP$_{z^+}$ (left) and LRPN$_{\alpha(\bar x),\beta(1-\bar x)}$
    (right), on the first cat and the first dog of
    Figure~\ref{fig:region-shares}. Bars: the eight channels $c$ of
    largest region relevance $\mu^M_c$ for $M=$ head, each bar being
    the row $(\mu^{M'}_c)_{M'}$ of the table -- the relevance of the
    channel split by the region its evidence comes from, the bias nodes
    upstream included --, normalized by $x_o$. Below: the evidence of
    the two leading channels of the head and of the body, i.e.\ the pixel
    relevance obtained by propagating the relevance of that channel alone
    (concept relevance); the channels of the body are ranked by their own
    column, and need not appear among the bars.}
    \label{fig:region-channels}
\end{figure}

\begin{table}[t]
    \centering
    \caption{Relevance by region on the $\RPN$ Oxford-IIIT Pet
    photographs (Figure~\ref{fig:region-shares}): median and, in
    parentheses, quartiles over the photographs. Area of the regions;
    share of $x_o$ carried by each region and by the bias nodes, at
    conv5\_3; for the head, its share of the pixel relevance, its
    relevance per unit area divided by that of the body, the number of
    channels of conv5\_3 and conv4\_3 that carry half of its evidence,
    the mean share of their evidence lying in the head over its ten
    leading channels (purity), and the number of channels common to the
    ten leading channels of the head and of the body.}
    \label{tab:regions}
    \small
    \begin{tabular}{lcc}
        \toprule
        & LRP$_{z^+}$ & LRPN$_{\alpha(\bar x),\beta(1-\bar x)}$ \\
        \midrule
        \multicolumn{3}{l}{\emph{area of the region (\% of the image)}} \\
\quad head & \multicolumn{2}{c}{21 (13--31)} \\
\quad body & \multicolumn{2}{c}{26 (21--34)} \\
\quad background & \multicolumn{2}{c}{49 (39--58)} \\
\addlinespace\multicolumn{3}{l}{\emph{share of $x_o$ carried by the region (\%)}} \\
\quad head & 42 (32--50) & 42 (31--50) \\
\quad body & 22 (17--28) & 25 (20--32) \\
\quad background & 18 (14--23) & 23 (18--28) \\
\quad bias nodes & 17 (15--19) & 9 (9--10) \\
\addlinespace\multicolumn{3}{l}{\emph{head}} \\
\quad share of the pixel relevance (\%) & 51 (40--60) & 46 (35--55) \\
\quad density of relevance, head / body & 2.5 (1.9--3.2) & 2.1 (1.7--2.6) \\
\quad conv5\_3 channels for half of its evidence & 24 (21--28) & 31 (28--35) \\
\quad conv4\_3 channels for half of its evidence & 65 (59--73) & 87 (77--96) \\
\quad purity of its ten leading channels (\%) & 76 (66--82) & 73 (65--80) \\
\quad channels common to the ten leading of head and body & 3 (2--4) & 2 (1--3) \\
        \bottomrule
    \end{tabular}
    \normalsize
\end{table}

The cut identity holds to within $\RPerr\,x_o$ for both rules, at both
layers, on every photograph. Under LRP$_{z^+}$, the head, which covers
$\RPareaheadlrpzp\%$ of the image (median), carries
$\RPshareheadlrpzp\%$ of $x_o$ and $\RPpixheadlrpzp\%$ of the relevance
of the pixels; per unit area, it receives $\RPdenslrpzp$ times as much
relevance as the body (median), and more than the body on
$\RPdensonelrpzp$ of the $\RPN$ photographs
(Figure~\ref{fig:region-shares}, right). Half of its evidence goes
through $\RPhalffivelrpzp$ of the $512$ channels of conv5\_3
($\RPhalffourlrpzp$ at conv4\_3), and the ten leading ones draw
$\RPpuritylrpzp\%$ of their evidence from the head, more than the share
of the head in the pixel relevance -- the purity of a channel whose
evidence would be spread like the whole map -- on $\RPpurityabovelrpzp$
of the $\RPN$ photographs. The head and the body are carried by
largely different channels, $\RPcommonlrpzp$ of their ten leading
channels being common (median). The concept maps of the leading
channels of the head (Figure~\ref{fig:region-channels}) lie on the
face, those of the body on the rest of the animal, and a channel can
carry evidence from both: a row of the table says where the evidence of
a channel lies, as concept relevance propagation does
\citep{Achtibat2023}, and a column which channels carry the evidence of
a region, with exact accounting in both directions.
LRPN$_{\alpha(\bar x),\beta(1-\bar x)}$ gives a close decomposition,
more diffuse as its pixel maps are (Section~\ref{sec:app-images}): the
bias nodes keep $\RPsharebiasxvn\%$ of $x_o$ instead of
$\RPsharebiaslrpzp\%$, the body and the background gain a few points,
the head carries $\RPshareheadxvn\%$ of $x_o$ and $\RPpixheadxvn\%$ of
the pixel relevance, with a density $\RPdensxvn$ times that of the body,
and its evidence needs more channels ($\RPhalffivexvn$ at conv5\_3,
$\RPhalffourxvn$ at conv4\_3), with a similar purity
($\RPpurityxvn\%$) and $\RPcommonxvn$ channels common to the head and
the body. The relevance by region therefore separates the rules as their
pixel maps do, and says in addition through which hidden units each part
of the image reaches the output. These shares account for the flow of the
rule chosen, on this sample and these masks; they are not a causal test
of the units.
\FloatBarrier

\section{Conclusion}
\label{sec:conclusion}

We have shown that relevance propagation for feed-forward neural
networks, classical or otherwise, is naturally cast as an entropic
projection: the redistribution of a forward activation flow onto a
set of admissible relevance flows on the network's active
computational graph, measured by a relative entropy between backward
transition laws, which, for a non-negative reference, is a relative
entropy on backward paths, as in Schrödinger's problem of entropic
optimal transport. This variational principle rests on a
measure-theoretic Filippov formalization of feed-forward networks
(Section~\ref{sec:feedforward}), which covers smooth and
piecewise-smooth architectures, such as ReLU and max-pooling networks,
alongside outright discontinuous ones, such as hard-routed
Mixture-of-Experts, and which, recast as a discrete-time optimal control
problem, also characterizes backpropagation as the adjoint equation of a
Lagrangian relaxation. On this formalization, the classical LRP$_{z^+}$
and LRP$_{\alpha,\beta}$ rules are the unique solutions of
relative-entropy minimization problems between backward transition
laws, and so is LRP$_0$, its reference being the signed forward
contributions themselves (Section~\ref{sec:LRP}); taking the relative
entropy edge by edge, as in entropically regularized flows, yields a new
rule, Sinkhorn Relevance Propagation (Section~\ref{sec:SRP}).

The same principle lets the weights of LRP$_{\alpha,\beta}$ vary from
node to node, a family, LRPN$_{\alpha,\beta}$, that contains LRP$_0$,
LRP$_{z^+}$ and LRP$_{\alpha,\beta}$ and keeps the sign of the output
whenever its weights lie in $[0,1]$
(Corollary~\ref{cor:lrp-references}); with weights that grow with the
activity of the node, a more active node sends more of its relevance to
its excitatory contributions, a Hebbian-inspired reading. On networks of ReLU and max-pooling
nodes, each rule of this family is an input times a modified backward
pass (Corollary~\ref{thm:RP0ReLU}): LRP$_0$ is Gradient$\times$Input,
linear on each region of constant gates, with an explicit jump across an
isolated kink, while LRP$_{z^+}$ is continuous there and, when the maxima
are non-negative, bounded by the input times a path sum of the weights; where the winner of a max-pooling
node changes, every rule may jump, the relevance of the node moving from one
winner to the other.

Aggregated over a region of the input, the flow of a non-negative
reference tells which hidden units carry the evidence of that region,
with exact accounting; the flow conditioned on the region is a
Schrödinger bridge, whose potentials follow from one forward pass
(Section~\ref{sec:region}).

The applications of Sections~\ref{sec:app-images}
and~\ref{sec:app-flows} check these statements on VGG-16: the
conservation identities hold to round-off, and a perturbation of
standard deviation $10^{-5}$ moves the LRP$_0$ map by $\KINKlrpzero\%$,
against $\KINKlrpzp$ to $\KINKxvn\%$ for the other rules, consistently
with Corollary~\ref{thm:RP0ReLU}. They also show
that the choice of the reference is not a detail: the rules differ in
the sign of their maps, in the share of the output they leave to the
bias nodes, in their faithfulness, in their localization and in their
dependence on the class. SRP follows the edges and textures of the
lower layers and is less faithful than LRP$_{z^+}$; with the scaled
reference, SRP$_1$ comes close to the faithfulness of LRP$_{z^+}$, at a
higher cost. Node-wise weights normalized by the activations
of their layer bring LRPN$_{\alpha,\beta}$ close to LRP$_{z^+}$ on
VGG-16. On ConvNeXt-Tiny, whose hidden values take both signs,
LRPN$_{\alpha,\beta}$ conserves the output and these weights give the
highest ABPC of the rules tested, whereas LRP$_{z^+}$ loses part of the output where the
hypothesis of Corollary~\ref{cor:lrp-references}(ii) fails and LRP$_0$ is
no longer Gradient$\times$Input. On $\RPN$ annotated photographs of cats
and dogs, the relevance by region gives the head about $\RPdenslrpzp$
times as much relevance per pixel as the body (median), carried by
largely different channels.

Two directions follow from this work. First, the formalism of
Section~\ref{sec:feedforward} reaches well beyond the image classifiers
evaluated here: hard-routed Mixture-of-Experts, for which its
non-generic machinery was designed, and normalization layers entering
the graph as nodes, rather than frozen at the input as in
Section~\ref{sec:app-images}; changing the reference of the
projection while keeping its constraint gives further conservative rules
for them, which the indicators of Section~\ref{sec:app-images} can
compare. Second, the same entropic-projection language that recovers
LRP$_{\alpha,\beta}$ suggests relaxing the conservation constraint into
a soft, tunable marginal penalty, in the sense of unbalanced optimal
transport \citep{Chizat2018,Sejourne2023,LieroMielkeSavare2018}, when an
exactly conserved flow is not what the modelling goal calls for. The characterization is relative to the reference: it does not by
itself select one, nor guarantee faithfulness, which the applications
compare. Taken
together, these results suggest that optimal transport is not merely a
convenient reformulation of existing relevance-propagation heuristics,
but the organizing principle from which they, and their natural
extensions, follow.

\clearpage
\bibliographystyle{plainnat}
\bibliography{biblio} %

\newpage
\appendix

\section{Proofs for Section~\ref{sec:feedforward}: Feed-Forward Filippov Neural Networks}

\subsection{Proof of Theorem \ref{thm:filippov-piecewise}}
\label{proof:thm:filippov-piecewise}

Let $\Phi=\pm1$ on $\{\pm z>0\}\subset\mathbb R$, $\Sigma=\{0\}$. The closed convex hull fills the
jump: $\mathcal F[\Phi](0)=[-1,1]$. The intersection over $\delta$ localizes:
$\mathcal F[\Phi](\tfrac12)=\{1\}$, although $\Phi(B(\tfrac12,\delta)\setminus\Sigma)=\{-1,1\}$ for
$\delta>\tfrac12$. The intersection over null sets ignores negligible values: if
$\tilde\Phi=\Phi$ except $\tilde\Phi(1/\ell)=5$, $\ell\in\mathbb N^*$, then
$\overline{\operatorname{co}}\,\tilde\Phi(B(0,\delta)\setminus\Sigma)=[-1,5]$ for every $\delta$,
and $\mathcal F[\tilde\Phi](0)=[-1,1]$; in general $\mathcal F$ only depends on the class of the map
modulo null sets.

Let $x\in\Omega$ and $J:=J_\Phi(x)$. The family $(\overline{\Omega^{(k)}})_k$ being locally finite,
$J$ is finite and there is $\delta_0>0$ with $B(x,\delta_0)\subset\Omega$ and $J_\Phi(y)\subseteq J$
on $B(x,\delta_0)$. If $J=\varnothing$, a neighborhood of $x$ lies in $\Sigma$, against
$\mu(\Sigma)=0$.

Let $\Psi:\Omega\setminus\Sigma\to\mathbb R^p$ agree on each $\Omega^{(k)}$ with a map $\Psi^{(k)}$
continuous on $\overline{\Omega^{(k)}}\cap\Omega$; then
\[
\bigcap_{\mu(\mathcal N)=0}\ \bigcap_{\delta>0}
\overline{\operatorname{co}}\,\Psi\bigl(B(x,\delta)\setminus\mathcal N\bigr)
=\operatorname{co}\bigl\{\Psi^{(k)}(x):k\in J\bigr\}=:C .
\]
Let $k\in J$, $\mathcal N$ null and $\delta,\varepsilon>0$. For $\delta'\le\delta$ small, the open
set $B(x,\delta')\cap\Omega^{(k)}$ is nonempty, hence not contained in $\mathcal N$, and a point $y$
of it outside $\mathcal N$ has $\|\Psi(y)-\Psi^{(k)}(x)\|<\varepsilon$. So $\Psi^{(k)}(x)$ lies in
the left-hand side, which is convex, and $C$ too. Conversely, for $\varepsilon>0$ and $\delta$ small,
$\Psi(B(x,\delta)\setminus\Sigma)$ lies in the closed $\varepsilon$-neighborhood of the compact convex
set $C$, closed and convex; hence so does the left-hand side, for every $\varepsilon$.

Theorem~\ref{thm:filippov-piecewise}(i) is the case $\Psi=\Phi$ and the case $\Psi=(\Phi,D\Phi)$,
$\Psi^{(k)}=(\Phi^{(k)},D\Phi^{(k)})$, continuous since $\Phi^{(k)}$ is $C^1$ on $U^{(k)}$. If
$x\notin\Sigma$, then $x$ lies in a single open region $\Omega^{(k_0)}$, so $J=\{k_0\}$, and
$(\Phi,D\Phi)=(\Phi^{(k_0)},D\Phi^{(k_0)})$ there. $\hfill\blacksquare$

\subsection{Example for Theorem \ref{thm:ffnn-output}}
\label{thm:ffnn-output_example}

Take two inputs $i_1,i_2$, two hidden nodes $v_1,v_2$ and one output $o$, with
$w_{i_1,v_1}=w_{i_2,v_1}=w_{i_1,v_2}=1$, $w_{i_2,v_2}=-1$, $w_{v_1,o}=3$, $w_{v_2,o}=2$, and
\[
\begin{aligned}
&\Phi_b^{(1)}\equiv-1\ \text{on}\ \{a+c<0\},&&
\Phi_b^{(2)}\equiv+1\ \text{on}\ \{a+c>0\}\qquad(b=v_1,v_2),\\
&\Phi_o^{(1)}=(z_1+z_2)^2\ \text{on}\ \{z_1+z_2<1\},&&
\Phi_o^{(2)}=z_1+z_2+3\ \text{on}\ \{z_1+z_2>1\},
\end{aligned}
\]
so that $z_{v_1}=(x_{i_1},x_{i_2})$, $z_{v_2}=(x_{i_1},-x_{i_2})$ and
$z_o=(3x_{v_1},2x_{v_2})$.

At $x=(2,1)$, $x\notin\mathcal D$, $x_{v_1}=x_{v_2}=1$, $z_o=(3,2)\in\Omega_o^{(2)}$, $x_o=8$,
$\mathcal M(x)=\{(2,2,2)\}$, $\partial x_o/\partial x_{v_1}=3$, $\partial x_o/\partial x_{v_2}=2$,
$\partial x_o/\partial x_{i_1}=\partial x_o/\partial x_{i_2}=0$ and
$\mathfrak F_{v_1,o}(x)=\{(8,3)\}$.

At $x=(1,-1)$, $z_{v_1}=(1,-1)\in\Sigma_{v_1}$, so $X_{v_1}(x)=[-1,1]$, $X_{v_2}(x)=\{1\}$ and
$Z_o(x)=[-3,3]\times\{2\}$, cut by $\Sigma_o$ at $z_1=-1$:
\[
X_o(x)=\underbrace{\{(z_1+2)^2:z_1\in[-3,-1)\}}_{=[0,1]}
\ \cup\ \underbrace{\operatorname{co}\{1,4\}}_{z_1=-1}
\ \cup\ \underbrace{\{z_1+5:z_1\in(-1,3]\}}_{=(4,8]}
=[0,8].
\]
By \eqref{eq:branch-admissible}, $\kappa_{v_2}=2$, $\kappa_{v_1}\in\{1,2\}$ is free and $\kappa_o$
is forced by $z_o^{(\kappa)}(x)$. Hence $\mathcal M(x)=\{\kappa',\kappa''\}$ and, by
\eqref{eq:branch-derivative},
\[
\begin{array}{c|c|c|c|c|c|c}
 & (\kappa_{v_1},\kappa_{v_2},\kappa_o) & x_{v_1}^{(\kappa)} & z_o^{(\kappa)} &
 x_o^{(\kappa)} & g^{(\kappa)}_{v_1,o} & g^{(\kappa)}_{v_2,o}\\\hline
\kappa' & (1,2,1) & -1 & (-3,2) & 1 & 3\times(-2)=-6 & 2\times(-2)=-4\\
\kappa'' & (2,2,2) & +1 & (3,2) & 8 & 3\times1=3 & 2\times1=2
\end{array}
\]
with $g^{(\kappa')}_{i_1,o}=g^{(\kappa'')}_{i_1,o}=0$, so that
\[
\mathfrak F_{v_1,o}(x)=\operatorname{co}\{(1,-6),(8,3)\},\qquad
\mathfrak F_{v_2,o}(x)=\operatorname{co}\{(1,-4),(8,2)\},\qquad
\mathfrak F_{i_1,o}(x)=[1,8]\times\{0\}.
\]
Thus $\{x_o^{(\kappa)}(x)\}=\{1,8\}\subsetneq[0,8]=X_o(x)$: the values in $[0,1)\cup(1,4]$ come from
the interior of the hull at $v_1$ and are reached by no selection.

At $x=(2,1)$, \eqref{eq:adjoint-recursion} gives $\lambda_{v_1}=3$, $\lambda_{v_2}=2$,
$\lambda_{i_1}=\lambda_{i_2}=0$, and Corollary~\ref{cor:backprop-weight-gradient} gives
$\partial x_o/\partial w_{v_1,o}=1$ and $\partial x_o/\partial w_{i_1,v_1}=3\times2\times0=0$. At
$x=(1,-1)$, the recursion gives $(\lambda_{v_1},\lambda_{v_2})=(-6,-4)$ for $\kappa'$ and $(3,2)$
for $\kappa''$, with $\lambda_{i_1}=\lambda_{i_2}=0$.

\subsection{Proof of Theorem \ref{thm:ffnn-output}}
\label{thm:ffnn-output_proof}

By Theorem~\ref{thm:filippov-piecewise}(i), $\mathcal F[\Phi_v](\zeta)$ is the interval between the
extremes of the $\Phi_v^{(k)}(\zeta)$, $k\in J_v(\zeta)$, nonempty and finite, and it is
$\{\Phi_v(\zeta)\}$ at a regular $\zeta$; this is the formula of the statement. Let the $X_a(x)$,
$a\in\operatorname{In}(v)$, be nonempty and compact; so is $Z_v(x)$. Near $\zeta_0$,
$J_v(\zeta)\subseteq J_v(\zeta_0)$ and the branches of $J_v(\zeta_0)$ are continuous, so
$\mathcal F[\Phi_v](\zeta)$ lies in the $\varepsilon$-neighborhood of
$\mathcal F[\Phi_v](\zeta_0)$. A finite cover of $Z_v(x)$ thus bounds $X_v(x)$; and if
$y_n\in\mathcal F[\Phi_v](\zeta_n)$, $y_n\to y$, $\zeta_n\to\zeta$, then
$y\in\mathcal F[\Phi_v](\zeta)$. Hence $X_v(x)$ is compact, along a topological order.

$\mathcal M(x)$ is finite. Along a topological order, the predecessors of $v$ fix
$z_v^{(\kappa)}(x)$ and $J_v(z_v^{(\kappa)}(x))\neq\varnothing$ provides $\kappa_v$: $\mathcal
M(x)\neq\varnothing$. For $\kappa\in\mathcal M(x)$, induction gives $z_v^{(\kappa)}(x)\in Z_v(x)$
and $x_v^{(\kappa)}(x)=\Phi_v^{(\kappa_v)}(z_v^{(\kappa)}(x))\in
\mathcal F[\Phi_v](z_v^{(\kappa)}(x))\subseteq X_v(x)$.

Replace the value of $v$ by $t$ and recompute its descendants with the branches $\kappa_u$. At
$t_0=x_v^{(\kappa)}(x)$, $z_u(t_0)=z_u^{(\kappa)}(x)\in U_u^{(\kappa_u)}$, so each $x_u(t)$ is $C^1$
near $t_0$, $\dot x_v=1$, $\dot x_i=0$ at the inputs $i\neq v$, and
$\dot x_u=\sum_{a\in\operatorname{In}(u)}w_{a,u}[D\Phi_u^{(\kappa_u)}(z_u^{(\kappa)}(x))]_a\,\dot x_a$.
A path of $\mathcal P(v,u)$ is a path of $\mathcal P(v,a)$ followed by $(a,u)$, which gives
\eqref{eq:branch-derivative} by induction. $\mathfrak F_{v,o}(x)$ is the hull of a finite nonempty
set.

Let $D_v:=\{x:Z_v(x)\cap\Sigma_v\neq\varnothing\}$ and $A_v$ the union of the $D_a$ over the
non-input strict ancestors $a$ of $v$. Let $x\notin A_v$ and $\kappa\in\mathcal M(x)$. Along a
topological order, each ancestor $a$ of $v$ has $Z_a(x)=\{z_a^{(\kappa)}(x)\}$, outside $\Sigma_a$;
by Theorem~\ref{thm:filippov-piecewise}(ii), $\kappa_a$ is its unique active branch and
$X_a(x)=\{x_a^{(\kappa)}(x)\}$. Hence $Z_v(x)=\{z_v^{(\kappa)}(x)\}$ and
\[
D_v\subseteq A_v\cup\bigl\{x:z_v^{(\kappa)}(x)\in\Sigma_v\text{ for some }\kappa\in\mathcal M(x)\bigr\} .
\]
Non-degeneracy gives $\mu(D_v)=0$ by induction along a topological order, and $\mu(\mathcal D)=0$.

If $x\notin\mathcal D$, this holds at every node: every $X_u(x)$ is a singleton, every $z_u(x)$ lies
in an open region $\Omega_u^{(k_u)}$, and $\mathcal M(x)=\{\kappa\}$, $\kappa_u=k_u$. These regions
being open, $\kappa$ remains the network when the value of $v$ moves near $t_0$, so
$\partial x_o/\partial x_v(x)=g^{(\kappa)}_{v,o}(x)$.

Assume now that the branches agree on the switching sets. Then $\mathcal F[\Phi_v](\zeta)$ is a
singleton for every $\zeta$, the extension of $\Phi_v$ to $\Sigma_v$ is continuous, and, along a
topological order, $X_v(x)=\{x_v^{(\kappa)}(x)\}$ for every $\kappa\in\mathcal M(x)$, continuous in $x$. Let
$x_0\in\mathbb R^d$. On a ball $B$ around $x_0$, every selection admissible at a point belongs to
$\mathcal M(x_0)$: otherwise a single $\kappa\notin\mathcal M(x_0)$ is admissible at points
$x_n\to x_0$; then $z_u^{(\kappa)}(x_n)=z_u(x_n)\in\overline{\Omega_u^{(\kappa_u)}}$, where $z_u$ is
the pre-activation of the network, continuous, so $z_u(x_0)\in\overline{\Omega_u^{(\kappa_u)}}$, and
induction gives $z_u^{(\kappa)}(x_0)=z_u(x_0)$, the branches agreeing there: $\kappa\in\mathcal
M(x_0)$. So $x_o$ is on $B$ a continuous selection of the $C^1$ maps $x_o^{(\kappa)}$,
$\kappa\in\mathcal M(x_0)$, hence locally Lipschitz \citep{Scholtes2012}.

If $x\notin\mathcal D$, the regions being open, $x_o=x_o^{(\kappa)}$ near $x$, $C^1$, of gradient
$(g^{(\kappa)}_{i,o}(x))_i$. Let $x$ be arbitrary, $g^\kappa:=(g^{(\kappa)}_{i,o}(x))_{i\in\mathcal
I\setminus\mathcal B}$, $v\in\mathbb R^d$, $m:=\max_{\kappa\in\mathcal M(x)}\langle g^\kappa,v\rangle$ and
$\varepsilon>0$. On a ball $B$ around $x$, $x_o$ is a selection of the $x_o^{(\kappa)}$,
$\kappa\in\mathcal M(x)$, and $x_o^{(\kappa)}(y')-x_o^{(\kappa)}(y)\le\langle
g^\kappa,y'-y\rangle+\varepsilon\|y'-y\|$. Let $y+sv\in B$, $s\in[0,t]$. Near $y+sv$, by continuity,
$x_o$ only equals those maps with the value $x_o(y+sv)$ at $y+sv$, so
$s\mapsto x_o(y+sv)$, continuous, has right increments at most $m+\varepsilon\|v\|$ per unit length and
$x_o(y+tv)-x_o(y)\le(m+\varepsilon\|v\|)t$. Therefore
\[
x^\circ_o(x;v):=\limsup_{y\to x,\ t\downarrow0}\frac{x_o(y+tv)-x_o(y)}{t}\le m ,
\]
and every $\xi\in\partial^\circ x_o(x)=\{\xi:\langle\xi,v\rangle\le x^\circ_o(x;v)\ \forall v\}$
\citep{Clarke1975} lies in $\operatorname{co}\{g^\kappa:\kappa\in\mathcal M(x)\}$ by separation.
$\hfill\blacksquare$

\subsection{Proof of Theorem \ref{thm:backprop-adjoint}}
\label{proof:thm:backprop-adjoint}

$\zeta_v(\xi^{(\kappa)})=z_v^{(\kappa)}(x)\in U_v^{(\kappa_v)}$ and $\zeta_v$ is linear, so
$\mathcal L^{(\kappa)}$ is $C^1$ near $\xi^{(\kappa)}$. The variable $\xi_v$ enters the cost if $v=o$,
the constraint of $v$ with coefficient $-\lambda_v$, and the constraint of each
$u\in\operatorname{Out}(v)$ through $w_{v,u}\xi_v$:
\[
\frac{\partial\mathcal L^{(\kappa)}}{\partial\xi_v}(\xi^{(\kappa)},\lambda)
=\mathbf 1_{\{v=o\}}-\lambda_v
+\sum_{u\in\operatorname{Out}(v)}\lambda_u\,w_{v,u}\,
\bigl[D\Phi_u^{(\kappa_u)}\bigl(z_u^{(\kappa)}(x)\bigr)\bigr]_v ,
\]
which vanishes for every $v$ if and only if \eqref{eq:adjoint-recursion} holds. The right-hand side
of \eqref{eq:adjoint-recursion} only involves the successors of $v$: the solution is unique, along a
reverse topological order. A path of $\mathcal P(v,o)$, $v\neq o$, is an edge $(v,u)$ followed by a
path of $\mathcal P(u,o)$, and $\mathcal P(o,o)$ is the empty path: $g^{(\kappa)}_{v,o}(x)$ solves
the recursion. The rest follows from \eqref{eq:admissible-value-gradient} and
Theorem~\ref{thm:ffnn-output}(iii). \qedhere

\subsection{Proof of Corollary \ref{cor:backprop-weight-gradient}}
\label{proof:cor:backprop-weight-gradient}

The $z_n$ lying in open regions, $\kappa$ remains the network for $w_{u,v}$ near its value, the node
values $\xi(w_{u,v})$ are $C^1$ and satisfy the state equations, so that
$x_o=\mathcal L^{(\kappa)}(\xi(w_{u,v}),\lambda^{(\kappa)},w_{u,v})$. Differentiating,
$\partial\mathcal L^{(\kappa)}/\partial\xi_n=0$ leaves
\[
\frac{\partial x_o}{\partial w_{u,v}}
=\frac{\partial\mathcal L^{(\kappa)}}{\partial w_{u,v}}
=\lambda_v\,\bigl[D\Phi_v^{(\kappa_v)}(z_v)\bigr]_u\,x_u .
\]
\eqref{eq:adjoint-recursion} reads $\lambda_v=\mathbf 1_{\{v=o\}}+\sum_{r\in\operatorname{Out}(v)}
w_{v,r}\delta_{v,r}$; multiplied by $[D\Phi_v^{(\kappa_v)}(z_v)]_u$, it is
\eqref{eq:edge-costate-recursion}. Conversely, for a solution $d$ of
\eqref{eq:edge-costate-recursion}, $\mathbf 1_{\{v=o\}}+\sum_rw_{v,r}d_{v,r}$ solves
\eqref{eq:adjoint-recursion}, so it is $\lambda_v$, and $d=\delta$. \qedhere

\FloatBarrier

\section{Proofs for Section~\ref{sec:LRP}: Relevance Propagation as an Entropic Projection}

\subsection{Proof of Proposition \ref{thm:RelevanceConservation}}
\label{proof:thm:RelevanceConservation}

Let $\Sigma$ be the sum of the divergences over $H(x)$ and of the inflows at the outputs. Every edge
of $E(x)$ counts once with $+$ at its head and with $-$ at its tail if the tail is hidden, so
$\Sigma=\sum_{u\in\mathcal I}\sum_{v:u\to v}\pi^o_{u,v}=\sum_{u\in\mathcal I}\mu^o_u$. The constraints
of $\mathcal A_o(x)$ give $\Sigma=x_o$. The same holds with $x_o$ replaced by any $b\in\mathbb R$.
\qedhere

\subsection{Proof of Theorem \ref{thm:RP0}}
\label{proof:thm:RP0}

Fix $\kappa$ and drop it; $\pi^\ast:=\pi^{o,\kappa,\ast}$. If $\tilde\rho_v\neq0$,
$\sum_uK_{u,v}=1$ and $\bar K_{\cdot,v}=|K_{\cdot,v}|/\|K_{\cdot,v}\|_1$ is a probability vector. For
probability vectors $q,k$, $\log t\le t-1$, with equality only at $t=1$, gives
$q_u\log(q_u/k_u)\ge q_u-k_u$, so $\mathrm{KL}(q\|k)\ge0$ with equality if and only if $q=k$.

Let $n_v(\pi)\neq0$, $\pi$ coherent at $v$, $q_u:=|\pi_{u,v}|/n_v(\pi)$. The term of
\eqref{eq:cond-entropy} at $v$ is $+\infty$ if $\tilde\rho_v=0$; otherwise it is $\ge0$, and it
vanishes if and only if $|\pi_{u,v}|=c_v|K_{u,v}|$ for all $u$, $c_v:=n_v(\pi)/\|K_{\cdot,v}\|_1>0$.
Coherence then gives $\pi_{u,v}=\sigma_vc_vK_{u,v}$, and summing over $u$,
$\pi_{u,v}=K_{u,v}m_v(\pi)$ with $m_v(\pi)\neq0$. Conversely, if $\tilde\rho_v\neq0$ and
$\pi_{u,v}=K_{u,v}m$, $m\neq0$, then $\pi$ is coherent at $v$ with $\sigma_v=\mathrm{sgn}(m)$,
$n_v(\pi)=|m|\,\|K_{\cdot,v}\|_1$ and $q=\bar K_{\cdot,v}$. A node with $n_v(\pi)=0$ has
$\pi_{u,v}=0$ and contributes nothing. Hence $\mathcal H(\pi\|K)\ge0$, and
\begin{equation}
    \label{eq:zero-set}
    \mathcal H(\pi\|K)=0\iff\pi_{u,v}=K_{u,v}\,m_v(\pi)\quad\text{for every }(u,v)\in E(x).
\end{equation}

Let $\pi\in\mathcal A_o(x)$ satisfy \eqref{eq:zero-set}. The constraints give $\pi_{u,o}=K_{u,o}x_o$
and $\pi_{u,o'}=0$; along a reverse topological order, at a hidden $v$, the divergence gives
$m_v(\pi)=\sum_{w:v\to w}\pi_{v,w}=:\mu_v$, already known, and $\pi_{u,v}=K_{u,v}\mu_v$. So
$\pi=\pi^\ast$, $\pi^\ast_{u,v}:=K_{u,v}\mu_v$, $\mu$ being given by the backward recursion.
Conversely, $\tilde\rho_v\neq0$ on $H(x)\cup\{o\}$ gives $m_v(\pi^\ast)=\mu_v$ there, the recursion
gives the outflow $\mu_v$ at a hidden $v$, $\mu_{o'}=0$ gives $m_{o'}(\pi^\ast)=0$: $\pi^\ast\in\mathcal
A_o(x)$, and it satisfies \eqref{eq:zero-set}. Thus $\pi^\ast$ is the unique minimizer, of value $0$.
Proposition~\ref{thm:RelevanceConservation} gives $\sum_{u\in\mathcal I(x)}\mu_u=x_o$. If neither the
node values nor the reference depend on $\kappa$, neither do the active graph and the
$\mu_u^{o,\kappa}$. \qedhere

\subsection{Path-space reading of the conditional entropy \eqref{eq:cond-entropy}}
\label{proof:path-space}

Under the assumptions of Theorem~\ref{thm:RP0}, let $x_o>0$, $\rho\ge0$, so that $\bar K=K$,
$\pi\in\mathcal A_o(x)$ non-negative, so that
$n_v(\pi)=m_v(\pi)$, with $\mathcal H(\pi\|K)<\infty$, and $\Pi$ the set of backward paths
$p=(o=v_0,\ldots,v_r)$, $v_{j+1}\to v_j$ in $E(x)$, $v_r\in\mathcal I(x)$. Let
\[
P_\pi(p):=x_o\prod_{j=0}^{r-1}q_{v_{j+1},v_j},
\qquad
Q_K(p):=x_o\prod_{j=0}^{r-1}K_{v_{j+1},v_j},
\qquad q_{u,v}:=\frac{\pi_{u,v}}{m_v(\pi)} ,
\]
$q_{\cdot,v}$ being any transition law where $m_v(\pi)=0$, be the backward chains from $o$ of mass
$x_o$. Then $\mathrm{KL}(P_\pi\|Q_K)=\mathcal H(\pi\|K)$.

Along a reverse topological order, the mass of the paths through a node $v\notin\mathcal I(x)$ is
$m_v(\pi)$: it is $x_o$ at $o$, and $\sum_{w:v\to w}m_w(\pi)q_{v,w}=\sum_w\pi_{v,w}=m_v(\pi)$ at a
hidden node, as $\pi_{v,w}=0$ where $m_w(\pi)=0$; through an input $i$, it is its outflow
$\sum_w\pi_{i,w}$. So the paths using $(u,v)$ have mass $\pi_{u,v}$, and those ending at an input
carry its outflow. Expanding
the products and exchanging the sums,
\[
\mathrm{KL}(P_\pi\|Q_K)=\sum_{(u,v)\in E(x)}\pi_{u,v}\log\frac{q_{u,v}}{K_{u,v}}
=\sum_{(u,v)\in E(x)}\pi_{u,v}\log\frac{\pi_{u,v}}{K_{u,v}\,m_v(\pi)}
=\mathcal H(\pi\|K). \qquad\blacksquare
\]

\subsection{Proof of Corollary \ref{cor:lrp-references}}
\label{proof:cor:lrp-references}

Each case checks $\tilde\rho_v\neq0$ on $H(x)\cup\{o\}$, computes $K=\rho/\tilde\rho$ and applies
Theorem~\ref{thm:RP0}.

For $\rho=z$, $\tilde\rho_v=\tilde z_v\neq0$, $K_{u,v}=z_{u,v}/\tilde z_v$. If moreover
$\tilde z_{o'}\neq0$, then $\bar K_{u,v}=|z_{u,v}|/\sum_{r:r\to v}|z_{r,v}|$ at every non-input node
(the formula can fail at an output with $\tilde z_{o'}=0$, where $\bar K=0$), and, $\tilde z_v$
having a fixed sign, coherence with $K$ says that the products $\pi_{u,v}z_{u,v}$ entering $v$ have a
common sign.

For $\rho=z^+$, $\tilde\rho_v=\tilde z^+_v>0$, $K_{u,v}=z^+_{u,v}/\tilde z^+_v$. For
LRP$_{\alpha,\beta}$, $\tilde\rho_v=\alpha+\beta=1$ and $K=\rho$, both channels being non-empty.

For LRPN$_{\alpha,\beta}$, the two channels are not both empty, and
$\sum_uz^\pm_{u,v}/\tilde z^\pm_v=1$ when $\tilde z^\pm_v>0$, so $\tilde\rho_v=\alpha_v+\beta_v=1$
when both are non-empty and $\tilde\rho_v=1$ otherwise: $K=\rho$. Theorem~\ref{thm:RP0} only
sees the numbers $\rho_{u,v}$, whatever the dependence of the weights on $v$ and $x$. \qedhere

\subsection{Proof of Corollary \ref{thm:RP0ReLU}}
\label{proof:thm:RP0ReLU}

On $E(x)$, $K_{u,v}x_v=x_u\,\omega_{u,v}s_v$. The recursion of $\mu$, each path of $\mathcal P(u,o)$
decomposing uniquely as a first edge followed by a path, gives $\mu_u=x_uW^\rho_u$ by induction along a
reverse topological order; off $V(x)$, $x_u=0=\mu_u$.

A kernel that vanishes with $z_{u,v}$ has the same relevance on $G(x)$ and on $\mathcal G$, a node of
value $0$ sending nothing, and $\tilde\rho_v$ is the same on both at $v\in V(x)$. So LRP$_0$ is the
relevance on $\mathcal G$ of the kernel $\rho_{u,v}/\tilde\rho_v$. At a linear or ReLU hidden node or at
$o$: if $\tilde z_v\neq0$, $x_v=\eta_{u,v}\tilde z_v$ gives $K_{u,v}x_v=x_uw_{u,v}\eta_{u,v}$; if
$\tilde z_v=0$, then $K_{u,v}=0$ and $\eta_{u,v}=0$, a linear hidden node with $\tilde z_v=0$ being
excluded and $\tilde z_o\neq0$. At a max node, the reference has the single term
$z_{\kappa_v,v}=x_v$, so that $K_{u,v}x_v=z_{u,v}\eta_{u,v}=x_uw_{u,v}\eta_{u,v}$, both sides vanishing
if $x_v=0$. As above, $\mu_u=x_uW_u$, with $W_u=\sum_{v:u\to v}w_{u,v}\eta_{u,v}W_v$, $W_o=1$,
$W_{o'}=0$; hence $W^\rho=W$ on $V(x)\setminus(\mathcal O\setminus\{o\})$.

If $x\notin\mathcal D$, Theorem~\ref{thm:ffnn-output}(iii) gives a unique $\kappa$ with every $z_v(x)$
in an open region: no gate changes near $x$, $x_o$ is affine there, and the branch Jacobians are the
gates, that of a max node $v$ being the coordinate vector of $\kappa_v$, so
$\partial x_o/\partial x_u(x)=W_u(x)$ by \eqref{eq:branch-derivative}. The recursion of $W$ is
\eqref{eq:adjoint-recursion}.

On $U$, the signs of the values and $\kappa$ fix $G(x)$, the channels, the cases of
Corollary~\ref{cor:lrp-references}(iv) and the kernels of the max nodes. The shares
$z^\pm_{u,v}/\tilde z^\pm_v$ are continuous, or zero on an empty channel, so the kernels and $\mu$ are
continuous. For LRP$_0$, with $\tilde z_v\neq0$ at the linear hidden nodes, the gates are constant, so
is $W$, and $\mu_u=x_uW_u$. For LRP$_{z^+}$ with ReLU hidden nodes, or max hidden nodes of non-negative value, and $x_o>0$, a reverse
topological order gives $0\le\mu_u\le|x_u|\bar W_u$: at a live ReLU head $v$,
$0<x_v=\tilde z_v\le\tilde z^+_v$, so $K_{u,v}\mu_v\le z^+_{u,v}\bar W_v\le|x_u|\,|w_{u,v}|\,\bar W_v$;
at a live max head, $x_v=z_{\kappa_v,v}>0$, and $K_{u,v}\mu_v$ is $0$, or
$\mu_v\le x_v\bar W_v=|x_u|\,|w_{u,v}|\,\bar W_v$ for $u=\kappa_v$; a dead head has $\mu_v=0$.

Near $\bar x$, in (iii) and (iv), $\kappa$ stays admissible at the max nodes other than the tie: a
single maximizer stays the only one, and a maximum $0$ attained at ReLU nodes $r$ with
$\tilde z_r(\bar x)<0$ stays $0$, the other contributions staying negative. Such a node acts as a
linear node with the single edge $(\kappa_r,r)$, whose value moves with that of $\kappa_r$, which is not
$v$ in (iii). Write $\bar\eta:=\eta(\bar x)$ and $\eta_r$ for the gate shared by the edges entering a
linear or ReLU node $r$, so $\bar\eta_v=0$ in (iii), and let $U$ be a
neighborhood of $\bar x$ on which every node $r\neq v$, hidden or $o$, keeps its gates. A path does not
cross $v$ twice, so $W_v(x)=W_v(\bar x)$ on $U$, and the first-edge recursion gives, by induction,
\[
W_u(x)=W_u(\bar x)+\eta_v(x)\,W^v_u(\bar x)\,W_v(\bar x)\qquad(u\neq v).
\]
With $\mu_u(x)=x_uW_u(x)$, this gives the two limits of LRP$_0$; at $v$, $\mu_v(x)=x_vW_v(\bar x)\to0$ and
$\bar x_v=0$. Along a topological order, for $r$ before $v$, $x_r=\bar\eta_r\tilde z_r$ at a linear
or ReLU node and $x_r=z_{\kappa_r,r}$ at a max node give $\sum_{i\in\mathcal I}\bar x_i\,\hat
W^r_i=\bar x_r$, $\hat W^r_i$ being the sum over $\mathcal P(i,r)$ of the products of $w\bar\eta$, gate
of $r$ included. Hence $\sum_{u\in\mathcal I}\bar x_uW^v_u(\bar x)=\sum_{r:r\to v}w_{r,v}\bar
x_r=\tilde z_v(\bar x)=0$ and $\sum_{u\in\mathcal I}\Delta_u=0$.

For LRPN$_{\alpha,\beta}$, the inputs and the $\tilde z_r(\bar x)$, $r\neq v$ linear or ReLU, are
nonzero, so on a neighborhood $U$ of $\bar x$ every node other than $v$ keeps the sign of its value, zero
at an inactive ReLU node, that of its winner at a max node, and so does every contribution it sends;
$G(x)=G(\bar x)$ where $\tilde z_v(x)<0$, and $G(x)$ is $G(\bar x)$ augmented by $v$ and its edges where
$\tilde z_v(x)>0$. By the recursion of $\mu$, it suffices that every kernel converge. Fix a node and a
channel. Where $\tilde z_v(x)<0$, $x_v=0$: the channel is empty or not as at $\bar x$, and its shares
converge to those of $\bar x$, so $\mu^-=\mu(\bar x)$. Where $\tilde z_v(x)>0$, at a successor $w$ of $v$
other than a max node, $z_{v,w}=w_{v,w}x_v\to0$ with the sign of $w_{v,w}$; a max successor keeps its
kernel, $v$ not being its winner. If the channel of that sign is non-empty at $\bar x$, its shares
converge to those of $\bar x$, $v$ having share $\to0$. If it is empty, $v$ fills it: share $1$ for $v$,
$0$ for the others. In all cases the kernels converge, to those of LRPN$_{\alpha,\beta}$ on the augmented
graph, at the values of $\bar x$, each successor $w$ counting $v$ in the channel of the sign of
$w_{v,w}$; they are the kernels of $\bar x$ if no channel is filled, and then $\mu^+=\mu(\bar x)$. For
LRP$_{z^+}$, $\beta_w=0$, so filling the negative channel of a node changes none of its kernels when its
positive channel is non-empty, which holds at the live nodes, and a positive channel can only be filled
at a dead node, of relevance $0$: $\mu^+=\mu(\bar x)$. Finally, the relevances of the inputs add up to
$x_o(x)$ at the inputs $x\in U$ with $\tilde z_v(x)>0$; as $x\to\bar x$,
$\sum_{u\in\mathcal I}\mu^+_u=x_o(\bar x)=\sum_{u\in\mathcal I}\mu_u(\bar x)$.

In (iv), the other predecessors of $v$ stay below $u$ and $u'$ near $\bar x$, so $\kappa$ is admissible
where $z_{u,v}>z_{u',v}$ and $\kappa'$ where $z_{u',v}>z_{u,v}$. No gate, sign or channel changes there,
and $x_v\neq0$, so $\mu=\mu^\kappa$ on the first side, continuous at $\bar x$ as in (ii), and
$\mu=\mu^{\kappa'}$ on the second. The kernels of $\kappa$ and $\kappa'$ differ only on the edges entering
$v$, where they are $1$ on $(u,v)$, resp.\ $(u',v)$, and $0$ on the others. Splitting each path of
$\mathcal P(i,o)$ through $v$ at the edge entering $v$, the paths to $u$ and $u'$ and the relevance
$\mu_v(\bar x)$ being the same for $\kappa$ and $\kappa'$,
\[
\mu^{\kappa'}_i(\bar x)-\mu^{\kappa}_i(\bar x)=\mu_v(\bar x)\sum_{a:a\to v}R^{(a)}_i\bigl(K^{\kappa'}_{a,v}-
K^{\kappa}_{a,v}\bigr)=\mu_v(\bar x)\bigl(R^{(u')}_i-R^{(u)}_i\bigr).
\]
Both conserve $x_o(\bar x)$, so $\sum_{i\in\mathcal I}\Delta_i=0$. For LRP$_0$, $\mu_v(\bar x)=\bar x_vW_v(\bar
x)$, the paths to $a$ give $R^{(a)}_i=\bar x_i\hat W^a_i(\bar x)/\bar x_a$ as in (i), and $\bar
x_v=w_{u,v}\bar x_u=w_{u',v}\bar x_{u'}$. \qedhere

\subsection{Proof of Theorem \ref{thm:srp}}
\label{proof:thm:srp}

Write $E^o:=E^o_\rho(x)$, $V^o:=V^o_\rho(x)$, and attach to $e=(u,v)\in E^o$ its arc, of tail
$\mathrm{tail}(e)$ and head $\mathrm{head}(e)$: $u\to v$ if $\rho_e>0$, $v\to u$ if $\rho_e<0$. For
$\nu\in\mathbb R^{V^o}$, $\nu_i=0$ at the inputs, let $\varphi:=e^\nu$ and $(\pi^+,\pi^-)$ be the pair
\eqref{eq:srp-form}: the mass $|\rho_e|e^{\nu_{\mathrm{head}(e)}-\nu_{\mathrm{tail}(e)}}$ along each
arc. Let
\[
    g(\nu):=\sum_{(u,v)\in E^o}\Bigl(\rho^+_{u,v}\bigl(1-e^{\nu_v-\nu_u}\bigr)
    +\rho^-_{u,v}\bigl(1-e^{\nu_u-\nu_v}\bigr)\Bigr)+y\,\nu_o ,
\]
strictly concave, the dual function of \eqref{eq:srp-problem}. Then $\partial g/\partial\nu_v$ is the
net outflow minus the net inflow of the pair at a hidden $v$, $\partial g/\partial\nu_o=y$ minus its
net inflow at $o$; multiplying by $\varphi_v$, the pair is feasible if and only if $\nabla g(\nu)=0$
if and only if $\varphi$ solves \eqref{eq:srp-fixed-point}.

Let $c\in\mathbb R$, $g(\nu)\ge c$, $M:=\max(0,\max_e(\nu_{\mathrm{head}(e)}-\nu_{\mathrm{tail}(e)}))$,
$r:=\min_e|\rho_e|$, $R:=\sum_e|\rho_e|$ and $L:=|V^o|$. Every node of $V^o$ is reached from an input by at
most $L$ arcs, so $\nu_v\le LM$, and $c\le g(\nu)\le R+r(1-e^M)+yLM$ bounds $M$. From $g(\nu)\le
R+y\nu_o$, $\nu_o$ is bounded below. From every node of $V^o$, at most $L$ arcs lead to an input or to
$o$, so $\nu_v\ge\min(0,\nu_o)-LM$. Thus $\{g\ge c\}$ is compact and $g$ has a maximizer $\nu^\ast$;
let $(\pi^{\mathrm S+},\pi^{\mathrm S-})$ be its pair, feasible, positive exactly where
$\rho^\pm>0$.

Let $(\pi^+,\pi^-)$ be feasible, of finite objective $J$. The difference $d$ of the two net flows
has zero divergence and zero inflow at $o$, so, summing by parts,
$\sum_{(u,v)}d_{u,v}(\nu_v-\nu_u)=0$ for every $\nu$ vanishing at the inputs. With $\nu^\ast$, as
$\log(\pi^{\mathrm S\pm}_{u,v}/\rho^\pm_{u,v})=\pm(\nu^\ast_v-\nu^\ast_u)$,
\[
J(\pi^+,\pi^-)=J(\pi^{\mathrm S+},\pi^{\mathrm S-})
+\mathrm{KL}(\pi^+\|\pi^{\mathrm S+})+\mathrm{KL}(\pi^-\|\pi^{\mathrm S-}),
\]
so $(\pi^{\mathrm S+},\pi^{\mathrm S-})$ is the unique minimizer. The same identity holds for every
feasible pair of the form \eqref{eq:srp-form}: such a pair is the minimizer, and, conversely, the
minimizer is feasible, whence \eqref{eq:srp-fixed-point}. On $E^o$, \eqref{eq:srp-form} fixes the
ratios of potentials along edges, every node of $V^o$ other than an input is reached from an input
along $E^o$, and $\varphi_i=1$ at the inputs: $\varphi$ is unique. Along a path of
positive reference from an input to $v$, the products telescope to $\varphi_v$.

Proposition~\ref{thm:RelevanceConservation} gives $\sum_{u\in\mathcal I(x)}\mu^{\mathrm S}_u=y$.

At a hidden $v$, $\partial g/\partial\nu_v$ is strictly decreasing in $\nu_v$ and vanishes exactly
where \eqref{eq:srp-fixed-point} holds at $v$, at
$\varphi_v=\bigl((\sum_w\rho^+_{v,w}\varphi_w+\sum_u\rho^-_{u,v}\varphi_u)/(\sum_u\rho^+_{u,v}/\varphi_u
+\sum_w\rho^-_{v,w}/\varphi_w)\bigr)^{1/2}$; at $o$, with $A^+_o:=\sum_u\rho^+_{u,o}/\varphi_u$ and
$A^-_o:=\sum_u\rho^-_{u,o}\varphi_u$, $\partial g/\partial\nu_o=y-\varphi_oA^+_o+A^-_o/\varphi_o$
vanishes at the positive root of $A^+_o\varphi_o^2-y\varphi_o-A^-_o$. So each update maximizes $g$ in one
coordinate, and the sweep $T$ is continuous, with $g\circ T\ge g$, and $g(T\nu)=g(\nu)$ only if
$\nabla g(\nu)=0$, that is, $\nu=\nu^\ast$. The iterates stay in the compact set $\{g\ge g(\nu^0)\}$ and
$g(\nu^{(t)})$ converges; at a limit point $\bar\nu$, $g(T\bar\nu)=g(\bar\nu)$, so $\bar\nu=\nu^\ast$,
and $\nu^{(t)}\to\nu^\ast$. \qedhere

\subsection{Proof of Proposition \ref{prop:srp-mass}}
\label{proof:prop:srp-mass}

$(\rho^+,0)$ is feasible, of objective $\mathrm{KL}(0\|\rho^-)=R^-\le R$, so $J\le R$ at the minimizer.
For $p,r\ge0$, $p\log(p/r)-p+r\ge p-(e-1)r$, the tangent at $p=er$ of the convex function
$p\mapsto p\log(p/r)$. Summing over the two sign parts, $J\ge M-(e-1)R$, with $M$ the total mass.
Hence $M\le eR$. \qedhere

\subsection{Proof of Proposition \ref{prop:srp-lrp}}
\label{proof:prop:srp-lrp}

The restriction of $\rho$ is feasible, of the form \eqref{eq:srp-form} with $\varphi\equiv1$: by
Theorem~\ref{thm:srp}(iii) it is $\pi^{\mathrm S}$. If every edge enters $o$, its source is an
input, $\pi^{\mathrm S}_{u,o}=\rho_{u,o}\varphi_o$, and the boundary condition gives
$\varphi_o=y/\sum_{(r,o)\in E^o_\rho(x)}\rho_{r,o}=y/\tilde\rho_o$, $\rho$ vanishing off
$E^o_\rho(x)$: this is $K_{u,o}\,y$. On the graph of the last statement, by
Theorem~\ref{thm:srp}(ii), the products of $\pi^{\mathrm S}/\rho$ along the two paths agree, and
conservation gives $\pi^{\mathrm S}_{i,h_k}=\pi^{\mathrm S}_{h_k,o}$, whence the stated ratio. For
Theorem~\ref{thm:RP0}, the flows into $o$ are $K_{h_k,o}\,y$. \qedhere

\FloatBarrier

\section{Proofs for Section~\ref{sec:transformers}: Relevance by Region}

\subsection{Proof of Corollary \ref{cor:region-relevance}}
\label{proof:cor:region-relevance}

Fix $\kappa$ and drop it. A path of $\mathcal P(u,v)$, $u\neq v$, is a path of $\mathcal P(u,w)$
followed by $(w,v)$, or an edge $(u,w)$ followed by a path of $\mathcal P(w,v)$:
\begin{equation}
    \label{eq:R-recursion}
    R^{(v)}_u=\sum_{w:\,w\to v}K_{w,v}\,R^{(w)}_u=\sum_{w:\,u\to w}K_{u,w}\,R^{(v)}_w
    \quad(u\neq v),\qquad R^{(v)}_v=1 .
\end{equation}
The second form is the recursion of Theorem~\ref{thm:RP0} from $v$ with unit mass, so
$\mu_u=x_o\,R^{(o)}_u$.

$R^{(v)}\ge0$. $R^{(v)}$ is the Dirac mass at $v$ if $v$ is an input; for $v\in H(x)\cup\{o\}$,
$\sum_uK_{u,v}=1$, and summing \eqref{eq:R-recursion} over $i$ gives $\sum_iR^{(v)}_i=1$ along a
topological order. By Appendix~\ref{proof:path-space}, with $q=K$, the path $p$ has mass
$Q_K(p)=x_o\prod_jK_{v_{j+1},v_j}$; a path through $v$ ending at $i$ splits at $v$, so these paths
have mass $x_o\,R^{(o)}_v\,R^{(v)}_i=\mu_vR^{(v)}_i$.

Summing over the regions of a partition gives $\sum_M\mu^M_v=\mu_v$. A path of positive mass ends
at an input and meets a cut $C$ exactly once, so $\sum_{v\in C}\mu_vR^{(v)}_i$ is the mass of the
paths ending at $i$, which is $\mu_i$ by Appendix~\ref{proof:path-space}.

Let $c:=x_o/\mu(M)$, $h(u):=R^{(u)}(M)$, so that $h(i)=\mathbf 1_M(i)$ at the inputs,
$h(v)=\sum_uK_{u,v}h(u)$ elsewhere by \eqref{eq:R-recursion}, and $h(o)=1/c$. Then $\pi^M\ge0$. Its
inflow at $v\in H(x)\cup\{o\}$ is $c\,\mu_vh(v)$, its outflow at a hidden $v$ too; it has inflow
$c\,x_oh(o)=x_o$ at $o$, $0$ at the other outputs, and outflow $c\,\mu_ih(i)$ at an input, and the
node relevances are $c\,\mu_uh(u)=c\,\mu^M_u$.

Let $\pi$ be feasible with $\mathcal H(\pi\|K)<\infty$. By Appendix~\ref{proof:path-space},
$\mathcal H(\pi\|K)=\mathrm{KL}(P_\pi\|Q_K)$, and the endpoint has law the outflow $\nu$ of the
inputs under $P_\pi$, $\mu$ under $Q_K$, with $\nu_i=c\,\mu_i\mathbf 1_M(i)$. By the chain rule for
relative entropy,
\[
\mathrm{KL}(P_\pi\|Q_K)=\mathrm{KL}(\nu\|\mu)+\sum_{i\in M}\nu_i\,
\mathrm{KL}\bigl(P_\pi(\cdot\mid i)\,\big\|\,Q_K(\cdot\mid i)\bigr)
\ \ge\ \mathrm{KL}(\nu\|\mu)=x_o\log c,
\]
with equality if and only if $P_\pi=P^M$, $P^M(p):=c\,Q_K(p)\,\mathbf 1_M(v_r)$. The ratios
$h(v_{j+1})/h(v_j)$ telescope to $c\,\mathbf 1_M(v_r)$ along a path of positive $P^M$-mass, so
$P^M$ is the chain of transitions $K_{u,v}h(u)/h(v)=\pi^M_{u,v}/m_v(\pi^M)$, any law at the nodes
where $h(v)=0$, which carry no $P^M$-mass: $P^M=P_{\pi^M}$. Hence
$\mathcal H(\pi^M\|K)=x_o\log c$, and a feasible $\pi$ attaining the bound has $P_\pi=P_{\pi^M}$,
so $\pi=\pi^M$, a flow being the edge marginal of its path measure. \qedhere

\FloatBarrier

\section{Architectures of VGG-16 and ConvNeXt-Tiny}
\label{sec:architectures}

Both networks are used with their ImageNet weights from torchvision, on
$224\times224$ colour images, and have $1000$ linear outputs, the logits.
Tables~\ref{tab:arch-vgg} and~\ref{tab:arch-convnext} list their layers;
the sizes are those of the output of each stage.

\begin{table}[H]
    \centering
    \caption{VGG-16 \citep{SimonyanZisserman2015}: $138.4$ million
    parameters, $71.6\%$ top-1 accuracy on ImageNet. Every convolution is
    $3\times3$, of stride $1$ and padding $1$, followed by a ReLU; every
    max-pooling has $2\times2$ windows of stride $2$.}
    \label{tab:arch-vgg}
    \small
    \begin{tabular}{lll}
        \toprule
        stage & layers & output size \\
        \midrule
        input & -- & $224\times224\times3$ \\
        block 1 & 2 convolutions, $64$ channels; max-pooling & $112\times112\times64$ \\
        block 2 & 2 convolutions, $128$ channels; max-pooling & $56\times56\times128$ \\
        block 3 & 3 convolutions, $256$ channels; max-pooling & $28\times28\times256$ \\
        block 4 & 3 convolutions, $512$ channels; max-pooling & $14\times14\times512$ \\
        block 5 & 3 convolutions conv5\_1--conv5\_3, $512$ channels; max-pooling & $7\times7\times512$ \\
        head & dense $4096$ + ReLU, dense $4096$ + ReLU, dense $1000$ & $1000$ \\
        \bottomrule
    \end{tabular}
\end{table}

As a Filippov network, VGG-16 is layered
(Definition~\ref{def:layered-DAG}): one input node per pixel and colour
channel, one ReLU node per unit of a convolution or dense layer
($\KINKunits$ million in all), with one bias node of value $1$ each, one
max node (Corollary~\ref{thm:RP0ReLU}) of four predecessors of weight $1$
per window of a max-pooling layer, and $1000$ linear outputs, each with
its bias node. The adaptive average pooling of torchvision is the identity
at this image size, and dropout is inactive at inference. The regions of
Section~\ref{sec:app-flows} are cut in conv5\_3, of size
$14\times14\times512$.

\begin{table}[H]
    \centering
    \caption{ConvNeXt-Tiny \citep{Liu2022ConvNeXt}: $28.6$ million
    parameters, $82.5\%$ top-1 accuracy on ImageNet. A block of width $C$
    maps $a$ to $a+\gamma\odot W_2\,\mathrm{GELU}\bigl(W_1\,
    \mathrm{LN}(D\,a)\bigr)$, with $D$ a depthwise $7\times7$ convolution
    (one kernel per channel, padding $3$), $\mathrm{LN}$ a layer
    normalization over the $C$ channels at each position, $W_1$, $W_2$ two
    $1\times1$ convolutions $C\to4C\to C$, and $\gamma\in\mathbb R^C$ the
    layer scale.}
    \label{tab:arch-convnext}
    \small
    \begin{tabular}{lll}
        \toprule
        stage & layers & output size \\
        \midrule
        input & -- & $224\times224\times3$ \\
        stem & convolution $4\times4$ of stride $4$, $96$ channels; LN & $56\times56\times96$ \\
        stage 1 & 3 blocks, $C=96$ & $56\times56\times96$ \\
        stage 2 & LN, convolution $2\times2$ of stride $2$; 3 blocks, $C=192$ & $28\times28\times192$ \\
        stage 3 & LN, convolution $2\times2$ of stride $2$; 9 blocks, $C=384$ & $14\times14\times384$ \\
        stage 4 & LN, convolution $2\times2$ of stride $2$; 3 blocks, $C=768$ & $7\times7\times768$ \\
        head & global average pooling, LN, dense $1000$ & $1000$ \\
        \bottomrule
    \end{tabular}
\end{table}

ConvNeXt-Tiny is not layered: each block adds its input to the output of
its branch. Read at $x$ as in Section~\ref{sec:app-images}, it has linear
and GELU nodes only, with no ReLU and no max node: a convolution or dense
unit is a linear node with its bias node, a depthwise unit having its
predecessors in its own channel only; the node $c$ of a layer
normalization at a position whose $C$ channels have mean $m$ and
stabilized standard deviation $\sigma_\varepsilon:=\bigl(C^{-1}\sum_{c'}(x_{c'}-m)^2+\varepsilon\bigr)^{1/2}$,
$\varepsilon=10^{-6}$ being the constant of the module, held constant,
receives the contributions
$g_c(\delta_{c,c'}-1/C)\,x_{c'}/\sigma_\varepsilon$ and the contribution $b_c$ of a bias
node, $g$ and $b$ being the affine parameters of the normalization; each unit of $W_1$ is a GELU node ($8.6$ million in all); the
layer scale $\gamma$ is folded into $W_2$ and its bias; a residual sum is
a node with two contributions of weight $1$; and the global average
pooling is a linear node of weights $1/49$.

\end{document}